\documentclass[11pt]{article}

\usepackage{amsmath , geometry , graphicx , amssymb, amsthm}
\usepackage{subcaption}
\usepackage{float}
\usepackage{adjustbox}
\usepackage[ruled,vlined,linesnumbered]{algorithm2e}

\usepackage[style=authoryear,citestyle=authoryear,natbib=true,backend=bibtex]{biblatex}
\newcommand{\figureWidth}{0.45}

\title{ Sequential Learning in the Dense Associative Memory }
\author{ Hayden McAlister$^{1}$, Anthony Robins$^{1}$, Lech Szymanski$^{1}$ }
\date{
  $^{1}$School of Computing, University of Otago, Dunedin, New Zealand.
}

\begin{document}
\maketitle	
\pagebreak

\begin{abstract}
Sequential learning involves learning tasks in a sequence, and proves challenging for most neural networks. Biological neural networks regularly conquer the sequential learning challenge and are even capable of transferring knowledge both forward and backwards between tasks. Artificial neural networks often totally fail to transfer performance between tasks, and regularly suffer from degraded performance or catastrophic forgetting on previous tasks. Models of associative memory have been used to investigate the discrepancy between biological and artificial neural networks due to their biological ties and inspirations, of which the Hopfield network is the most studied model. The Dense Associative Memory (DAM), or modern Hopfield network, generalizes the Hopfield network, allowing for greater capacities and prototype learning behaviors, while still retaining the associative memory structure. We give a substantial review of the sequential learning space with particular respect to the Hopfield network and associative memories. We perform foundational benchmarks of sequential learning in the DAM using various sequential learning techniques, and analyze the results of the sequential learning to demonstrate previously unseen transitions in the behavior of the DAM. This paper also discusses the departure from biological plausibility that may affect the utility of the DAM as a tool for studying biological neural networks. We present our findings, including the effectiveness of a range of state-of-the-art sequential learning methods when applied to the DAM, and use these methods to further the understanding of DAM properties and behaviors.
\end{abstract}

\section{Introduction}

Sequential learning is an important area of study in machine learning, where models are exposed to tasks in sequence and must learn new tasks without further exposure to previous ones. Biological neural networks can perform extremely well in sequential environments, incorporating information from previous tasks to help learn new tasks quicker (forward transfer) and achieve better-than-random performance on unseen tasks, (zero-shot performance). However, artificial neural networks struggle immensely in sequential environments, often forgetting previous tasks entirely (catastrophic forgetting) let alone showing any signs of forward transfer. There have been many techniques developed to improve the ability of artificial neural networks to learn sequential tasks, with varying degrees of success. 

Autoassociative memories have been of particular interest in the study of sequential learning due to their use in bridging the gap between computer science, psychology, human cognition, and neuroscience. In general, autoassociative memories are models that store patterns of activation by “associating” them with themselves (rather than with a different output pattern) using some learning rule that encodes the patterns in the model parameters. This process creates ``attractors'' in the state space of possible patterns, which allows the model to recall learned states when presented with similar probes by iterating the probe towards the attractor through some (usually physics-inspired) update mechanism. Sequential learning is performed in autoassociative memories by learning one set of states (a task) before learning another set of states without reference to the first. If the attractors of the second task disrupt those of the first, then the model has experienced catastrophic forgetting --- it is unable to recall earlier tasks. The Hopfield network \citep{Hopfield1982} is the most well studied model of autoassociative memory. Of particular note is its simplicity and the biological plausibility of the Hebbian learning algorithm \citep{Hebb1949}. The Hopfield network has also been the focus of ample research on sequential learning, again thanks to its parallels with biological networks. 

Recently, an abstraction on top of the classical Hopfield network has been proposed, the Dense Associative Memory \citep{KrotovHopfield2016,KrotovHopfield2018} (sometimes called the modern Hopfield network). In this abstraction, the quadratic attractors of the Hopfield network are replaced by attractors of any increasing function, allowing for steeper attractor basins and hence less cross-talk between learned items. A new hyperparameter, the interaction vertex, parametrizes the steepness of the attractors, leading to a natural question about how the interaction vertex influences sequential learning performance. The DAM has increased capacity, improved training times, and induces new learned representations compared to the classical Hopfield network. In exchange, the DAM loses some of the biologically plausible features of the classical network. The Hebbian learning rule, simple weight matrix, and single-step update rule are respectively replaced by gradient descent, set of memory vectors, and a contrastive difference of similarities. We discuss these differences further in Section \ref{Section: Literature Review Hopfield}. 

Unlike the Hopfield network, there is a distinct lack of research into sequential learning in the DAM. Although the changes between the Hopfield network and the DAM take a step away from biological plausibility, it is still worth studying in the context of sequential learning as an autoassociative memory. In particular, the unique architecture compared to feed-forward networks results in a range of new behaviors when using traditional sequential learning methods. Furthermore, the DAM is still a relatively new architecture, and its properties and characteristics are not yet fully understood.
 
In this paper we investigate sequential learning in the DAM. In Section \ref{Section: Literature Review} we conduct an extensive literature review of sequential learning in the Hopfield network. Section \ref{Section: Methods} details each of the sequential learning methods we will investigate, with much greater detail being found in Appendix \ref{Appendix: Methods}. Section \ref{Section: Experiments} discusses our experimental design and formalizes the DAM. In Section \ref{Section: Results} we analyze the results of our hyperparameter tunings for each sequential learning method, from which we infer several novel properties of the DAM, including:
 \begin{itemize}
    \item the need for increased data volumes to stabilize high interaction vertex DAMs,
    \item the variation of recalled items as the interaction vertex increases,
    \item and the existence of other behavior transitions beyond the known feature-to-prototype transition.
 \end{itemize}
 In Section \ref{Section: Results} we also give our results of the sequential learning performances in the DAM which lay a foundation for further work on sequential learning methods specifically designed for the model.
\section{Literature Review}
\label{Section: Literature Review}

\subsection{The Hopfield Network and Dense Associative Memory}
\label{Section: Literature Review Hopfield}

The classical Hopfield network is a relatively simple artificial neural network that operates as a model of autoassociative memory \citep{Hopfield1982}. Although earlier autoassociative memories exist \citep{Steinbuch1963, Steinbuch1965, Kohonen1972, Kohonen1978}, the Hopfield network is perhaps the most studied in the class. Typically, a network of dimension \(N\) operates over states that are drawn from a discrete domain, usually the binary (\(\{0,1\}^N\)) or bipolar (\(\{-1,1\}^N\)). The network can be operated over a continuous domain, but it has been shown that the behavior of the network is the same whether operating over continuous or discrete domains \citep{Hopfield1984}, and hence the simpler discrete domain is more common. The network usually employs the Hebbian learning rule \citep{Hebb1949}, or a variation of the Hebbian such as the Widrow-Hof \citep{WidrowHoff1960} or Storkey \citep{Storkey1997} learning rule. The classical Hopfield network has been analyzed extensively from the perspective of statistical physics, where the model is equivalent to that of a Sherrington-Kirkpatrick spin-glass system \citep{Amit1985A, Amit1985B, KirkpatrickSherrington1978, Bovier2001}. The long range spinglass is a mature physical model with well studied classes of states, phase transitions, and dynamics which can all be applied nearly directly to the Hopfield network, although statistical physics rarely looks at the equivalent of sequential learning. The Hebbian learning rule has been shown to give rise to a network capacity of only \(0.14 N\) states from both the computer science \citep{McEliece1987, Hertz1991} and physics perspectives \citep{Amit1985B, Amit1987}. This makes the classical Hopfield network difficult to work with, as the linear network capacity is made impractical by the quadratic weight matrix size \(N^2\). More relevant to our work, there has been significant analysis of the network in the field of psychology thanks to its biological plausibility \citep{Amit1994, Maass1997, Tsodyks1999, Hopfield1999}. Of note is the work on prototype formation in the Hopfield network, which may help explain prototype learning in the human brain \citep{Robins2004, McAlister2024A}.

The Dense Associative Memory abstracts the Hopfield network by allowing variation in the steepness of the energy wells forming the attractor space \citep{KrotovHopfield2016}. This is done by altering a hyperparameter known as the interaction vertex \(n \in \mathbb{R}^{+}\). It has been shown that the DAM has the same behavior as the classical network for \(n=2\) \citep{KrotovHopfield2016, Demircigil2017}, meaning the modern network could  replace the classical in all results. However, the modern network has carved its own distinct niche thanks to several other alterations. The shift from a Hebbian based learning rule to a gradient descent has removed a significant portion of the biological plausibility \citep{Stork1989}, making the network potentially less attractive in the study of psychology and neuroscience. This is compounded by a change from a weight matrix (which can describe the connection strength between a set of \(N\) neurons) to a set of memory vectors that are isolated from one another. However, as a model of autoassociative memory, the gains to the network capacity are hard to ignore. Specifically, Krotov and Hopfield find a network capacity of
\begin{align*}
    \frac{1}{2\left(2n-3\right)!!} \frac{N^{n-1}}{\ln(N)},
\end{align*}
meaning the network capacity is superlinear for \(n>2\). Furthermore, the network memories have been observed to have different representations between low and high interaction vertices. The memories appear to take on feature-like values at low interaction vertices, but transition to prototype-like values at high interaction vertices, indicating the network is learning to model the training data differently as the interaction vertex changes. This is entirely different to the prototype behavior observed in the classical Hopfield network, which emerges from properties of the Hebbian learning rule \citep{McAlister2024A}.

Despite the steps away from biological plausibility, the DAM has still been used in studies of psychology and neuroscience, with links made to the human brain \citep{Snow2024, Checiu2024}. The main focus of research around the DAM recently, however, has been on the correspondence to the attention mechanism used in transformers and Large Language Models \citep{Vaswani2017, HopfieldIsAllYouNeed2021}. This shows there is an interest in autoassociative memories and the possibility for biologically plausible explanations of network mechanics. Moreover, LLMs and attention based models often undergo a fine-tuning step in which the model learns on a small set of data corresponding to a new, specific task without retraining on the massive amount of general data seen previously. This is very similar to a sequential learning environment, and our work may have consequences for how fine-tuning is applied in the DAM equivalent attention mechanism.

\subsection{Sequential Learning in the Hopfield Network}

The classical Hopfield network has been studied in sequential learning environments at length. \citet{Nadal1986} analyzed the recall of the Hopfield network after training on a sequence of items. The weights of the network were rigorously analyzed as new items were added, and the magnitudes associated with each item were quantified with results that are strikingly similar to the spin glass literature \citep{Amit1985B, Toulouse1986,Personnaz1986}. Notably, the catastrophic forgetting observed when the network's capacity is exceeded is linked directly to an item's associated weight magnitude dropping below a derived threshold --- something that occurs nearly simultaneously for all stored items above the network capacity. 

\citet{Burgess1991} investigated list-learning in the Hopfield network, a form of sequential learning where items are presented one at a time for a single epoch, effectively forming tasks of a single item. Recall is tested over all previously presented items at the end of each epoch. It was found that the accuracy is high for both early and late items, but lower for intermediate items. This was linked to the biological memory model of short-term and long-term memory \citep{Shiffrin1969} as well as the concepts of recency and primacy/familiarity \citep{Lund1925}. \citeauthor{Burgess1991} also experimented with small modifications to the learning algorithms in the Hopfield network, such as increasing weights in proportion to the weight magnitude. Thanks to the simplicity of Hebbian learning rules, this is effectively equivalent to rehearsal in this domain. When using this method, it was found that the network could be tuned to exhibit strong primacy or strong recency behavior easily, and with some difficulty both behaviors could be observed. 

Robins investigated a more explicit form of rehearsal, looking at feed-forward neural networks with new tasks alongside items from previous tasks. Of particular interest are the variations on rehearsal studied. Sweep rehearsal \citep{Robins1993} involves rapidly updating the items rehearsed from some larger buffer of previous task items, which should provide a constant error signal for the new task every batch, while the rehearsal error signal is effectively smoothed out over many batches. Robins found this resulted in better sequential learning performance when using sweep rehearsal compared to traditional buffer rehearsal. However, it was also found that as the number of items became large the learned behavior was to ``pass-through'' items rather than learning a representation, limiting the usefulness of these techniques for sequential learning. Pseudorehearsal \citep{Robins1995} is another variation of rehearsal based techniques, where buffer items are not sampled from previous tasks but instead from the network itself. Pseudorehearsal is described in Section \ref{Section: Methods}, and discussed in depth in Appendix \ref{Appendix: Methods}, but effectively alleviates catastrophic forgetting without requiring access to previous task data by probing the network with random inputs and using the resulting output as a buffer item --- something that has been proposed to model the function of dreaming in memory consolidation in the human brain. Later Robins and McCallum moved from feed-forward networks to the Hopfield network \citep{Robins1998}. In particular, rehearsal based mechanisms were implemented in the Hopfield network, and for the first time sequential learning in a batched environment was studied in the Hopfield network (in contrast to list learning, such as in \citep{Burgess1991}). Using the thermal perceptron learning rule \citep{Frean1992} rather than the Hebbian learning rule (which, as discussed  above, implements sequential learning inherently), Robins and McCallum are able to push past the performance caps previously set by the low network capacity associated with Hebbian learning, and find that catastrophic forgetting still occurs but now in a similar fashion to feed-forward networks; later batches ``wipe out'' earlier ones. Robins and McCallum also showed that pseudorehearsal improved the sequential learning performance of the Hopfield network, particularly when the generated items are representative of the actual task items.

In contrast to the work done on sequential learning in the classical Hopfield network, there has been very little on sequential learning in the DAM. Indeed, it is still unknown how traditional sequential learning methods perform in combatting catastrophic forgetting, or how the interaction vertex effects forgetting. Our work aims to lay the foundations of this intersection between DAMs and sequential learning.

\section{Sequential Learning Methods}
\label{Section: Methods}

Sequential learning methods can be broadly classified into three categories: architectural, rehearsal, regularization. Architectural sequential learning methods make some change to the network architecture in an attempt to improve the sequential learning performance of the network. This may include freezing specific weights that are important to previous task, avoiding forgetting by preventing those weights from changing, or adding new modules to the network that are able to fit new tasks, minimizing disruption to previous tasks. Rehearsal-based sequential learning methods keep a buffer of items associated with previous tasks to replay during training of the next task. This ensures an error signal is still present from previous tasks and allows a model to weigh learning new data against forgetting old data. There is a natural tradeoff between the memory requirements of the buffer and the task representation of buffer items; while we would like to rehearse all of the data, it is often not feasible to store it. Finally, regularization-based methods introduce a surrogate term to the loss function that models the true loss of previous tasks. Regularization-based methods are often more memory efficient than rehearsal-based methods, storing only a weight importance rather than many task items, but requires a good approximation of the previous task's loss function to operate effectively. Moreover, most regularization-based techniques approximate the loss with a quadratic term; a good approximation near a minima of the previous task loss, but ignores the existence of other well-performing minima.

We investigate several well-established sequential learning methods in this paper as applied to the Dense Associative Memory. A comprehensive discussion of all methods can be found in Appendix \ref{Appendix: Methods}. Starting with rehearsal-based methods:

\begin{itemize}
    \item Naive Rehearsal: a rehearsal-based method in which a proportion of previous task items are stored in a buffer and presented for learning alongside new tasks.
    \item Pseudorehearsal \citep{Robins1995, Robins1998}: another rehearsal-based method, but rather than storing task items directly we first probe the network to recover \textit{some} attractor --- this may be a learned task item, but is not guaranteed to be one.
    \item Gradient Episodic Memories (GEM) \citep{Lopez-Paz2017}: a hybrid regularization- and rehearsal-based method. A proportion of task items are used to condition the gradient used to update the parameters of the model. Specifically, the gradient of the loss on a new task is projected such that the dot product with all previous task's gradients is non-negative, which in theory should prevent forgetting.
    \item Averaged Gradient Episodic Memories (A-GEM) \citep{Chaudhry2018B}: a hybrid regularization- and rehearsal-based method similar to GEM, but computes a single gradient over all previous tasks rather than considering all previous task gradients separately. This massively decreases the complexity of the sequential learning method, even offering a closed form solution for the projected gradient unlike GEM, at the cost of some theoretical soundness.
\end{itemize}

The other methods we investigate are all regularization-based, using a quadratic penalty in the loss as a surrogate for previous tasks. The main difference between each of these methods are the weight importance measures: how much is each parameter of the model constrained? A good sequential learning technique will strongly constrain parameters that are important to the previous task while allowing parameters that are unimportant to update freely. 

\begin{itemize}
    \item L2 Regularization: uses a uniform weight importance measure across model parameters, a naive solution that acts as a baseline for the regularization family of methods.
    \item Elastic Weight Consolidation \citep{Kirkpatrick2017}: uses the Fisher Information of the model on task data to determine weight importances. This is an expensive computation, but can be approximated using only first order gradients near a local minima of the loss function \citep{Pascanu2014}.
    \item Memory Aware Synapses \citep{Aljundi2018}: uses a Taylor expansion of the model to compute weight importances, taking the average magnitude of first-order derivatives evaluated at the task items to estimate how much each parameter contributes to the task.
    \item Synaptic Intelligence \citep{Zenke2017}: weight importance is estimated using a path integral of the task gradient over the path through parameter space walked during learning. The weight importances of each task are combined to give only a single quadratic loss term, rather than one for each task.
\end{itemize}
\section{Hyperparameter Tuning and Experiment Design}
\label{Section: Experiments}

\subsection{Dense Associative Memory Formalization and Hyperparameters}
\label{Section: Modern Hopfield Formalization}

The DAM is an autoassociative memory consisting of a set of memory vectors \(\bar{\zeta}\) that learns a set of items \(\bar{\xi}\) via a gradient descent on the difference between the initial probe item \(\xi \in \bar{\xi}\) and the relaxed item \(\xi^\prime\). In effect the learning algorithm checks if the learned states are stable, and if not, updates the memory vectors --- the weights of the DAM --- to increase the stability, minimizing the difference \(\lVert\xi-\xi^\prime\rVert\).

To relax a probe item, we update neurons until all neurons are stable, i.e. further updates do not result in change. In contrast to the classical Hopfield network \citep{Hopfield1984}, the DAM (particularly when used as a classifier) updates all neurons synchronously. The update for a specific neuron \(\xi_i\) is computed by:
\begin{equation}
    \label{Eqn: Modern Hopfield Relaxation}
    \begin{aligned}
        \xi_i\left(t+1\right) &:= \sigma\left( \sum_{\zeta \in \bar{\zeta}} \left( f_n ( \beta \zeta \cdot \xi_{+i}\left(t\right)) - f_n ( \beta \zeta \cdot \xi_{-i}\left(t\right) ) \right) \right),
    \end{aligned}
\end{equation}
where \(\sigma\) is an activation function, \(f_n\) is the interaction function parameterized by the interaction vertex \(n\), \(\beta\) is a scaling factor used to adjust the gradient during training, and \(\xi_{+i}, \xi_{-i}\) are the probe item \(\xi\) with the \(i^\text{th}\) neuron clamped on or off;
\begin{align*}
    \xi_{+i} &= \begin{cases}
        +1 \quad &\text{if } i=j \\ 
        \xi_{j} \quad &\text{if } i \neq j
    \end{cases}, \\
    \xi_{-i} &= \begin{cases}
        -1      \quad   &\text{if } i=j \\ 
        \xi_{j} \quad   &\text{if } i \neq j
    \end{cases}.
\end{align*}

When items are drawn from the bipolar domain, \(\xi \in \{-1,1\}^N\), we use \(\sigma = \text{sign}\) during relaxation, so items remain in the bipolar domain, and \(\sigma = \tanh\) during training so we can take the gradient across the function. We use a linear activation on the classification neurons so the results represent logits of the predicted probability distribution for an item. Throughout our experiments we have used the leaky rectified polynomial interaction function
\begin{equation}
    \label{Eqn: Leaky Rectified Polynomial Interaction Function}
    f_n(x) = \begin{cases}
        x^n             \quad   &\text{if } x>0 \\
        -\epsilon x     \quad   &\text{otherwise}
    \end{cases}
\end{equation}
with \(\epsilon = 10^{-2}\). Previous experiments show that the polynomial and rectified polynomial interaction functions \citep{KrotovHopfield2016} had very similar performances to the leaky rectified polynomial such that including them here would be redundant. We do not include results using more exotic interaction functions, such as the exponential \citep{Demircigil2017} or log-sum-exp \citep{HopfieldIsAllYouNeed2021}, as these make further constraints and assertions on the model architecture that may cloud our results and analysis. Further research should investigate sequential learning using these interaction functions, including how those results differ from our own.

\begin{algorithm}[H]
    \caption{Learning Process of the Dense Associative Memory} \label{Alg: Modern Hopfield Learning Algorithm}
    \SetAlgoLined
    \KwIn{A set of items \(\bar{\xi}\)}
    \KwResult{Trained memory vectors \(\bar{\zeta}\)}
    \KwData{Number of training epochs MaxEpochs, initial learning rate InitialLearningRate, learning rate decay LearningRateDecay, initial temperature \(T_{i}\), final temperature \(T_{f}\), momentum \(p\), error exponent \(m\)}
    
    Initialize \(\bar{\zeta} \gets \mathcal{N}\left(\mu=0, \sigma=0.1\right)\) \\
    Initialize MomentumGradient \(\gets 0\)\\
    \For{epoch \(\gets 1\) \KwTo MaxEpochs}{
        \(\text{lr} \gets \text{InitialLearningRate} \times \text{LearningRateDecay}^{\text{epoch}}\) \\
        \(T \gets T_i + \left(T_f - T_i\right) \times \frac{\text{epoch}}{\text{MaxEpochs}}\) \\
        \(\beta \gets \frac{1}{T}\) \\
        Error \(\gets 0\) \\
        \For{\(\xi \in \bar{\xi}\)}{
            \For{\(i \gets 1\) \KwTo \(N\)}{
                \(\xi_{+i} = \begin{cases}
                    +1 \quad &\text{if } i=j \\ 
                    \xi_{j} \quad &\text{if } i \neq j
                \end{cases}\) \\
                \(\xi_{-i} = \begin{cases}
                    -1      \quad   &\text{if } i=j \\ 
                    \xi_{j} \quad   &\text{if } i \neq j
                \end{cases}\) \\
                \(\xi_i^\prime \gets \tanh\left( \sum_{\zeta \in \bar{\zeta}} \left( f_n ( \beta \zeta \cdot \xi_{+i}) - f_n ( \beta \zeta \cdot \xi_{-i} ) \right) \right)\) \\
                Error \(:=\) Error \(+ \left(\xi_i - \xi_i^\prime\right)^{2 m}\)\\
            }
        }
        Compute EpochGradient \(\gets \nabla_{\bar{\zeta}}\) Error \\
        MomentumGradient \(\gets p \times \text{MomentumGradient} + \text{EpochGradient}\) \\
        \(\bar{\zeta} := \bar{\zeta} - \text{MomentumGradient} \times \text{lr}\)\\
    }    
\end{algorithm}
Algorithm \ref{Alg: Modern Hopfield Learning Algorithm} details the learning process of the DAM \citep{KrotovHopfield2016}, including the numerous hyperparameters that can greatly impact the behavior of the network. Previous research \citep{McAlister2024B} has shown a small modification to the definition of the network's relaxation function (namely, moving \(\beta\) inside the interaction function, which we have done in Equation \ref{Eqn: Modern Hopfield Relaxation}) results in very stable hyperparameter selection across a wide range of interaction vertices. These same modifications also allow for stable trainings for very high interaction vertices \(n\geq30\), but vertices this high are not investigated in this paper. Using these modifications, we are able to tune the network hyperparameters once on the permuted MNIST dataset and use those parameters again throughout the experiments ensuring consistency. See Appendix \ref{Appendix: General Hyperparameter Search} for the results of our grid-search over the initial learning rate and temperatures, which demonstrates the stability of hyperparameters across interaction vertex values. We used \(500\) training epochs, an initial learning rate of \(8\times10^{-2}\), a learning rate decay of \(0.999\), a momentum  of \(0.6\), an initial and final temperature value of \(0.95\), and an error exponent of \(1\) throughout our experiments. Krotov and Hopfield \citep{KrotovHopfield2016,KrotovHopfield2018} found that increasing the error exponent and temperature difference can improve performance at larger interaction vertices, but we found good performance without altering these values thanks to the modifications made to the algorithm. 

Algorithm \ref{Alg: Modern Hopfield Learning Algorithm} is modified slightly in practice --- implementing minibatches over the items and clamping memory weights to the interval \([-1,1]\) each update. The classification neuron weights are left unclamped, in consistency with other literature.

\subsection{Sequential Learning Methods Hyperparameters}
\label{Section: Method Hyperparameters}

Each sequential learning method in Section \ref{Section: Methods} has associated hyperparameters that must also be tuned to maximize performance. Instead of fixing the hyperparameters across all interaction vertices as we have done for the general hyperparameters of the DAM, we will tune the parameters per interaction vertex, as the optimal hyperparameter choice changed significantly with interaction vertex.

For rehearsal based methods, we varied the proportion of task items added to the rehearsal buffer. We conduct a linear search from a rehearsal proportion of \(0.0\) to \(1.0\). For naive rehearsal, this range corresponds to vanilla sequential learning (no sequential learning method) at \(0.0\) to presenting all tasks non-sequentially at \(1.0\). For pseudorehearsal, the lower end of the search again corresponds to no method, but the upper end no longer has a nice interpretation. We can generate as many buffer items as we like, including more than the number of items in the original task. Since we do not require unique generated items, we are likely to generate several repeated items in the buffer, but as the generated items are representative of the network's attractor space, this is not only acceptable, but perhaps preferable to ensure important regions are well sampled. We do not explore beyond a pseudorehearsal proportion of \(1.0\) as we see a plateau in the performance. Gradient Episodic Memories and Averaged Gradient Episodic Memories also use a memory buffer, and we again measure the amount of data added to the buffer as a proportion of the task size. This allows us to compare the rehearsal methods with the GEM family of methods. 

Regularization based methods --- L2 Regularization, Elastic Weight Consolidation, Memory Aware Synapses, and Synaptic Intelligence --- all have the general form of the loss function:
\begin{equation}
    \mathcal{L}\left(\theta\right) = \mathcal{L}_{\text{Base}} \left(\theta\right) + \lambda \sum_{k} \omega_{k} \left(\theta_{k}^\star - \theta_k\right)^2,
\end{equation}
for model parameters \(\theta\), \(k\) indexing over the model parameters, base loss \(\mathcal{L}_{\text{Base}}\), weight importances \(\omega\), and optimal parameters on the previous task \(\theta^\star\). We must tune the regularization parameter \(\lambda\), which balances the base loss (learning the new task) and the quadratic penalty (remembering the old tasks). Unlike regularization-based methods, in which we could guess that an increased memory proportion would improve sequential learning performance, the regularization hyperparameter \(\lambda\) has a more nuanced impact on performance --- too small and the network will forget earlier tasks, too large and the network will not learn new tasks. Furthermore, the search space of \(\lambda\) is not bounded, and it is not clear where the optimal region will lie. For these reasons, we have not conducted a grid search over \(\lambda\) but instead use Optuna \citep{Optuna2019}, a hyperparameter optimization software library, in conjunction with a Tree-structured Parzen Estimator sampler over the search space. This sampler attempts to estimate the optimal hyperparameter value using a Gaussian Mixture Model, resulting in dense sampling near the optimal region (of interest to us) and sparser sampling far from the optimal region. To estimate the error in the average accuracy across the \(\lambda\) search space, we have used a moving window of size \(20\) and plot the average and standard deviation (as an error band) of the window. This gives us better performance and error estimates close to the optimal region, which is more useful for our analysis. In each of the regularization-based methods we have conducted \(100\) trials for each interaction vertex.

\subsection{Experimental Design}

In our experiments, we consistently use a series of five permuted MNIST tasks. The MNIST dataset \citep{LeCun1998} consists of images of handwritten digits, lending itself to classification tasks. Permuted MNIST \citep{Kirkpatrick2017} creates separate sequential learning tasks from MNIST by applying a permutation of all pixels in each image. Each task has a separate, consistent permutation. Since the models we are studying do not rely on the spatial relationships of the data (compared to, for example, convolutional neural network) the permuted task is no harder than the unpermuted task. By permuting all pixels in the image we remove any common features that the model may use to transfer learning between tasks --- we say the tasks are unrelated. By permuting only some pixels, applying or applying a different transformation, we could create tasks that are more related, which is more typical of practical sequential learning problems. However, we use only permuted MNIST in this work as we are interested in laying the foundation of sequential learning in the DAM rather than investigating other effects such as zero shot performance. Appendix \ref{Appendix: Datasets} has further discussion of sequential learning datasets. Studies on other datasets, including continuous ones, are determined to be outside the scope of this paper, but may be of interest to further research.

For hyperparameter tuning we use tasks consisting of \(2000\) items, and for the final results we use \(10000\) items per task. Items are randomly sampled from the full MNIST dataset. We do not train on the full MNIST dataset in each task due to computational constraints. We use only five tasks due to instability of the network with additional training --- we found the network weights fluctuate increasingly wildly as the number of tasks is increased (see Appendix \ref{Appendix: Appendix Task Performances} for examples of this). Decreasing the learning rate and / or increasing the momentum helped reduce this unwanted behavior, but at the cost of sequential learning performance. MNIST is not an overly challenging dataset by contemporary standards, but we find that even this dataset provides insights into the behavior of the Dense Associative Memory, and still differentiates between the various sequential learning techniques we investigate. 

Each task has \(20\%\) of its data set aside for testing, with the remaining \(80\%\) used for training. Each task has a different randomly generated permutation of pixels. Task items are encoded as binary vectors; starting with the flattened, permuted pixel data, followed by a one-hot encoded task ID, and ten neurons to represent the digit classes. To conform to the literature that uses this dataset with the DAM \citep{KrotovHopfield2016,KrotovHopfield2018}, we update only the ten neurons corresponding to the class of the probe item, and update them only once. We do not use the feed-forward equivalent network in this literature, keeping the autoassociative memory architecture throughout our experiments. 

We present the tasks to a DAM with \(512\) memory vectors for \(500\) training epochs each. Using a greater number of memory vectors or epochs did not improve network performance significantly. The training and relaxation procedures are discussed in Section \ref{Section: Modern Hopfield Formalization}. After training, we present the network with items from the test dataset, which we relax and compute the test F1 score. The two measures we use throughout our experiments are individual task accuracy and average accuracy on all tasks \citep{Caccia2021}. The average accuracy of a series of sequential learning tasks is defined as the accuracy of all tasks seen so far \(\nu \leq \mu\) after training on task \(\mu\).  To ensure our random sampling of data is not biasing our results, we use average F1 score in place of accuracy, so any class imbalances are accounted for. However, we refer to F1 score as accuracy throughout our discussions for flow, and the metrics are effectively the same for such large sample sizes.

To tune the general network hyperparameters (see Section \ref{Section: Modern Hopfield Formalization}), we train a DAM on five tasks and measure the test accuracy of each task at the end of the respective training period. A grid search over network hyperparameters is performed, with the objective to maximize the minimum test accuracy of any task --- that is, we are seeking a network that can learn tasks as accurately as possible even after training on other tasks. This ensures that the network learns quickly enough to perform well on the first task (overcoming initially random memory vectors) but is also plastic enough to learn subsequent tasks. Note that at this stage we are not tuning for the optimal \textit{sequential learning} results, only for \textit{individual task} results, so that the network learns quickly and remains plastic. It is likely that previous tasks are forgotten quickly, however new tasks are learned to near completion and are generalized well.

To tune sequential learning method hyperparameters (Section \ref{Section: Method Hyperparameters}), we again train a DAM on five sequential tasks but now measure the average accuracy at the end of all tasks. Now, instead of tuning the hyperparameters to make a network as plastic as possible, we are tuning parameters to ensure previous tasks are not forgotten, while new tasks are still picked up. This combination of parameter tunings mirrors what we might expect in a practical scenario; starting with a model that is tuned to learn as generally as possible, we apply sequential learning techniques that are tuned to improve sequential learning performance.

\section{Experiment Results}
\label{Section: Results}

\subsection{Sequential Learning Methods Hyperparameter Tuning}

We present the average accuracy of the DAM on five permuted MNIST tasks when trained with different sequential learning methods. For a baseline average accuracy with no sequential learning method (i.e. vanilla learning) see Table \ref{Table: Aggregated Results} later in this Section. Plots of the individual task accuracy for each sequential learning method can be found in Appendix \ref{Appendix: Appendix Task Performances}.

\subsubsection{Rehearsal Methods}

\begin{figure}[H]
    \centering
    \begin{subfigure}[t]{\figureWidth\textwidth}
        \centering
        \includegraphics[width=\textwidth]{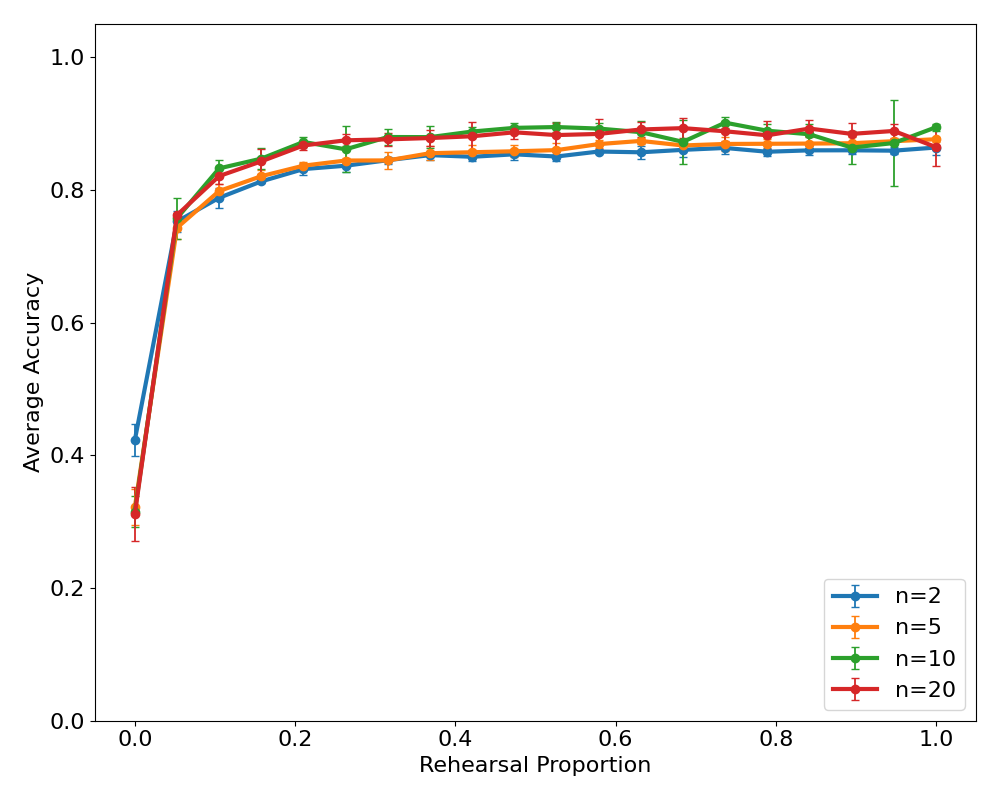}
        \caption{Naive Rehearsal.}
        \label{Fig: Rehearsal Hyperparameter Search}
    \end{subfigure}
    \begin{subfigure}[t]{\figureWidth\textwidth}
        \centering
        \includegraphics[width=\textwidth]{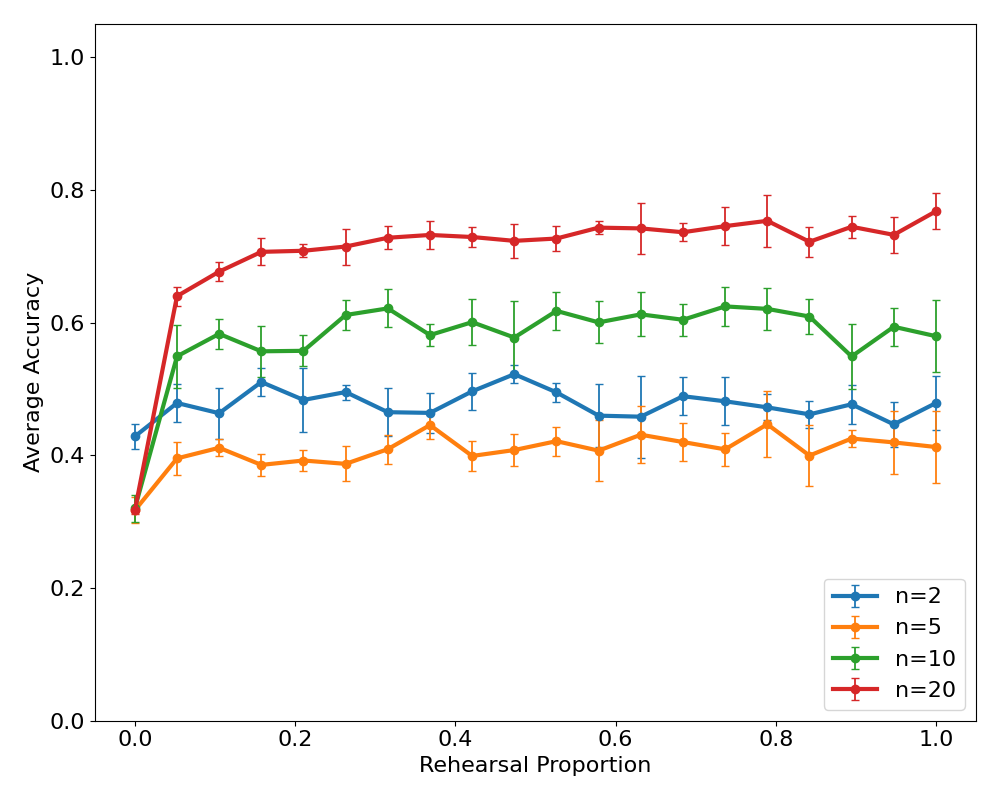}
        \caption{Pseudorehearsal.}
        \label{Fig: Pseudorehearsal Hyperparameter Search}
    \end{subfigure}
    \caption{Hyperparameter search over rehearsal proportion, measuring the average accuracy on the test data split. A rehearsal proportion of \(0.0\) corresponds to vanilla learning. For naive rehearsal, a proportion of \(1.0\) corresponds to presenting all previous tasks alongside the new task. Different interaction vertices, \(n\), are shown by color. A higher average accuracy reflects better performance on sequential learning tasks.}
\end{figure}

Our hyperparameter search for naive rehearsal, Figure \ref{Fig: Rehearsal Hyperparameter Search}, gives the expected response curve with respect to the rehearsal proportion. As more items are added to the buffer, the average accuracy increases, as previous tasks are rehearsed more thoroughly. This sets a good benchmark for future sequential learning methods, as no method should be able to outperform the simple solution of presenting all tasks for training at once. Perhaps most interestingly, and unique to naive rehearsal, all interaction vertices appear to perform effectively equally well at non-trivial rehearsal proportions. This indicates that five Permuted MNIST tasks are not overly difficult for this network with a number of memories \(|\bar{\zeta}|=512\), no matter the interaction vertex. For a rehearsal proportion of zero, \(n=2\) seems to outperform \(n>2\), which we attribute to higher interaction vertices learning new tasks, and forgetting old tasks, more quickly (see Appendix \ref{Appendix: Appendix Task Performances}).

Pseudorehearsal, Figure \ref{Fig: Pseudorehearsal Hyperparameter Search} exhibits more interesting response curves than naive rehearsal. Low interaction vertices, \(n=2\) and \(n=5\), have a reasonably flat response to the rehearsal proportion, while high interaction vertices, \(n=10\) and \(n=20\) have their average accuracy improve considerably with the rehearsal proportion. We conjecture that the Dense Associative Memory with a high interaction vertex recalls pseudoitems that are extremely representative of previously tasks, while low interaction vertex networks do not recall such pseudoitems. This can be attributed to the feature-to-prototype transition described by \citet{KrotovHopfield2016} --- the memory vectors of the network encode different representations which allow different items to be recalled, of which only some are useful for pseudorehearsal.  While we do not reach naive rehearsal levels of performance, we do observe very good sequential learning performance for even modest rehearsal proportions. These results also suggest that only a very small number of pseudoitems are required to saturate the sequential performance of the DAM under pseudorehearsal, although higher interaction vertices require more items than lower interaction vertices.

\begin{figure}[H]
    \centering
    \begin{subfigure}[t]{\figureWidth\textwidth}
        \centering
        \includegraphics[width=\textwidth]{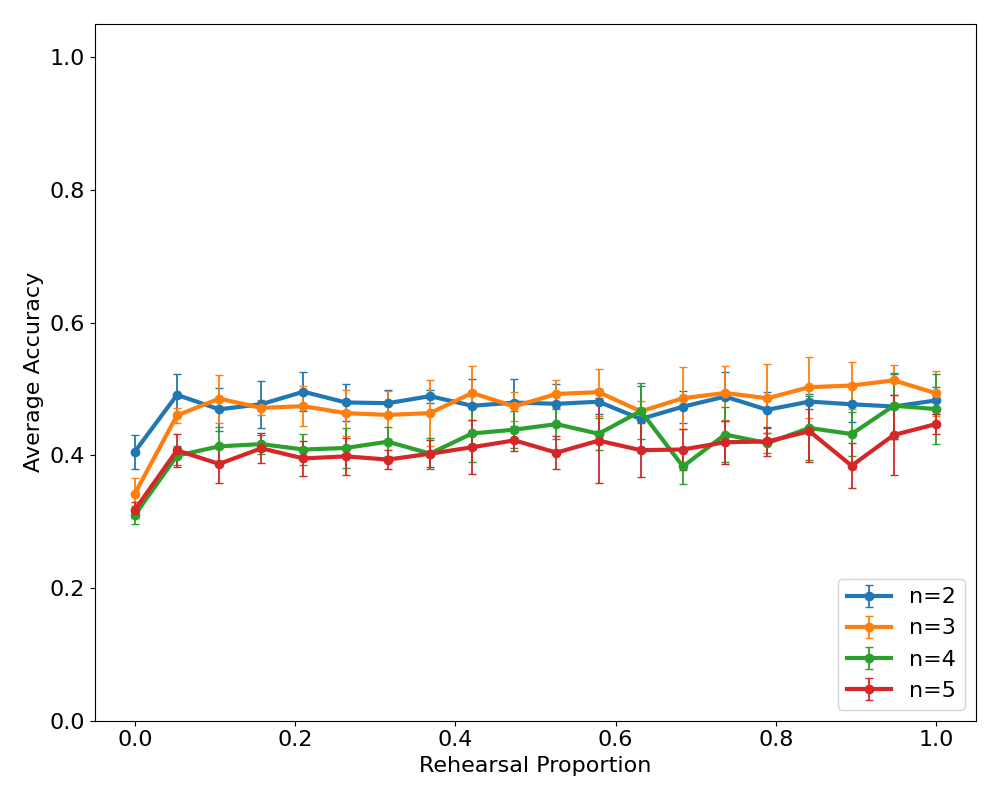}
        \caption{Small interaction vertices.}
        \label{Fig: Pseudorehearsal Small n}
    \end{subfigure}
    \begin{subfigure}[t]{\figureWidth\textwidth}
        \centering
        \includegraphics[width=\textwidth]{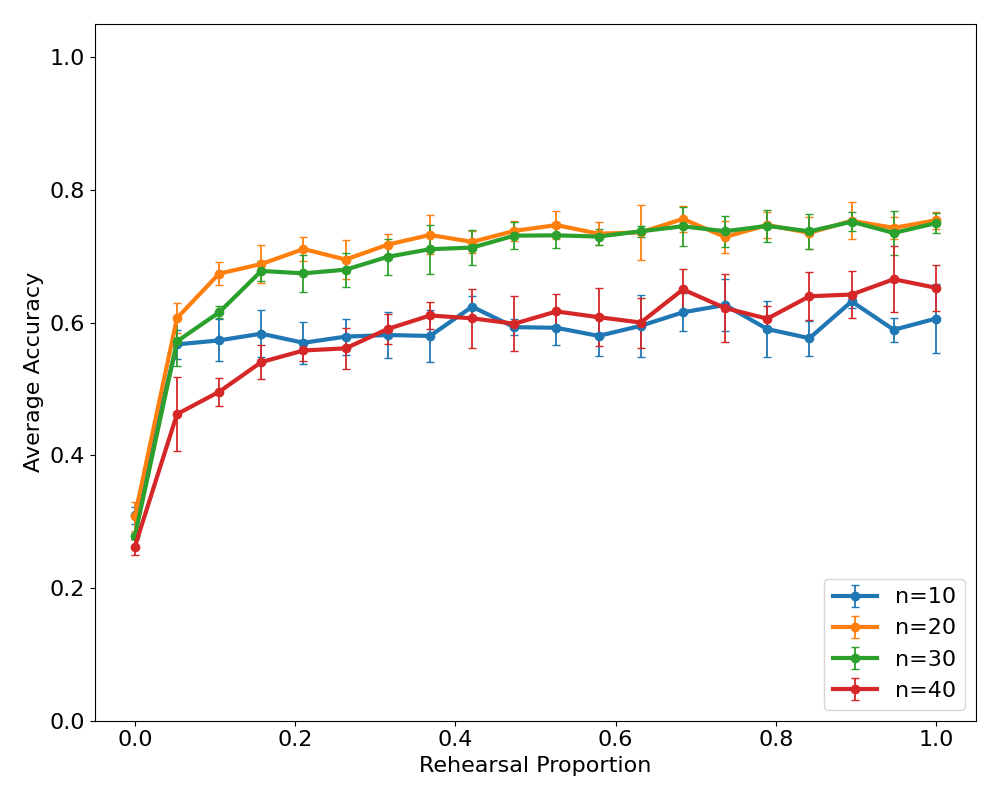}
        \caption{Large interaction vertices.}
        \label{Fig: Pseudorehearsal Large n}
    \end{subfigure}
    \caption{Pseudorehearsal hyperparameter search over rehearsal proportion,  measuring the average accuracy on the test data split. Note the legend in these Figures is different from others in this Section. These figures explore finer granularity at low and high interaction vertices.}
\end{figure}

Since pseudorehearsal offers insight into the recall behavior of the Dense Associative Memory, we have extended our investigation to other interaction vertices. We observe a behavior transition in the network for low interaction vertices --- notably, at a different point than the well known feature-to-prototype transition. In Figure \ref{Fig: Pseudorehearsal Small n} we see \(n\leq3\) has higher average accuracy than \(n=4,5\) (not considerably higher, but significantly higher). This may indicate that pseudoitems generated at \(n\leq3\) are more useful in preserving previous tasks compared to \(n=4,5\), which would infer a behavior transition of the DAM that has not yet been studied. However, unlike the performance differences in Figure \ref{Fig: Pseudorehearsal Hyperparameter Search}, which we attribute to the visually striking feature-to-prototype transition of memory structure, the cause of the performance difference in Figure \ref{Fig: Pseudorehearsal Small n} is less obvious. The pseudoitems generated from each network in the Figure are extremely similar, and the memory vectors have not obviously changed on inspection. The performance difference may also be due to a change in how the network responds to the pseudoitems, rather than a change to the pseudoitems themselves; that is, we are observing a transition in the learning behavior of the DAM. Further research is required to understand the exact nature of this discrepancy, although its existence is enough to provoke interest in the model's additional, unstudied modes.

We also study pseudorehearsal for larger interaction vertices, up to \(n=40\), Figure \ref{Fig: Pseudorehearsal Large n}.  We find that the sequential learning performance plateaus for \(n=20,30\) and drops again at \(n=40\). This could indicate that large interaction vertices undergo yet another behavior change, although we could also be observing a drop in the stability of the network, leading to worse performance due to unstable training. Typically, high interaction vertex networks are more temperamental, even with the modifications made to improve stability \citep{McAlister2024B}.

\subsubsection{GEM and A-GEM}

\begin{figure}[H]
    \centering
    \begin{subfigure}[t]{\figureWidth\textwidth}
        \centering
        \includegraphics[width=\textwidth]{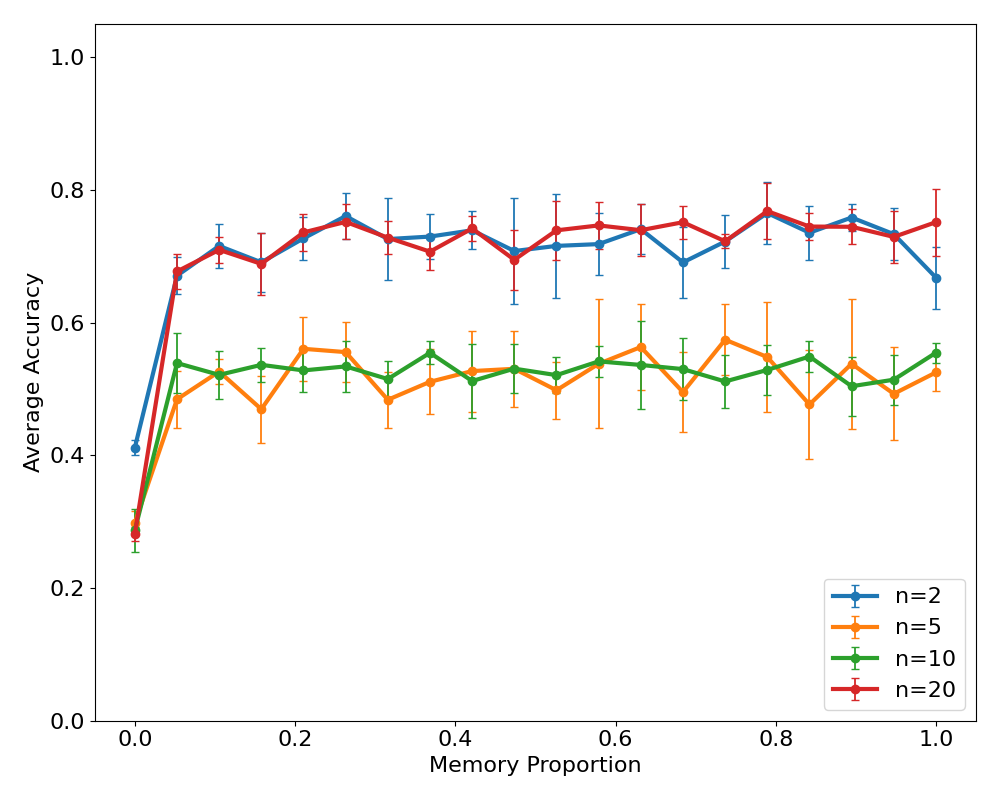}
        \caption{Gradient Episodic Memory.}
        \label{Fig: GEM Hyperparameter Search}
    \end{subfigure}
    \begin{subfigure}[t]{\figureWidth\textwidth}
        \centering
        \includegraphics[width=\textwidth]{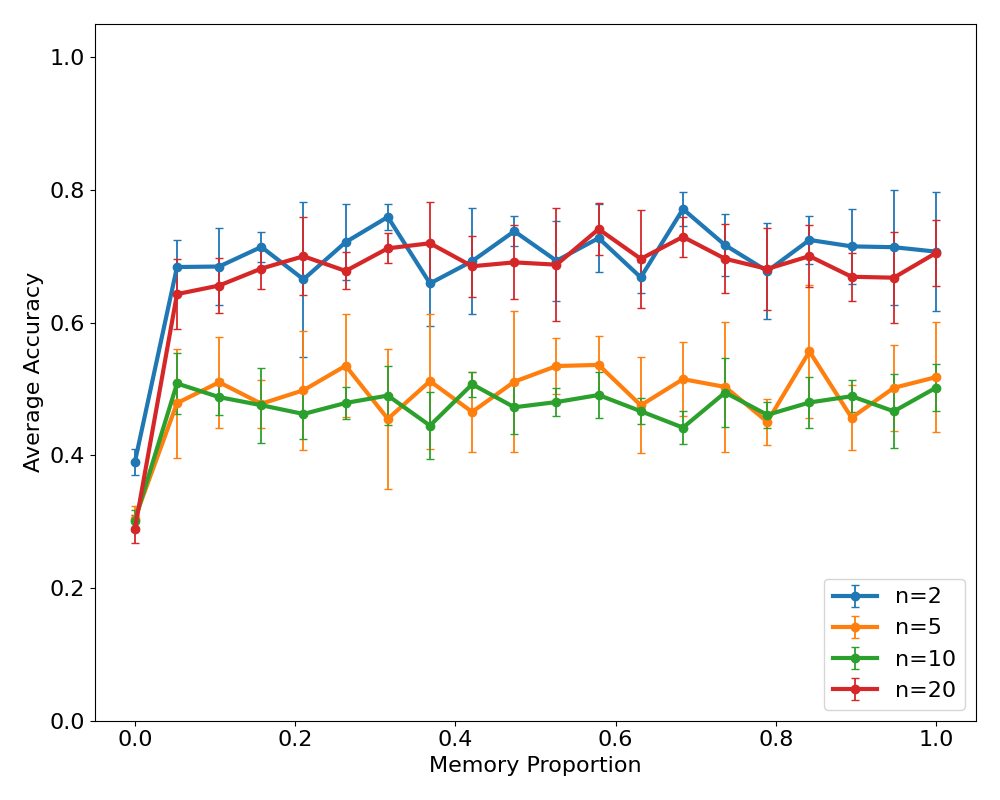}
        \caption{Averaged Gradient Episodic Memory.}
        \label{Fig: A-GEM Hyperparameter Search}
    \end{subfigure}
    \caption{Gradient Episodic Memories hyperparameter search, measuring the average accuracy on the test data split. A memory proportion of \(0.0\) corresponds to vanilla learning, while \(1.0\) checks the gradient across  all previous task items. A higher average accuracy reflects better performance on sequential learning tasks.}
\end{figure}

For both GEM and A-GEM (Figure \ref{Fig: GEM Hyperparameter Search} and \ref{Fig: A-GEM Hyperparameter Search} respectively), we see high average accuracy for \(n=2, 20\) and low average accuracy for \(n=5,10\), although all interaction vertices see a marked improvement over the baseline (a memory proportion of \(0\)). GEM and A-GEM appear to work well only for the extreme interaction vertices. It is possible that the memory vectors of intermediate interaction vertices may be somehow incompatible with the gradient constraints introduced by these methods. The standard deviations in both methods are much larger than in naive rehearsal or pseudorehearsal, indicating that the DAM is more sensitive to gradient constraints across all interaction vertices --- training is much more unstable in these methods than rehearsal-based methods. Comparing GEM and A-GEM we see there is little difference in performance, corroborating \citet{Chaudhry2019} who found that A-GEM, despite relaxing constraints on the theory of GEM, performed equally well. Between rehearsal-based and gradient-based methods, looking now at the memory proportion required for each, naive rehearsal unsurprisingly has better performance across all memory proportions, but pseudorehearsal, GEM, and A-GEM have similar performances for large interaction vertices. For very small interaction vertices, the same memory proportions give better performances when used with gradient-based methods, although none of pseudorehearsal, GEM, or A-GEM work particularly well for the intermediate interaction vertices.

\subsubsection{Regularization-Based Methods}
\label{Section: Regularization Method Parameters}

\begin{figure}[H]
    \centering
    \begin{subfigure}[t]{\figureWidth\textwidth}
        \centering
        \includegraphics[width=\textwidth]{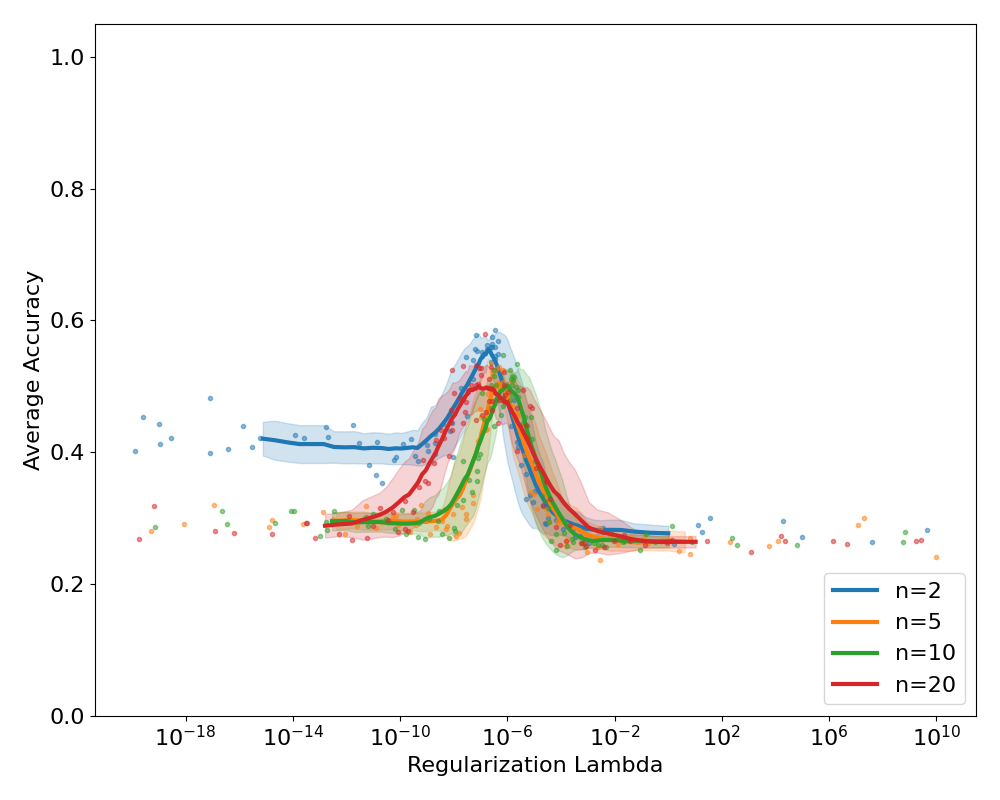}
        \caption{L2 Regularization.}
        \label{Fig: L2 Hyperparameter Search}
    \end{subfigure}
    \begin{subfigure}[t]{\figureWidth\textwidth}
        \centering
        \includegraphics[width=\textwidth]{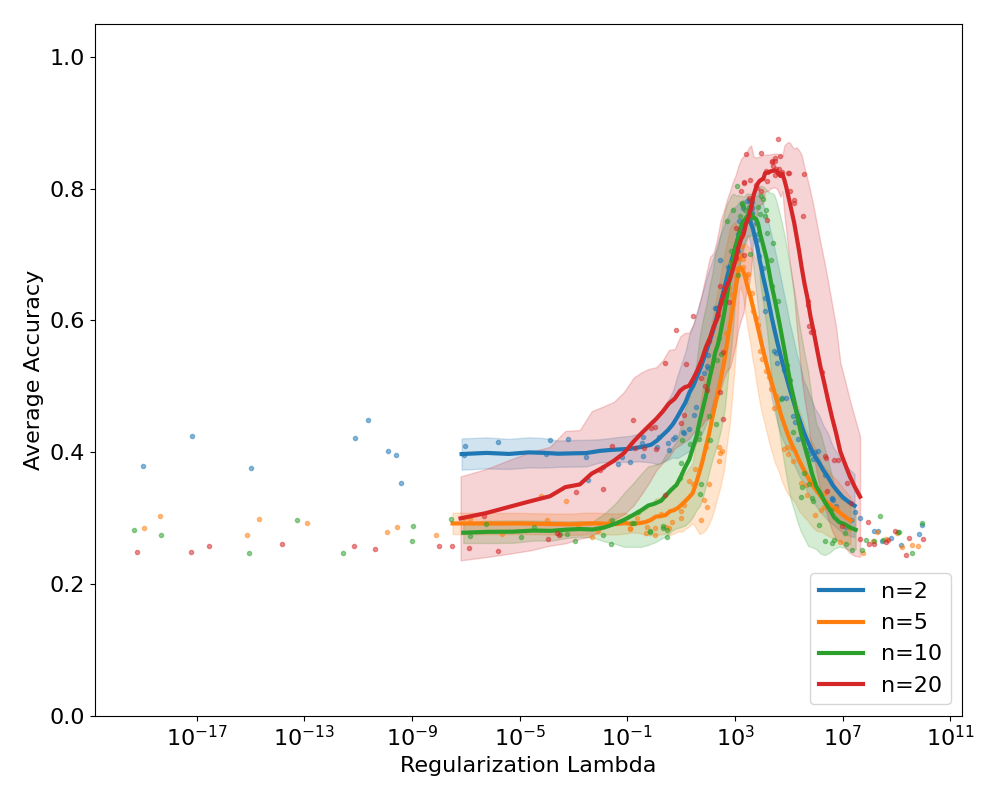}
        \caption{Elastic Weight Consolidation.}
        \label{Fig: EWC Small Data Hyperparameter Search}
    \end{subfigure}
    \hfill  
    \begin{subfigure}[t]{\figureWidth\textwidth}
        \centering
        \includegraphics[width=\textwidth]{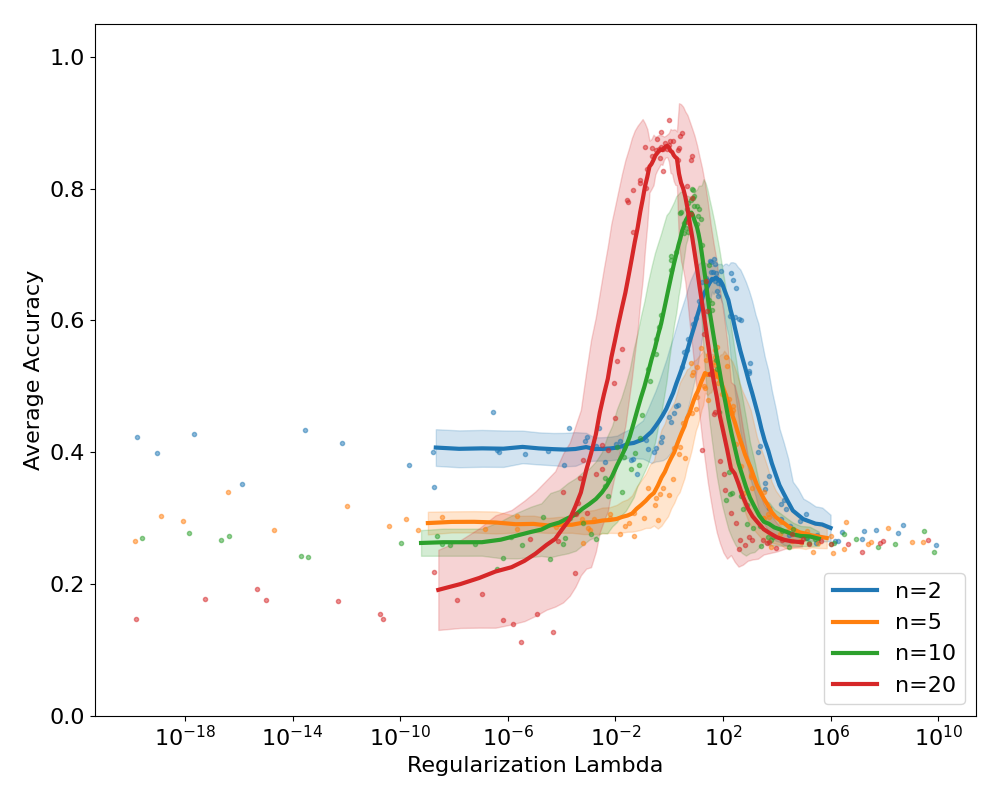}
        \caption{Memory Aware Synapses.}
        \label{Fig: MAS Hyperparameter Search}
    \end{subfigure}
    \begin{subfigure}[t]{\figureWidth\textwidth}
        \centering
        \includegraphics[width=\textwidth]{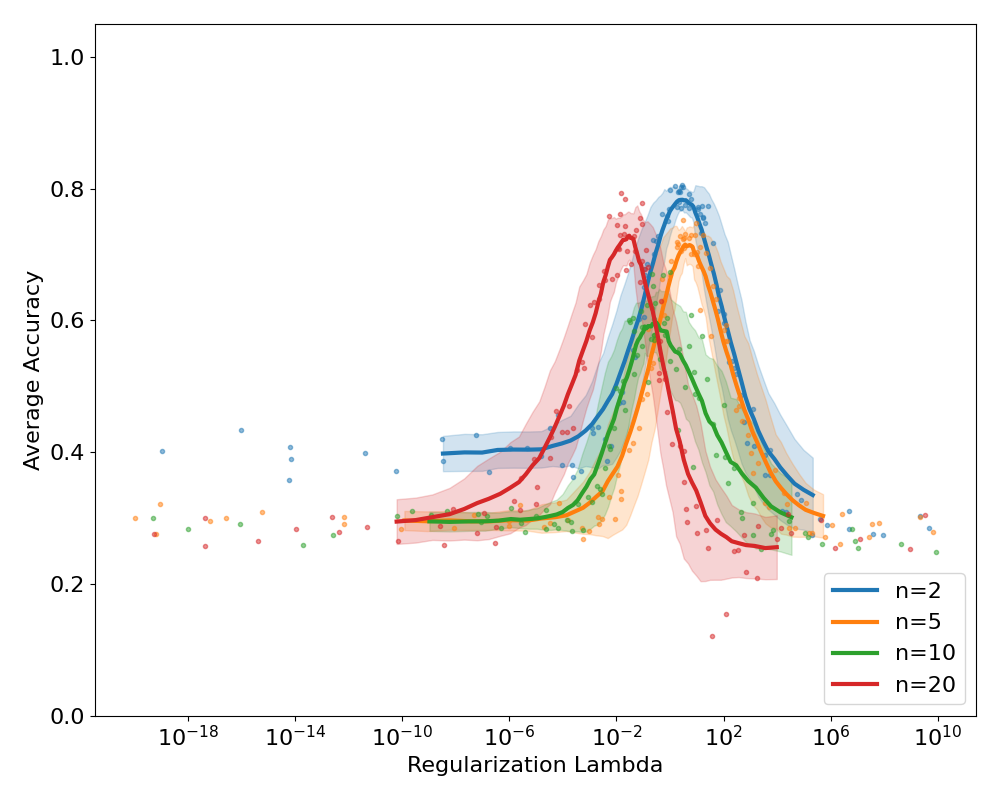}
        \caption{Synaptic Intelligence.}
        \label{Fig: SI Hyperparameter Search}
    \end{subfigure}
    
    \caption{Hyperparameter searches for regularization-based sequential learning methods over the regularization hyperparameter \(\lambda\), measuring the average accuracy on the test data split. Individual trials are shown as points, and a moving window of \(20\) trials is used to calculate the average (solid lines) and standard deviation (error band). A higher average accuracy reflects better performance on sequential learning tasks.}
    \label{Fig: Regularization Methods Hyperparameter Search}
\end{figure}

In Figure \ref{Fig: Regularization Methods Hyperparameter Search}, across all methods and all interaction vertices we see a definite peak in average accuracy with tails on either side. Our intuition is realized in the data: very high regularization \(\lambda\) constrain weights near the first task's optimal parameters but fail to learn new tasks, and very low regularization \(\lambda\) allow the network to forget each task and only remember the most recent. A sweet-spot between these extremes has the DAM remember each task and corresponds to the peaks in average accuracy. Across all methods we find that \(n=2\) gives a unique response curve compared to higher interaction vertices; lower-than-optimal values of \(\lambda\) perform much better than higher-than-optimal values. We can attribute this difference to the ``faster'' learning at higher interaction vertices \citep{KrotovHopfield2016}, which would cause the \(n=2\) networks to forget previous tasks ``slower'' even when under-constrained (again, see Appendix \ref{Appendix: Appendix Task Performances}). For higher interaction vertices, the response is often much more symmetrical, showing the improved remembering of previous tasks disappears.

L2 regularization, Figure \ref{Fig: L2 Hyperparameter Search}, acts as our baseline for the regularization-based methods. Unlike some other regularization methods, L2 regularization has \(n=2\) as the best performing interaction vertex, counter to what we found with rehearsal-based methods. L2 regularization also has a relatively low peak average accuracy, which is to be expected for a crude weight importance measure that is not based on any measurement of the network itself. The optimal regularization parameter (the peak of the response curve) shifts slightly as the interaction vertex increases. Unique to L2 regularization, we can interpret this shift directly, as the weight importances are identical and constant. Higher regularization hyperparameters indicate that the network requires stronger constraints on previously learned memories to avoid forgetting. Lower regularization parameters indicate that these constraints can be more relaxed and still retain that performance. In Figure \ref{Fig: L2 Hyperparameter Search} \(n=2,20\) both have slightly smaller optimal \(\lambda\) than \(n=5,10\). Alongside our results for GEM and A-GEM, this indicates that the memory vectors of the intermediate interaction vertices behave very differently to the feature- or prototype-regime memory vectors. In this case, it seems that intermediate interaction vertex DAMs require more constraints to avoid forgetting. This could mean that intermediate interaction vertex DAMs are fundamentally worse at representing these tasks, or perhaps that conventional sequential learning methods are not as applicable to intermediate interaction vertex networks.  The mechanism behind this is not clear, but hints at more nuanced behaviors in the DAM.

\begin{figure}[H]
    \centering
    \includegraphics[width=0.65\textwidth]{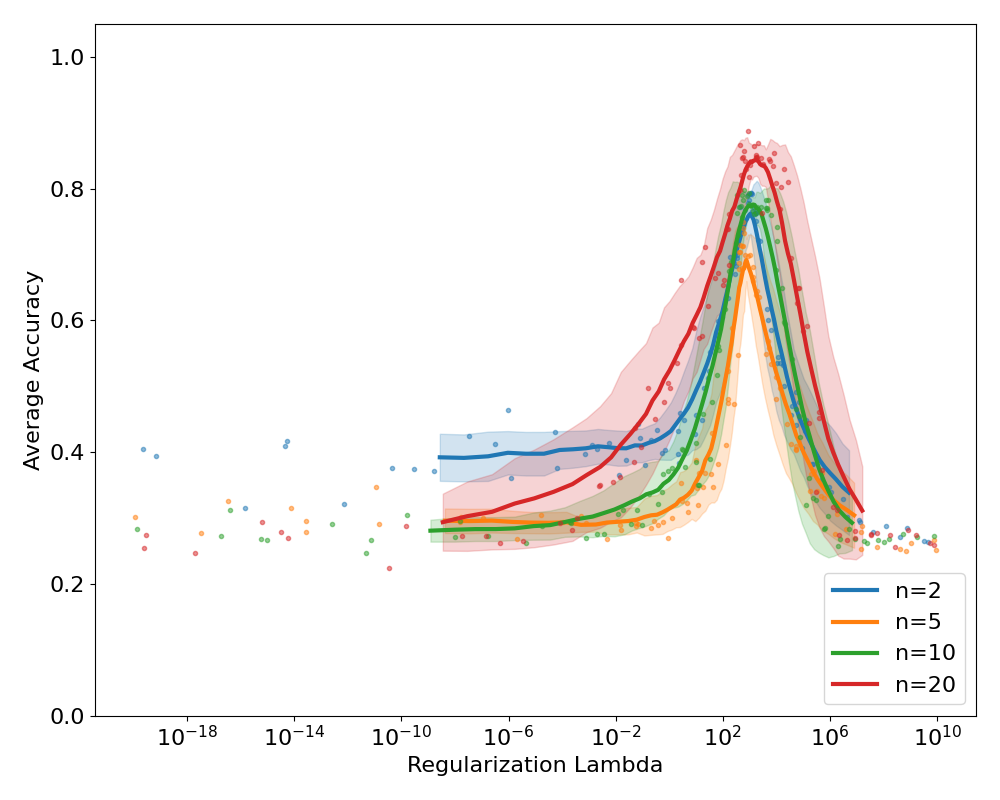}
    \caption{Elastic Weight Consolidation hyperparameter search using \(10000\) items per task. Compare this to Figure \ref{Fig: EWC Small Data Hyperparameter Search}, which uses \(2000\) items per task.}
    \label{Fig: EWC Large Data Hyperparameter Search}
\end{figure}

Elastic Weight Consolidation, Figure \ref{Fig: EWC Small Data Hyperparameter Search} and Figure \ref{Fig: EWC Large Data Hyperparameter Search}, presented some serious issues in our testing. Unfortunately, unlike all other regularization methods, we found a significant disparity in the performance of EWC when using the regularization lambda tuned for ``small'' tasks (\(2000\) items per task, Figure \ref{Fig: EWC Small Data Hyperparameter Search}) to train ``large'' tasks (\(10000\) items per task). We have retuned \(\lambda\) using \(10000\) items per task, shown in Figure \ref{Fig: EWC Large Data Hyperparameter Search}. Transferring hyperparameters from less computationally expensive tunings is a vital part of deep learning, as tuning on an entire dataset is often intractable. This is a mark against using Elastic Weight Consolidation with the DAM, especially since the most notable change between the two tunings occurs at higher interaction vertices, where the DAM is most interesting. Interpreting the differences between Figure  \ref{Fig: EWC Small Data Hyperparameter Search} and Figure \ref{Fig: EWC Large Data Hyperparameter Search}, the peak for \(n=20\) has shifted considerably which could indicate that high interaction vertex networks require more data to stabilize the memory vectors in a way that the Fisher Information measures. That is to say, higher interaction vertex DAMs are more data-hungry than lower ones. For a fair comparison, we tested the larger data tunings for the other regularization-based methods but did not find a significant disparity.

Looking now only at Figure \ref{Fig: EWC Large Data Hyperparameter Search}, the peak performance of EWC increases with the interaction vertex, reaching near naive rehearsal performance for \(n=20\). This may indicate that the weight importances are more concentrated in only a few memory vectors, resulting in only some weights being constrained, allowing strong constrains (approaching weight freezing) without disrupting plasticity in uncommitted memory vectors. We expect the Fisher Information to approximate the importance of network memories much better than a fixed importance measure, and therefore allow more nuanced constraining of weights than a fixed strategy. This would also explain the improved performance of EWC compared to L2 regularization. Aside from the aforementioned shifts relating to the task sizes, there is considerably less optimal hyperparameter shifting between the interaction vertices --- that is, the peaks are aligned at a specific \(\lambda\) value, which may allow practical applications to forgo retuning \(\lambda\) for each new network when using Elastic Weight Consolidation.

Memory Aware Synapses, Figure \ref{Fig: MAS Hyperparameter Search}, displays different behavior again compared to Elastic Weight Consolidation and L2 regularization. In Memory Aware Synapses not only do we find the common asymmetrical response curve for \(n=2\) but also an asymmetrical response for \(n=20\), where lower-than-optimal \(\lambda\) values produce terrifically poor average accuracies. As a tradeoff for the sensitivity, Memory Aware Synapses with a high interaction vertex offers some of the best accuracy we have observed, rivaling even naive rehearsal. There is a consistent shift to smaller optimal regularization parameters as the interaction vertex increases, similar to the shift towards higher regularization parameter values seen for Elastic Weight Consolidation when using small task sizes (Figure \ref{Fig: EWC Small Data Hyperparameter Search}). However, retuning Memory Aware Synapses with larger task sizes did not change the shifts as it did with Elastic Weight Consolidation. Perhaps this indicates that the computed importances are of an increasing magnitude, or that the ``concentration'' of weight importance increases to give rise to a consistent left-ward shift in \(\lambda\) as the interaction vertex increases. Memory Aware Synapses uses a Taylor expansion across the network outputs to compute the weight importances; for high interaction vertices we would expect the gradients in this expansion to increase due to the larger exponent, increasing the magnitude of the weight importance measures, offering a potential mechanism for the shift towards smaller regularization parameters.

Synaptic Intelligence, Figure \ref{Fig: SI Hyperparameter Search}, displays some decidedly different behavior compared to the above methods. The lowest interaction vertex network now performs better than other networks (similar to L2 Regularization), although the \(n=20\) network approaches the \(n=2\) in terms of performance (similar to GEM and A-GEM). This may suggest that Synaptic Intelligence interacts particularly well with feature-regime memories, unlike Elastic Weight Consolidation and Memory Aware Synapses.  As in Memory Aware Synapses the optimal regularization parameter decreases with the interaction vertex. Synaptic Intelligence includes the total weight drift over a task, which could be drastically changed by the interaction vertex. For example, altering a feature-regime memory on one task to a feature-regime memory on a second task may result in a smaller drift than altering a prototype-regime memory on one task to a prototype-regime memory on another, even for tasks that are extremely unrelated such as permuted MNIST. It would be interesting to see if this behavior also arose using more related tasks. The differences between L2 Regularization and Elastic Weight Consolidation compared to Memory Aware Synapses and Synaptic Intelligence may also be caused by the slightly different regularization terms --- the former add a new loss term per task, while the latter add only a single term that is updated with each task. Aggregating all previous task constraints into a single term may somehow result in decreasing optimal regularization parameters with increasing interaction vertices, although the mechanism for this is not obvious if it exists at all. Testing more regularization-based sequential learning methods may provide more insight.

\subsection{Tuned Sequential Learning Methods}

Now we have tuned the DAM and all sequential learning methods we can finally compare the performance of each method to one another in a consistent manner. For rehearsal-based methods, GEM, and A-GEM, we investigate several values of the rehearsal / memory proportion, ranging from a relatively small proportion (\(0.05\)) to a modest proportion (\(0.5\)). Since the rehearsal-based and gradient-based methods plateaued after this point, we did not investigate larger proportions due to computational cost. For the regularization-based methods, we use regularization parameters \(\lambda\) as determined by the tunings conducted throughout Section \ref{Section: Regularization Method Parameters}. The values selected by method and interaction vertex are shown in Table \ref{Table: Regularization Lambda Selection}. Our experiments here consist of five permuted MNIST tasks of \(10000\) items each. Each experiment is repeated ten times for each combination of method and interaction vertex, aggregating the results in Table \ref{Table: Aggregated Results}.

\begin{table}[H]
\centering
    \begin{tabular}{ r  c  c  c  c }
        \hline
        & \(n=2\) & \(n=5\) & \(n=10\) & \(n=20\) \\
        \hline
        L2 Regularization	& \(2.0 \times 10^{-7}\) & \(4.8 \times 10^{-7}\) & \(1.0 \times 10^{-6}\) & \(9.0 \times 10^{-8}\) \\
        Elastic Weight Consolidation	& \(1.1 \times 10^{3}\) & \(7.5 \times 10^{2}\) & \(1.3 \times 10^{3}\) & \(1.8 \times 10^{3}\) \\
        Memory Aware Synapses	& \(5.8 \times 10^{1}\) & \(4.1 \times 10^{1}\) & \(6.7 \times 10^{0}\) & \(8.4 \times 10^{-1}\) \\
        Synaptic Intelligence	& \(3.0 \times 10^{0}\) & \(3.7 \times 10^{0}\) & \(3.2 \times 10^{-1}\) & \(3.1 \times 10^{-2}\) \\     
        \hline        
    \end{tabular}
    \caption{Selection of regularization hyperparameter \(\lambda\) across interaction vertices and sequential learning method.}
    \label{Table: Regularization Lambda Selection}
\end{table}

\begin{table}[H]
    \centering
    \begin{adjustbox}{center}
        \begin{tabular}{ l  c  c  c  c }
            \hline
            & \(n=2\) & \(n=5\) & \(n=10\) & \(n=20\) \\
            \hline
            Vanilla & \(0.366 \pm 0.016\) & \(0.292 \pm 0.010\) & \(0.284 \pm 0.014\) & \(0.277 \pm 0.007\)\\
            
            \hline
            \multicolumn{5}{c}{Rehearsal Methods - Memory Proportion of \(0.05\)}\\
            \hline
            Rehearsal & \boldmath\(0.822 \pm 0.003\) & \boldmath\(0.830 \pm 0.008\) & \boldmath\(0.857 \pm 0.019\) & \boldmath\(0.862 \pm 0.008\)\\
            Pseudorehearsal & \(0.421 \pm 0.021\) & \(0.398 \pm 0.026\) & \(0.535 \pm 0.022\) & \(0.667 \pm 0.029\)\\
            GEM & \(0.680 \pm 0.029\) & \(0.539 \pm 0.031\) & \(0.564 \pm 0.032\) & \(0.777 \pm 0.021\)\\
            A-GEM & \(0.690 \pm 0.019\) & \(0.506 \pm 0.028\) & \(0.517 \pm 0.025\) & \(0.736 \pm 0.069\)\\
            
            \hline
            \multicolumn{5}{c}{Rehearsal Methods - Memory Proportion of \(0.5\)}\\
            \hline
            Rehearsal & \boldmath\(0.873 \pm 0.005\) & \boldmath\(0.885 \pm 0.009\) & \boldmath\(0.907 \pm 0.007\) & \boldmath\(0.884 \pm 0.022\)\\
            Pseudorehearsal & \(0.399 \pm 0.031\) & \(0.358 \pm 0.025\) & \(0.522 \pm 0.047\) & \(0.754 \pm 0.016\)\\
            GEM & \(0.717 \pm 0.016\) & \(0.547 \pm 0.040\) & \(0.566 \pm 0.030\) & \(0.805 \pm 0.028\)\\
            A-GEM & \(0.710 \pm 0.019\) & \(0.544 \pm 0.033\) & \(0.539 \pm 0.018\) & \(0.774 \pm 0.044\)\\
            
            \hline
            \multicolumn{5}{c}{Regularization Methods}\\
            \hline
            L2 Regularization & \(0.585 \pm 0.016\) & \(0.525 \pm 0.013\) & \(0.479 \pm 0.031\) & \(0.511 \pm 0.029\)\\
            EWC & \(0.626 \pm 0.031\) & \(0.507 \pm 0.015\) & \(0.580 \pm 0.056\) & \(0.821 \pm 0.034\)\\
            MAS & \(0.735 \pm 0.013\) & \(0.594 \pm 0.020\) & \boldmath\(0.828 \pm 0.016\) & \boldmath\(0.827 \pm 0.019\)\\
            SI & \boldmath\(0.770 \pm 0.009\) & \boldmath\(0.683 \pm 0.020\) & \(0.576 \pm 0.048\) & \(0.725 \pm 0.042\)\\

            \hline        
        \end{tabular}
    \end{adjustbox}
    \caption{Average Accuracy over five Permuted MNIST tasks using different sequential learning methods. Vanilla indicates training with no additional sequential learning method. Methods with a number note the memory proportion used, either as rehearsal items or for gradient calculations. Results are aggregated over ten trials each, and the standard deviation is reported after the \(\pm\). The best average accuracy in each category is bolded.}
    \label{Table: Aggregated Results}
\end{table}

We find results that are largely in agreement with the hyperparameter tunings from Section \ref{Section: Method Hyperparameters}. The notable exception, of course, being Elastic Weight Consolidation which required tuning with tasks of a larger number of items than all other methods. Because of this, the Elastic Weight Consolidation results in Table \ref{Table: Aggregated Results} are at an advantage to the other methods.

Our baseline measurements, vanilla sequential learning, have the average accuracy decrease with the interaction vertex even up to \(n=20\). When employing a sequential learning method, however, we often find that performances increase with the interaction vertex (or at least increase between \(n=2\) to \(n=20\), even if the intermediate values are worse). The performance increase is therefore even more impressive for larger interaction vertices, since the baseline measurements were significantly lower than smaller vertices.

Naive rehearsal has high average accuracy and low standard deviation across all interaction vertices. Pseudorehearsal is consistently better than vanilla learning, even for low interaction vertices, but improves considerably for larger interaction vertices. GEM and A-GEM both have relatively high standard deviations, indicative of the (additional) instability these methods introduce. GEM and A-GEM both significantly outperform pseudorehearsal for \(n=2\) but a combination of the improved performance of pseudorehearsal and poor applicability of gradient-based methods to intermediate interaction vertices means the methods are far more even at other vertices. However, if one has access to buffer items that are sampled from the tasks themselves (which GEM and A-GEM require), it is always better to apply naive rehearsal. Pseudorehearsal, using only generated pseudoitems and hence requiring no access to previous task data, performs exceptionally well in spite of the different buffer makeup.

For regularization-based methods, L2-Regularization continues its shockingly consistent performance across each interaction vertex, although the accuracy leaves much to be desired. At \(n=2\), Memory Aware Synapses and Synaptic Intelligence perform very well, and even beat GEM and A-GEM using a memory proportion of \(0.5\). Synaptic Intelligence is the best regularization method for \(n=5\), however just like with rehearsal-based methods the accuracy is considerably lower than at \(n=2\). Memory Aware Synapses performs extremely well at \(n=10\) where other methods still struggle with the intermediate interaction vertex memory vectors. At \(n=20\), all methods perform relatively well, although Synaptic Intelligence lags behind Elastic Weight Consolidation and Memory Aware Synapses somewhat. It is disappointing that Elastic Weight Consolidation does not significantly outperform the other regularization-based methods, even when tuned specifically for these task sizes while other methods were not.
\section{Conclusions}

We have investigated the Dense Associative Memory in the context of sequential learning. We believe the application of this model in a sequential learning environment is novel, and gives some insight into the behavior of this associative memory architecture. Our studies include rehearsal-based (naive rehearsal, pseudorehearsal), gradient-based (Gradient Episodic Memories, Averaged Gradient Episodic Memories) and regularization-based (L2 Regularization, Elastic Weight Consolidation, Memory Aware Synapses, Synaptic Intelligence) sequential learning methods. Our studies shed light on several aspects of the DAM; the performance of the network in sequential learning tasks, the most effective sequential learning method as the interaction vertex varies, and the properties of the network more broadly.


Our work is a small step into the much larger potential intersection between sequential learning and the Dense Associative Memory and is by no means a comprehensive study on all aspects of either the field or the architecture. We hope our results can serve as a foundation for future work on sequential learning with the Dense Associative Memory. To give a non-exhaustive list of potential future research directions: our work focuses only on discrete domain tasks, are any results affected by the shift to a continuous domain? How do other, more exotic interaction functions (such as the exponential or log-sum-exp) affect the results here? Are there sequential learning methods that can take advantage of the highly interpretable nature of the DAM architecture rather than relying on general sequential learning methods developed for deep neural networks? How do the attractors of the model respond to sequential learning of related and unrelated tasks, and is this similar to the classical Hopfield network? Can the stability issues be resolved, allowing for learning of more than a small handful of tasks sequentially?

Before this paper there was a single well known behavior transition in the DAM: the feature-to-prototype transition. This transition is extremely prominent and visually striking. Our work found other behaviors that change based on the interaction vertex, such as the response to sequential learning methods, which may indicate that other transitions exist. However, these transitions are less obvious than the feature-to-prototype transition, and their exact properties are still elusive. Certainly this work shows that there are still aspects of the DAM that are yet to be discovered and understood, let alone the architectures derived from the DAM such as the Hierarchical Associative Memory \citep{Krotov2021B} and the Continuous Modern Hopfield Network \citep{HopfieldIsAllYouNeed2021}. In the context of sequential learning, our findings seem to indicate that the DAM operates reasonably typically with respect to other deep learning architectures, and existing sequential learning methods can be applied directly. Rehearsal-based methods were extremely effective, often approaching non-sequential accuracies. Gradient-based methods proved more volatile, destabilizing the learning of the DAM, fatally so for intermediate interaction vertices. Regularization-based methods, which incidentally are all weight-importance-based methods, show that the common methods for determining weight importance (e.g. Fisher Information) are applicable to the DAM, but typically underperformed for intermediate interaction vertices. This could indicate that the architecture of the DAM is unique when compared to other model architectures, and these general weight importance measures do not capture the essence of the DAM as well as in other deep neural network architectures. Finally, our work also found that the memory vectors of large interaction vertex networks are only strongly stabilized when a large volume of data is presented, which is not the case for lower interaction vertex networks --- that is, high interaction vertices are more data-hungry than lower ones.

\printbibliography

@misc{Aich2021,
  title      = {Elastic {Weight} {Consolidation} ({EWC}): {Nuts} and {Bolts}},
  shorttitle = {Elastic {Weight} {Consolidation} ({EWC})},
  doi        = {10.48550/arXiv.2105.04093},
  publisher  = {arXiv},
  author     = {Aich, Abhishek},
  month      = may,
  year       = {2021},
  note       = {arXiv:2105.04093 [cs, stat]}
}

@misc{Aljundi2018,
  title      = {Memory {Aware} {Synapses}: {Learning} what (not) to forget},
  shorttitle = {Memory {Aware} {Synapses}},
  doi        = {10.48550/arXiv.1711.09601},
  publisher  = {arXiv},
  author     = {Aljundi, Rahaf and Babiloni, Francesca and Elhoseiny, Mohamed and Rohrbach, Marcus and Tuytelaars, Tinne},
  month      = oct,
  year       = {2018},
  note       = {arXiv:1711.09601 [cs, stat]}
}

@article{Amit1985A,
  title    = {Spin-glass models of neural networks},
  volume   = {32},
  doi      = {10.1103/PhysRevA.32.1007},
  number   = {2},
  journal  = {Physical Review A},
  author   = {Amit, Daniel J. and Gutfreund, Hanoch and Sompolinsky, H.},
  month    = aug,
  year     = {1985},
  note     = {Publisher: American Physical Society},
  pages    = {1007--1018}
}

@article{Amit1985B,
  title    = {Storing {Infinite} {Numbers} of {Patterns} in a {Spin}-{Glass} {Model} of {Neural} {Networks}},
  volume   = {55},
  doi      = {10.1103/PhysRevLett.55.1530},
  number   = {14},
  journal  = {Physical Review Letters},
  author   = {Amit, Daniel J. and Gutfreund, Hanoch and Sompolinsky, H.},
  month    = sep,
  year     = {1985},
  note     = {Publisher: American Physical Society},
  pages    = {1530--1533}
}

@article{Amit1987,
  title    = {Statistical mechanics of neural networks near saturation},
  volume   = {173},
  issn     = {0003-4916},
  doi      = {10.1016/0003-4916(87)90092-3},
  number   = {1},
  journal  = {Annals of Physics},
  author   = {Amit, Daniel J and Gutfreund, Hanoch and Sompolinsky, H},
  month    = jan,
  year     = {1987},
  pages    = {30--67}
}

@incollection{Amit1994,
  address    = {Amsterdam},
  series     = {Pergamon {Studies} in {Neuroscience}},
  title      = {11 - {Psychology}, {Neurobiology} and {Modeling}: {The} {Science} of {Hebbian} {Reverberations}},
  shorttitle = {11 - {Psychology}, {Neurobiology} and {Modeling}},
  booktitle  = {Neural {Modeling} and {Neural} {Networks}},
  publisher  = {Pergamon},
  author     = {Amit, Daniel J.},
  editor     = {Ventriglia, F.},
  month      = jan,
  year       = {1994},
  doi        = {10.1016/B978-0-08-042277-0.50016-2},
  pages      = {251--281}
}

@article{Bovier2001,
  title    = {The spin-glass phase-transition in the {Hopfield} model with p-spin interactions},
  volume   = {5},
  doi      = {10.4310/ATMP.2001.v5.n6.a2},
  journal  = {Advances in Theoretical and Mathematical Physics},
  author   = {Bovier, Anton and Niederhauser, Beat},
  month    = sep,
  year     = {2001},
  pages    = {1001--1046}
}

@article{Burgess1991,
  title     = {Neural network models of list learning},
  copyright = {© 1991 Informa UK Ltd All rights reserved: reproduction in whole or part not permitted},
  doi       = {10.1088/0954-898X_2_4_005},
  language  = {EN},
  journal   = {Network: Computation in Neural Systems},
  author    = {Burgess, Neil and Shapiro, J. L. and Moore, M. A.},
  month     = jan,
  year      = {1991},
  note      = {Publisher: Taylor \& Francis}
}

@misc{Caccia2021,
  title      = {Online {Fast} {Adaptation} and {Knowledge} {Accumulation}: a {New} {Approach} to {Continual} {Learning}},
  shorttitle = {Online {Fast} {Adaptation} and {Knowledge} {Accumulation}},
  language   = {en},
  publisher  = {arXiv},
  author     = {Caccia, Massimo and Rodriguez, Pau and Ostapenko, Oleksiy and Normandin, Fabrice and Lin, Min and Caccia, Lucas and Laradji, Issam and Rish, Irina and Lacoste, Alexandre and Vazquez, David and Charlin, Laurent},
  month      = jan,
  year       = {2021},
  note       = {arXiv:2003.05856 [cs]}
}

@misc{Chaudhry2019,
  title     = {On {Tiny} {Episodic} {Memories} in {Continual} {Learning}},
  doi       = {10.48550/arXiv.1902.10486},
  publisher = {arXiv},
  author    = {Chaudhry, Arslan and Rohrbach, Marcus and Elhoseiny, Mohamed and Ajanthan, Thalaiyasingam and Dokania, Puneet K. and Torr, Philip H. S. and Ranzato, Marc'Aurelio},
  month     = jun,
  year      = {2019},
  note      = {arXiv:1902.10486 [cs, stat]}
}

@article{Checiu2024,
  title      = {Reconstructing creative thoughts: {Hopfield} neural networks},
  volume     = {575},
  issn       = {0925-2312},
  shorttitle = {Reconstructing creative thoughts},
  doi        = {10.1016/j.neucom.2024.127324},
  journal    = {Neurocomputing},
  author     = {Checiu, Denisa and Bode, Mathias and Khalil, Radwa},
  month      = mar,
  year       = {2024},
  pages      = {127324}
}

@article{Demircigil2017,
  title    = {On a {Model} of {Associative} {Memory} with {Huge} {Storage} {Capacity}},
  volume   = {168},
  issn     = {0022-4715, 1572-9613},
  doi      = {10.1007/s10955-017-1806-y},
  language = {en},
  number   = {2},
  journal  = {Journal of Statistical Physics},
  author   = {Demircigil, Mete and Heusel, Judith and Löwe, Matthias and Upgang, Sven and Vermet, Franck},
  month    = jul,
  year     = {2017},
  pages    = {288--299}
}

@misc{Douillard2022,
  title      = {{DyTox}: {Transformers} for {Continual} {Learning} with {DYnamic} {TOken} {eXpansion}},
  shorttitle = {{DyTox}},
  doi        = {10.48550/arXiv.2111.11326},
  publisher  = {arXiv},
  author     = {Douillard, Arthur and Ramé, Alexandre and Couairon, Guillaume and Cord, Matthieu},
  month      = aug,
  year       = {2022},
  note       = {arXiv:2111.11326 [cs]}
}

@misc{Goodfellow2015,
  title     = {An {Empirical} {Investigation} of {Catastrophic} {Forgetting} in {Gradient}-{Based} {Neural} {Networks}},
  language  = {en},
  urldate   = {2024-08-20},
  publisher = {arXiv},
  author    = {Goodfellow, Ian J. and Mirza, Mehdi and Xiao, Da and Courville, Aaron and Bengio, Yoshua},
  month     = mar,
  year      = {2015},
  note      = {arXiv:1312.6211 [cs, stat]}
}

@book{Hebb1949,
  address   = {Oxford, England},
  series    = {The organization of behavior; a neuropsychological theory},
  title     = {The organization of behavior; a neuropsychological theory},
  publisher = {Wiley},
  author    = {Hebb, D. O.},
  year      = {1949},
  note      = {Pages: xix, 335}
}

@book{Hertz1991,
  address   = {Boca Raton},
  title     = {Introduction {To} {The} {Theory} {Of} {Neural} {Computation}},
  isbn      = {978-0-429-49966-1},
  publisher = {CRC Press},
  author    = {Hertz, John A.},
  year      = {1991},
  doi       = {10.1201/9780429499661}
}

@article{Hopfield1982,
  title        = {Neural networks and physical systems with emergent collective computational abilities.},
  volume       = {79},
  issn         = {0027-8424},
  pages        = {2554--2558},
  number       = {8},
  journaltitle = {Proceedings of the National Academy of Sciences of the United States of America},
  shortjournal = {Proc Natl Acad Sci U S A},
  author       = {Hopfield, J J},
  urldate      = {2024-02-22},
  date         = {1982-04},
  year         = {1982},
  pmid         = {6953413},
  pmcid        = {PMC346238}
}

@article{Hopfield1984,
  title   = {Neurons with {Graded} {Response} {Have} {Collective} {Computational} {Properties} like {Those} of {Two}-{State} {Neurons}},
  volume  = {81},
  issn    = {0027-8424},
  number  = {10},
  journal = {Proceedings of the National Academy of Sciences of the United States of America},
  author  = {Hopfield, J. J.},
  year    = {1984},
  note    = {Publisher: National Academy of Sciences},
  pages   = {3088--3092}
}

@article{Hopfield1999,
  title   = {Brain, neural networks, and computation},
  volume  = {71},
  doi     = {10.1103/RevModPhys.71.S431},
  number  = {2},
  journal = {Reviews of Modern Physics},
  author  = {Hopfield, J. J.},
  month   = mar,
  year    = {1999},
  note    = {Publisher: American Physical Society},
  pages   = {S431--S437}
}

@misc{HopfieldIsAllYouNeed2021,
  title     = {Hopfield {Networks} is {All} {You} {Need}},
  language  = {en},
  publisher = {arXiv},
  author    = {Ramsauer, Hubert and Schäfl, Bernhard and Lehner, Johannes and Seidl, Philipp and Widrich, Michael and Adler, Thomas and Gruber, Lukas and Holzleitner, Markus and Pavlović, Milena and Sandve, Geir Kjetil and Greiff, Victor and Kreil, David and Kopp, Michael and Klambauer, Günter and Brandstetter, Johannes and Hochreiter, Sepp},
  month     = apr,
  year      = {2021},
  note      = {arXiv:2008.02217 [cs, stat]}
}

@inproceedings{Houlsby2019,
  title     = {Parameter-{Efficient} {Transfer} {Learning} for {NLP}},
  language  = {en},
  urldate   = {2024-08-20},
  booktitle = {Proceedings of the 36th {International} {Conference} on {Machine} {Learning}},
  publisher = {PMLR},
  author    = {Houlsby, Neil and Giurgiu, Andrei and Jastrzebski, Stanislaw and Morrone, Bruna and Laroussilhe, Quentin De and Gesmundo, Andrea and Attariyan, Mona and Gelly, Sylvain},
  month     = may,
  year      = {2019},
  note      = {ISSN: 2640-3498},
  pages     = {2790--2799}
}

@article{Kirkpatrick2017,
  title    = {Overcoming catastrophic forgetting in neural networks},
  volume   = {114},
  issn     = {0027-8424, 1091-6490},
  doi      = {10.1073/pnas.1611835114},
  language = {en},
  number   = {13},
  journal  = {Proceedings of the National Academy of Sciences},
  author   = {Kirkpatrick, James and Pascanu, Razvan and Rabinowitz, Neil and Veness, Joel and Desjardins, Guillaume and Rusu, Andrei A. and Milan, Kieran and Quan, John and Ramalho, Tiago and Grabska-Barwinska, Agnieszka and Hassabis, Demis and Clopath, Claudia and Kumaran, Dharshan and Hadsell, Raia},
  month    = mar,
  year     = {2017},
  note     = {arXiv:1612.00796 [cs, stat]},
  pages    = {3521--3526}
}

@article{KirkpatrickSherrington1978,
  title    = {Infinite-ranged models of spin-glasses},
  volume   = {17},
  doi      = {10.1103/PhysRevB.17.4384},
  number   = {11},
  journal  = {Physical Review B},
  author   = {Kirkpatrick, Scott and Sherrington, David},
  month    = jun,
  year     = {1978},
  note     = {Publisher: American Physical Society},
  pages    = {4384--4403}
}

@article{Kohonen1972,
  title    = {Correlation {Matrix} {Memories}},
  volume   = {C-21},
  issn     = {0018-9340},
  doi      = {10.1109/TC.1972.5008975},
  language = {en},
  number   = {4},
  journal  = {IEEE Transactions on Computers},
  author   = {Kohonen, Teuvo},
  month    = apr,
  year     = {1972},
  pages    = {353--359}
}

@book{Kohonen1978,
  title      = {Associative {Memory}: {A} {System}-{Theoretical} {Approach}},
  isbn       = {978-3-642-96384-1},
  shorttitle = {Associative {Memory}},
  language   = {en},
  publisher  = {Springer Science \& Business Media},
  author     = {Kohonen, T.},
  month      = dec,
  year       = {1978},
  note       = {Google-Books-ID: 7wPtCAAAQBAJ}
}

@article{Krizhevsky2009,
  title    = {Learning {Multiple} {Layers} of {Features} from {Tiny} {Images}},
  year     = {2009},
  language = {en},
  journal  = {University of Toronto, ON, Canada},
  author   = {Krizhevsky, Alex and Hinton, Geoffrey}
}

@inproceedings{Krizhevsky2012,
  title     = {{ImageNet} {Classification} with {Deep} {Convolutional} {Neural} {Networks}},
  volume    = {25},
  urldate   = {2024-08-20},
  booktitle = {Advances in {Neural} {Information} {Processing} {Systems}},
  publisher = {Curran Associates, Inc.},
  author    = {Krizhevsky, Alex and Sutskever, Ilya and Hinton, Geoffrey E},
  year      = {2012}
}

@misc{KrotovHopfield2016,
  title     = {Dense {Associative} {Memory} for {Pattern} {Recognition}},
  doi       = {10.48550/arXiv.1606.01164},
  publisher = {arXiv},
  author    = {Krotov, Dmitry and Hopfield, John J.},
  month     = sep,
  year      = {2016},
  note      = {arXiv:1606.01164 [cond-mat, q-bio, stat]}
}

@article{KrotovHopfield2018,
  title    = {Dense {Associative} {Memory} {Is} {Robust} to {Adversarial} {Inputs}},
  volume   = {30},
  issn     = {0899-7667},
  doi      = {10.1162/neco_a_01143},
  number   = {12},
  journal  = {Neural Computation},
  author   = {Krotov, Dmitry and Hopfield, John},
  month    = dec,
  year     = {2018},
  pages    = {3151--3167}
}

@article{Lampert2009,
  title    = {Learning to detect unseen object classes by between-class attribute transfer},
  doi      = {10.1109/CVPR.2009.5206594},
  urldate  = {2024-08-20},
  journal  = {2009 IEEE Conference on Computer Vision and Pattern Recognition},
  author   = {Lampert, Christoph H. and Nickisch, Hannes and Harmeling, Stefan},
  month    = jun,
  year     = {2009},
  note     = {Conference Name: 2009 IEEE Computer Society Conference on Computer Vision and Pattern Recognition Workshops (CVPR Workshops)
              ISBN: 9781424439928
              Place: Miami, FL
              Publisher: IEEE},
  pages    = {951--958}
}

@misc{LeCun1998,
  title    = {{MNIST} handwritten digit database, {Yann} {LeCun}, {Corinna} {Cortes} and {Chris} {Burges}},
  year     = {1998},
  author   = {LeCun, Yann and Cortes, Corrina and Burges, Christopher}
}

@article{Lewis2004,
  title      = {{RCV1}: {A} {New} {Benchmark} {Collection} for {Text} {Categorization} {Research}.},
  volume     = {5},
  shorttitle = {{RCV1}},
  journal    = {Journal of Machine Learning Research},
  author     = {Lewis, David and Yang, Yiming and Russell-Rose, Tony and Li, Fan},
  month      = apr,
  year       = {2004},
  pages      = {361--397}
}

@misc{Li2022,
  title      = {Technical {Report} for {ICCV} 2021 {Challenge} {SSLAD}-{Track3B}: {Transformers} {Are} {Better} {Continual} {Learners}},
  shorttitle = {Technical {Report} for {ICCV} 2021 {Challenge} {SSLAD}-{Track3B}},
  doi        = {10.48550/arXiv.2201.04924},
  urldate    = {2024-08-20},
  publisher  = {arXiv},
  author     = {Li, Duo and Cao, Guimei and Xu, Yunlu and Cheng, Zhanzhan and Niu, Yi},
  month      = jan,
  year       = {2022},
  note       = {arXiv:2201.04924 [cs]}
}

@misc{Lopez-Paz2017,
  title     = {Gradient {Episodic} {Memory} for {Continual} {Learning}},
  doi       = {10.48550/arXiv.1706.08840},
  publisher = {arXiv},
  author    = {Lopez-Paz, David and Ranzato, Marc'Aurelio},
  month     = sep,
  year      = {2017},
  note      = {arXiv:1706.08840 [cs]}
}

@article{Lund1925,
  title    = {The psychology of belief},
  volume   = {20},
  issn     = {0096-851X},
  doi      = {10.1037/h0076047},
  number   = {1},
  journal  = {The Journal of Abnormal and Social Psychology},
  author   = {Lund, F. H.},
  year     = {1925},
  note     = {Place: US
              Publisher: American Psychological Association},
  pages    = {63--81; 174--195}
}

@article{Maass1997,
  title    = {Networks of spiking neurons can emulate arbitrary {Hopfield} nets in temporal coding},
  volume   = {8},
  issn     = {0954-898X},
  doi      = {10.1088/0954-898X_8_4_002},
  number   = {4},
  journal  = {Network: Computation in Neural Systems},
  author   = {Maass, Wolfgang and Natschläger, Thomas},
  month    = jan,
  year     = {1997},
  note     = {Publisher: Taylor \& Francis
              \_eprint: https://doi.org/10.1088/0954-898X\_8\_4\_002},
  pages    = {355--371}
}

@article{McAlister2024A,
  title = {Prototype {{Analysis}} in {{Hopfield Networks With Hebbian Learning}}},
  author = {McAlister, Hayden and Robins, Anthony and Szymanski, Lech},
  year = {2024},
  month = oct,
  journal = {Neural Computation},
  volume = {36},
  number = {11},
  pages = {2322--2364},
  issn = {0899-7667},
  doi = {10.1162/neco_a_01704},
  urldate = {2025-01-23}
}

@misc{McAlister2024B,
  title     = {Improved {Robustness} and {Hyperparameter} {Selection} in {Modern} {Hopfield} {Networks}},
  doi       = {10.48550/arXiv.2407.08742},
  urldate   = {2024-08-20},
  publisher = {arXiv},
  author    = {McAlister, Hayden and Robins, Anthony and Szymanski, Lech},
  month     = jul,
  year      = {2024},
  note      = {arXiv:2407.08742 [cs]}
}

@article{McEliece1987,
  title    = {The capacity of the {Hopfield} associative memory},
  volume   = {33},
  issn     = {1557-9654},
  doi      = {10.1109/TIT.1987.1057328},
  number   = {4},
  journal  = {IEEE Transactions on Information Theory},
  author   = {McEliece, R. and Posner, E. and Rodemich, E. and Venkatesh, S.},
  month    = jul,
  year     = {1987},
  note     = {Conference Name: IEEE Transactions on Information Theory},
  pages    = {461--482}
}

@article{Nadal1986,
  title    = {Networks of {Formal} {Neurons} and {Memory} {Palimpsests}},
  volume   = {1},
  issn     = {0295-5075},
  doi      = {10.1209/0295-5075/1/10/008},
  language = {en},
  number   = {10},
  journal  = {Europhysics Letters},
  author   = {Nadal, J. P. and Toulouse, G. and Changeux, J. P. and Dehaene, S.},
  month    = may,
  year     = {1986},
  pages    = {535}
}

@inproceedings{Robins1993,
  title      = {Catastrophic forgetting in neural networks: the role of rehearsal mechanisms},
  shorttitle = {Catastrophic forgetting in neural networks},
  doi        = {10.1109/ANNES.1993.323080},
  booktitle  = {Proceedings 1993 {The} {First} {New} {Zealand} {International} {Two}-{Stream} {Conference} on {Artificial} {Neural} {Networks} and {Expert} {Systems}},
  author     = {Robins, Anthony},
  month      = nov,
  year       = {1993},
  pages      = {65--68}
}

@article{Robins1995,
  title     = {Catastrophic {Forgetting}, {Rehearsal} and {Pseudorehearsal}},
  copyright = {Copyright Taylor \& Francis Group, LLC},
  doi       = {10.1080/09540099550039318},
  language  = {EN},
  journal   = {Connection Science},
  author    = {Robins, Anthony},
  month     = jun,
  year      = {1995},
  note      = {Publisher: Taylor \& Francis Group}
}

@article{Robins1998,
  title     = {Catastrophic {Forgetting} and the {Pseudorehearsal} {Solution} in {Hopfield}-type {Networks}},
  copyright = {Copyright Taylor and Francis Group, LLC},
  doi       = {10.1080/095400998116530},
  language  = {EN},
  journal   = {Connection Science},
  author    = {Robins, Anthony and McCallum, Simon},
  month     = jun,
  year      = {1998},
  note      = {Publisher: Taylor \& Francis Group}
}

@article{Robins2004,
  title    = {A robust method for distinguishing between learned and spurious attractors},
  volume   = {17},
  issn     = {0893-6080},
  doi      = {10.1016/j.neunet.2003.11.007},
  language = {eng},
  number   = {3},
  journal  = {Neural Networks: The Official Journal of the International Neural Network Society},
  author   = {Robins, Anthony V. and McCallum, Simon J. R.},
  month    = apr,
  year     = {2004},
  pmid     = {15037350},
  pages    = {313--326}
}

@article{Shiffrin1969,
  title    = {Storage and retrieval processes in long-term memory},
  volume   = {76},
  issn     = {1939-1471},
  doi      = {10.1037/h0027277},
  number   = {2},
  journal  = {Psychological Review},
  author   = {Shiffrin, R. M. and Atkinson, R. C.},
  year     = {1969},
  note     = {Place: US
              Publisher: American Psychological Association},
  pages    = {179--193}
}

@article{Snow2024,
  title    = {Biological {Plausibility} in {Modern} {Hopfield} {Networks}},
  language = {en},
  author   = {Snow, Mallory},
  month    = dec,
  year     = {2022},
  note     = {Publisher: University of Waterloo}
}

@article{Steinbuch1963,
  title    = {Learning matrices and their applications},
  volume   = {EC-12},
  issn     = {0367-7508},
  doi      = {10.1109/PGEC.1963.263588},
  number   = {6},
  journal  = {IEEE Transactions on Electronic Computers},
  author   = {Steinbuch, K. and Piske, U. A. W.},
  month    = dec,
  year     = {1963},
  note     = {Conference Name: IEEE Transactions on Electronic Computers}
}

@article{Steinbuch1965,
  title    = {Adaptive networks using learning matrices},
  volume   = {2},
  issn     = {1432-0770},
  doi      = {10.1007/BF00272311},
  language = {en},
  number   = {4},
  journal  = {Kybernetik},
  author   = {Steinbuch, Karl},
  month    = feb,
  year     = {1965},
  pages    = {148--152}
}

@inproceedings{Stork1989,
  title     = {Is backpropagation biologically plausible?},
  doi       = {10.1109/IJCNN.1989.118705},
  booktitle = {International 1989 {Joint} {Conference} on {Neural} {Networks}},
  author    = {{Stork}},
  month     = jun,
  year      = {1989},
  pages     = {241--246 vol.2}
}

@inproceedings{Storkey1997,
  address   = {Berlin, Heidelberg},
  series    = {Lecture {Notes} in {Computer} {Science}},
  title     = {Increasing the capacity of a hopfield network without sacrificing functionality},
  isbn      = {978-3-540-69620-9},
  doi       = {10.1007/BFb0020196},
  language  = {en},
  booktitle = {Artificial {Neural} {Networks} — {ICANN}'97},
  publisher = {Springer},
  author    = {Storkey, Amos},
  editor    = {Gerstner, Wulfram and Germond, Alain and Hasler, Martin and Nicoud, Jean-Daniel},
  year      = {1997},
  pages     = {451--456}
}

@article{Tsodyks1999,
  title     = {Attractor neural network models of spatial maps in hippocampus},
  volume    = {9},
  copyright = {Copyright © 1999 Wiley-Liss, Inc.},
  issn      = {1098-1063},
  doi       = {10.1002/(SICI)1098-1063(1999)9:4<481::AID-HIPO14>3.0.CO;2-S},
  language  = {fr},
  number    = {4},
  journal   = {Hippocampus},
  author    = {Tsodyks, Misha},
  year      = {1999},
  pages     = {481--489}
}

@article{Vaswani2017,
  address   = {Red Hook, NY, USA},
  series    = {{NIPS}'17},
  title     = {Attention is all you need},
  isbn      = {978-1-5108-6096-4},
  booktitle = {Proceedings of the 31st {International} {Conference} on {Neural} {Information} {Processing} {Systems}},
  publisher = {Curran Associates Inc.},
  author    = {Vaswani, Ashish and Shazeer, Noam and Parmar, Niki and Uszkoreit, Jakob and Jones, Llion and Gomez, Aidan N. and Kaiser, Lukasz and Polosukhin, Illia},
  month     = dec,
  year      = {2017},
  pages     = {6000--6010}
}

@article{Welinder2010,
  title    = {Caltech-{UCSD} {Birds} 200},
  author   = {Welinder, Peter and Branson, Steve and Mita, Takeshi and Wah, Catherine and Schroff, Florian and Belongie, Serge and Perona, Pietro},
  month    = sep,
  year     = {2010}
}

@article{WidrowHoff1960,
  title      = {Adaptive switching circuits},
  shorttitle = {(1960) {Bernard} {Widrow} and {Marcian} {E}. {Hoff}, {Adaptive} switching circuits, 1960 {IRE} {WESCON} {Convention} {Record}, {New} {York}},
  doi        = {10.7551/mitpress/4943.003.0012},
  language   = {en},
  author     = {Widrow, Bernard and Hoff, Marcian E.},
  year       = {1960}
}

@inproceedings{Yang2018,
  title      = {{SGM}: {Sequence} {Generation} {Model} for {Multi}-label {Classification}},
  shorttitle = {{SGM}},
  urldate    = {2024-08-20},
  author     = {Yang, Pengcheng and Sun, Xu and Li, Wei and Ma, Shuming and Wu, Wei and Wang, Houfeng},
  month      = jun,
  year       = {2018}
}

@inproceedings{Zenke2017,
  title     = {Continual {Learning} {Through} {Synaptic} {Intelligence}},
  language  = {en},
  booktitle = {Proceedings of the 34th {International} {Conference} on {Machine} {Learning}},
  publisher = {PMLR},
  author    = {Zenke, Friedemann and Poole, Ben and Ganguli, Surya},
  month     = jul,
  year      = {2017},
  note      = {ISSN: 2640-3498},
  pages     = {3987--3995}
}

@misc{Zubiaga2012,
  title     = {Enhancing {Navigation} on {Wikipedia} with {Social} {Tags}},
  doi       = {10.48550/arXiv.1202.5469},
  urldate   = {2024-08-20},
  publisher = {arXiv},
  author    = {Zubiaga, Arkaitz},
  month     = feb,
  year      = {2012},
  note      = {arXiv:1202.5469 [cs]}
}

@incollection{Silver2015,
	address = {Cham},
	title = {Consolidation {Using} {Sweep} {Task} {Rehearsal}: {Overcoming} the {Stability}-{Plasticity} {Problem}},
	volume = {9091},
	isbn = {978-3-319-18355-8 978-3-319-18356-5},
	shorttitle = {Consolidation {Using} {Sweep} {Task} {Rehearsal}},
	url = {https://link.springer.com/10.1007/978-3-319-18356-5_27},
	language = {en},
	urldate = {2024-08-22},
	booktitle = {Advances in {Artificial} {Intelligence}},
	publisher = {Springer International Publishing},
	author = {Silver, Daniel L. and Mason, Geoffrey and Eljabu, Lubna},
	editor = {Barbosa, Denilson and Milios, Evangelos},
	year = {2015},
	doi = {10.1007/978-3-319-18356-5_27},
	note = {Series Title: Lecture Notes in Computer Science},
	pages = {307--322},
}

@article{Atkinson2021,
	title = {Pseudo-{Rehearsal}: {Achieving} {Deep} {Reinforcement} {Learning} without {Catastrophic} {Forgetting}},
	volume = {428},
	issn = {09252312},
	shorttitle = {Pseudo-{Rehearsal}},
	url = {http://arxiv.org/abs/1812.02464},
	doi = {10.1016/j.neucom.2020.11.050},
	language = {en},
	urldate = {2024-08-22},
	journal = {Neurocomputing},
	author = {Atkinson, Craig and McCane, Brendan and Szymanski, Lech and Robins, Anthony},
	month = mar,
	year = {2021},
	note = {arXiv:1812.02464 [cs]},
	pages = {291--307},
}

@misc{Chaudhry2018B,
	title = {Efficient {Lifelong} {Learning} with {A}-{GEM}},
	url = {https://arxiv.org/abs/1812.00420v2},
	language = {en},
	urldate = {2024-07-31},
	journal = {arXiv.org},
	author = {Chaudhry, Arslan and Ranzato, Marc'Aurelio and Rohrbach, Marcus and Elhoseiny, Mohamed},
	month = dec,
	year = {2018},
}

@misc{Pascanu2014,
	title = {Revisiting {Natural} {Gradient} for {Deep} {Networks}},
	url = {http://arxiv.org/abs/1301.3584},
	doi = {10.48550/arXiv.1301.3584},
	urldate = {2024-08-28},
	publisher = {arXiv},
	author = {Pascanu, Razvan and Bengio, Yoshua},
	month = feb,
	year = {2014},
	note = {arXiv:1301.3584 [cs]},
}

@article{Toulouse1986,
	title = {Spin glass model of learning by selection},
	volume = {83},
	issn = {0027-8424},
	doi = {10.1073/pnas.83.6.1695},
	language = {eng},
	number = {6},
	journal = {Proceedings of the National Academy of Sciences of the United States of America},
	author = {Toulouse, G. and Dehaene, S. and Changeux, J. P.},
	month = mar,
	year = {1986},
	pmid = {3456609},
	pmcid = {PMC323150},
	pages = {1695--1698},
}

@article{Personnaz1986,
	title = {A biologically constrained learning mechanism in networks of formal neurons},
	volume = {43},
	issn = {1572-9613},
	url = {https://doi.org/10.1007/BF01020645},
	doi = {10.1007/BF01020645},
	language = {en},
	number = {3},
	urldate = {2024-09-09},
	journal = {Journal of Statistical Physics},
	author = {Personnaz, L. and Guyon, I. and Dreyfus, G. and Toulouse, G.},
	month = may,
	year = {1986},
	pages = {411--422},
}

@article{Frean1992,
	title = {A "{Thermal}" {Perceptron} {Learning} {Rule}},
	volume = {4},
	issn = {0899-7667},
	url = {https://doi.org/10.1162/neco.1992.4.6.946},
	doi = {10.1162/neco.1992.4.6.946},
	number = {6},
	urldate = {2024-09-09},
	journal = {Neural Computation},
	author = {Frean, Marcus},
	month = nov,
	year = {1992},
	pages = {946--957},
}

@inproceedings{Optuna2019,
	address = {New York, NY, USA},
	series = {{KDD} '19},
	title = {Optuna: {A} {Next}-generation {Hyperparameter} {Optimization} {Framework}},
	isbn = {978-1-4503-6201-6},
	shorttitle = {Optuna},
	url = {https://dl.acm.org/doi/10.1145/3292500.3330701},
	doi = {10.1145/3292500.3330701},
	urldate = {2024-09-09},
	booktitle = {Proceedings of the 25th {ACM} {SIGKDD} {International} {Conference} on {Knowledge} {Discovery} \& {Data} {Mining}},
	publisher = {Association for Computing Machinery},
	author = {Akiba, Takuya and Sano, Shotaro and Yanase, Toshihiko and Ohta, Takeru and Koyama, Masanori},
	month = jul,
	year = {2019},
	pages = {2623--2631},
}

@misc{Krotov2021B,
  title = {Hierarchical {{Associative Memory}}},
  author = {Krotov, Dmitry},
  year = {2021},
  month = jul,
  number = {arXiv:2107.06446},
  eprint = {2107.06446},
  primaryclass = {cond-mat, q-bio, stat},
  publisher = {arXiv},
  doi = {10.48550/arXiv.2107.06446},
  urldate = {2024-02-22},
  archiveprefix = {arXiv}
}

\appendix
\section{Sequential Learning Methods}
\label{Appendix: Methods}

\subsection{Naive Rehearsal}

While rehearsal-based methods may appear simple at first glance, there is some depth to them. The buffer items need not be taken directly from the training data of previous tasks and could instead be generated another way, meaning the assumption of access to previous tasks is not violated. Furthermore, the method by which rehearsal items are introduced during training may vary, from naively introducing the entire buffer at each epoch to sampling a new mini-buffer that is updated every epoch, known as sweep-rehearsal \citep{Robins1993, Robins1995, Silver2015}.

In naive rehearsal we take some proportion of the previous task items and store them in a buffer to introduce during the next task's training. We take a constant proportion across all tasks, e.g. 100 items from task one, 100 from task two, and so on. Typically, some predefined buffer size is used, and the buffer is utilized entirely throughout learning (i.e. initially the entire buffer are items from the first task, after task two the buffer is half items from task one, half from task two, and so on). However, we found excellent results using a growing buffer, with considerably shorter training times. Our rehearsal implementation combined the buffer with the new task data, then shuffled and batched this. This means individual buffer items are seen as frequently as individual new items --- once per epoch --- and effectively implements a type of sweep rehearsal (as the buffer items are newly sampled each batch). We may be able to save some memory by replaying the same buffer items more than once, but our investigation here is foundational, and hence the best-case scenario with a large number of buffer items is still interesting. We parameterize naive rehearsal using the rehearsal proportion: a rehearsal proportion of \(1.0\) would be to present the entire previous task alongside the new task, while \(0.1\) would be to present only \(10\%\) of the previous task items.

\subsection{Pseudorehearsal}

Pseudorehearsal \citep{Robins1995}, similar to naive rehearsal, is a rehearsal-based sequential learning method. In pseudorehearsal, however, buffer items are not sampled from the training data of previous tasks, but instead from randomly generated data that is fed through the model. This removes the need for access to the previous training data, as only the model itself is required to generate buffer items. Pseudorehearsal has been studied in the Hopfield network \citep{Robins1998}, both using both homogenous (a random state is relaxed to a stable point, which is taken as the buffer item) and heterogeneous methods (a random state is relaxed to a stable one, and the \textit{pair} is taken as a buffer item to ensure the mapping remains consistent during training of the next task). We will focus on the homogenous method. After each task, we will generate some number of random probes (based on the amount of training data in the previous task, so we can compare the buffer size to the rehearsal method above) and relax these probes to a stable state. We then treat these relaxed probes exactly as if they were sampled from the training data using naive rehearsal --- combine with the data for the next task, shuffle, batch, and train.

We theorize (and do indeed find) that pseudorehearsal will be extremely effective at higher interaction vertices. It is known that the memories of the DAM undergo a feature-to-prototype transition as the interaction vertex is raised \citep{KrotovHopfield2016,KrotovHopfield2018}. When the memories are learned to be prototypes of the training data, we expect random probes to stabilize on or near those prototype states, meaning we get buffer items that are extremely representative of the previous task training data without requiring explicit access to the data itself. This should improve the efficacy of pseudorehearsal, which has always suffered from buffer items that are not representative of training data \citep{Atkinson2021}.

\subsection{Gradient Episodic Memories}

The following methods, GEM and A-GEM, are somewhat hybrid regularization- and rehearsal-based methods. Gradient Episodic Memories \citep{Lopez-Paz2017} aims to preserve performance on previous tasks by ensuring the gradient vector of the loss on the current task never points against the gradient of the loss on previous tasks. To measure the gradient on previous tasks, a buffer of previous task items is kept and evaluated each batch to find the gradients. Therefore, Gradient Episodic Memories \textit{regularizes} the weight updates by \textit{rehearsing} previous task items. One of the main benefits of GEM over other sequential learning methods, particularly the quadratic-penalty weight-importance methods we discuss below, is the explicit allowance of improvement on previous tasks. Because the gradient of the loss is allowed to point in the same direction as previous task's losses, this allows the model to update weights away from the optimal weights found at the end of the previous task as long as the update is believed to improve the loss on that previous task (rather, improve the loss on the buffer sampled from the previous task). The motivation for this difference in behavior is to allow for backwards transfer, where future tasks can improve performance on previous tasks, something that is much more difficult to achieve when weight updates are penalized significantly.

Lopez-Paz and Ranzato alter the typical learning objective with constraints on previous task performance:
\begin{equation}
    \begin{aligned}
        \text{minimize}_{\theta}    &\qquad \mathcal{L}\left(\theta, X_\mu\right) \\
        \text{subject to}           &\qquad \mathcal{L}\left(\theta, \mathcal{M}_\nu\right) \le \mathcal{L}\left(\theta_{\mu-1}, \mathcal{M}_\nu\right) \text{ for } \nu<\mu,
    \end{aligned}
\end{equation}
that is, minimize the current task loss \(\mathcal{L}\left(\theta, X_\mu\right)\) as long as the loss on the previous task's buffer \(\mathcal{M}_\nu\) does not increase with respect to the start of the current task. Here, \(\mathcal{L}\) is the base loss, as no surrogate loss terms are introduced, and \(\mu\) indexes over tasks. Notably, we do not need to store the entire old model as long as we store just the loss on the buffer items. This objective can be optimized while respecting the constraints by noting that if the gradient of the loss points in the opposite direction to any previous task's gradient, we may project the current gradient on to the closest vector that doesn't. Formally, given the gradient of the loss with respect to \(\theta\), which we label \(g\), and the gradient of the loss on previous tasks, \(g_\nu\), we calculate the projected gradient \(\tilde{g}\) by solving another optimization problem:
\begin{equation}
    \label{Eqn:GEM Gradient Projection}
    \begin{aligned}
        \text{minimize}_{\tilde{g}} &\qquad \frac{1}{2} \lvert\lvert g - \tilde{g} \rvert\rvert _2^2 \\
        \text{subject to} &\qquad \langle \tilde{g}, g_\nu \rangle \ge 0 \quad \forall \nu<\mu.
    \end{aligned}
\end{equation}
We can express this as a quadratic programming problem of the number of parameters in the network:
\begin{equation}
    \begin{aligned}
        \text{minimize}_{z} &\qquad \frac{1}{2} z^\top z  - g^\top z + \frac{1}{2} g^\top g \\
        \text{subject to} &\qquad Gz \ge 0,
    \end{aligned}
\end{equation}
Where \(G = \left(g_1, g_2, \dots, g_{\mu-1}\right)\). The constant term \(g^\top g\) is discarded. However, since the network is likely to have many parameters, this problem is often intractable to solve, so we shift the problem to the dual space, which has only \(\mu-1\) variables instead:
\begin{equation}
    \begin{aligned}
        \text{minimize}_{v} &\qquad \frac{1}{2} v^\top G G^\top v + g^\top G^\top v \\
        \text{subject to} &\qquad v \ge 0.
    \end{aligned}
\end{equation}
Once we solve this problem to find \(v^\star\), we can get the projected gradient vector:
\begin{align}
    \tilde{g} &= G^\top v^\star + g,
\end{align}
which we can use for weight updates as usual.

While GEM is theoretically sound, its main issue comes with an increasing number of constraints as the number of tasks grows. Every new task adds a new constraint to the primal problem, which in turns adds a new parameter to the dual problem that is actually solved. These increasingly limit the directions in which the projected gradient is allowed to point, potentially leading to very poor learning on new tasks when the number of previous tasks is large. However, for large networks the curse of dimensionality means the number of directions is extremely large, so this is not as significant a problem as it may seem. To find the vector \(G\), we must compute the gradient on a buffer of items from the previous task, which is also computationally expensive, and requires us to store a buffer of items like a rehearsal-based task. We will again parameterize this method using a rehearsal proportion hyperparameter to see how different buffer sizes impacts the sequential learning performance of the model. On a practical note, quadratic programming solvers are typically iterative and implemented on the CPU, meaning the GPU bound model gradients must be transferred to the CPU, used in the optimization for \(v^\star\), and transferred back to the GPU for use in gradient descent, which is tremendously expensive when performed every batch.

\subsection{Averaged Gradient Episodic Memory}

Average Gradient Episodic Memory \citep{Chaudhry2018B} aims to solve some of the issues with Gradient Episodic Memories above by relaxing the conditions on the optimization problems. Rather than requiring that the projected gradient does not point against \textit{any} previous task's gradient, leading to one constraint per previous task, we require the projected gradient to not point against the gradient of \textit{all} previous tasks \textit{simultaneously}. That is, rather than having a gradient per previous task, we have a single gradient after running the model on all tasks. This is the \textit{average} in Averaged Gradient Episodic Memory. By reducing the number of constraints to one, we also greatly reduce the computation required to find the projected gradient \(\tilde{g}\); in fact, a closed form solution for this optimization problem exists. However, by only constraining the gradient to point in the same direction as the gradient on the combined task buffers we may allow learning that goes against the gradient for an individual task, potentially leading to forgetting, but only if the learning would benefit all previous tasks \textit{on average}.

A-GEM looks to learn a new task with the objective:
\begin{equation}
    \begin{aligned}
        \text{minimize}_{\theta}    &\qquad \mathcal{L}\left(\theta, X_\mu\right) \\
        \text{subject to}           &\qquad \mathcal{L}\left(\theta, \mathcal{M}_\mu\right) \le \mathcal{L}\left(\theta_{\mu-1}, \mathcal{M}_\nu\right) \text{ where } \mathcal{M}_\mu = \cup_{\nu<\mu} \mathcal{M}_\nu.
    \end{aligned}
\end{equation}
Note by combining the individual task buffers \(\mathcal{M}_\nu\) into a single buffer \(\mathcal{M}_\mu = \cup_{\nu<\mu} \mathcal{M}_\nu\) we enforce only a single constraint. The projected gradient is then given by:
\begin{equation}
    \label{Eqn:A-GEM Gradient Projection}
    \begin{aligned}
        \text{minimize}_{\tilde{g}} &\qquad \frac{1}{2} \lvert\lvert g - \tilde{g} \rvert\rvert _2^2 \\
        \text{subject to} &\qquad \langle \tilde{g}, g_\mu \rangle \ge 0,
    \end{aligned}
\end{equation}
where \(g_\mu\) is the gradient on the buffer \(\mathcal{M}_\mu\). This optimization problem is much simpler than the one in Equation \ref{Eqn:GEM Gradient Projection} and does not require the dual space problem and expensive quadratic programming solvers. Instead, the closed form solution of Equation \ref{Eqn:A-GEM Gradient Projection} is:
\begin{align}
    \tilde{g} = g - \frac{g^\top g_\mu}{g^\top_\mu g_\mu} g_\mu.
\end{align}
The closed form is simple enough to be implemented entirely on the GPU, avoiding the computation problems of GEM above, although we must still compute the gradient on the previous task buffers.

\subsection{L2 Regularization}

Regularization-based methods add an additional term to the loss function that approximates the previous task's loss. Usually that additional term is proportional to the squared difference of the previous task's optimal weights and the current weights, meaning large shifts from the previous optimal weights are penalized heavily. Many regularization-based methods include a measure of weight importance to allow ``unimportant'' weights to shift more than ``important'' ones. We will investigate several such regularization methods with weight importance measures, but as a sanity check we will also investigate the trivial case of assigning an equal importance to every weight. This is functionally equivalent to L2 regularization, also called weight decay.

Formally, this class of quadratic term regularization methods look like
\begin{align}
    \mathcal{L}\left(\theta\right) &= \mathcal{L}_{\text{Base}} \left(\theta\right) + \lambda \sum_{\mu} \sum_{k} \omega_{k, \mu} \left(\theta_{k, \mu}^\star - \theta_k\right)^2,
\end{align}
where \(\mathcal{L}_{\text{Base}} \left(\theta\right)\) is the base loss on the next task (e.g. mean squared error, binary cross entropy, \dots), \(\mu\) indexes over tasks, \(k\) indexes over the weights of the model, \(\omega_k\) is the importance of \(\theta_k\), and \(\theta_{k, \mu}^\star\) is the optimal value of \(\theta_k\) found at the end of the previous task. \(\lambda\) is a hyperparameter to weigh between learning the new task (low \(\lambda\)) and remembering the old task (high \(\lambda\)).  Some methods include a new quadratic term for each task completed, therefore also including another importance and optimal weight value too. In L2 regularization, we set \(\omega = 1\) for each weight, and include a new quadratic term, with optimal weights \(\theta_{k, \mu}^\star\), after each task.

\subsection{Elastic Weight Consolidation}

Elastic Weight Consolidation \citep{Kirkpatrick2017} is the first regularization-based technique we will discuss with a nontrivial weight importance measure. In EWC, weight importance is measured using the Fisher information matrix of the network parameters, effectively quantifying how each parameter affects the prediction for a sample of the training data. The presence of previous task items means that this calculation must be done either with a buffer of previous task items (in which case, rehearsal methods may be more effective) or at the end of the task (i.e. it is known the model is to undergo sequential learning). The Fisher information matrix has several useful properties for this application \citep{Pascanu2014}, the most important of which are that it can be computed from the first derivative of the loss, and it approaches the second order derivative of the base loss function near a minima. The first property allows us to calculate the Fisher information matrix efficiently, as directly computing the second derivative of a large model is often extremely expensive if not entirely intractable. The second property allows us to approximate the minima of the loss provably effectively without having to directly measure the second derivative.

Over a sample of training data the Fisher information matrix, and hence the weight importance, is approximated by:
\begin{align}
    \omega_k &= \left(\frac{\partial}{\partial \theta_k} \mathcal{L}_{\text{Base}} \left(\theta\right) \right)^2.
\end{align}
Aich \citep{Aich2021} has a good discussion and formalization of the derivation of this approximation.

The loss function on future tasks is defined by:
\begin{align}
    \mathcal{L}\left(\theta\right) &= \mathcal{L}_{\text{Base}} \left(\theta\right) + \frac{\lambda}{2} \sum_{\mu} \sum_{k} \omega_{k,\mu} \left(\theta_{k,\mu}^\star - \theta_k\right)^2.
\end{align}
Note in EWC we have a quadratic term for each completed task, indexed by \(\mu\), rather than a single term that is carried through. The additional factor of \(\frac{1}{2}\) on the \(\lambda\) is included for consistency with the original literature.

\subsection{Memory Aware Synapses}

Memory Aware Synapses \citep{Aljundi2018} is another weight importance regularization-based sequential learning technique. In contrast to EWC, which uses the Fisher information and the loss function of the task, MAS measures the model sensitivity directly. A Taylor expansion of the model output with respect to each network parameter gives a weight importance measure of:
\begin{align}
    \omega_k &= \frac{1}{N} \sum_{i=1}^{N} \Bigl \lvert\Bigl \lvert \frac{\partial F\left(X, \theta\right)}{\partial \theta_k} \Bigr \rvert\Bigr \rvert,
\end{align}
again for a sample of the training data \(X\). Note that the network output \(F\left(X, \theta\right)\) itself is used, not the loss function \(\mathcal{L}_{\text{Base}} \left(\theta\right)\), and that the training data \(X\) is not indexed to a task \(\mu\), as we are using only the most recent task data to calculate the weight importances. The partial derivative of \(F\) with respect to each possible network output is taken and averaged, indicated by the sum over \(N\) (e.g. for a task with \(10\) outputs / classes, \(N=10\)). The value of \(\omega_k\) is accumulated over each task, so we keep some information about previous task's weight importances without requiring a new term per task, as in EWC.

\citeauthor{Aljundi2018} note a slightly more efficient implementation of this technique for models with many outputs. Rather than taking the derivative of each output with respect to the model parameters, i.e. one expansion per output, instead take the Taylor expansion of the squared L2 norm of the model output as a vector. This results in only a single expansion and hence much less computation for the same result.

The modified loss used by each is task is:
\begin{align}
    \mathcal{L}\left(\theta\right) &= \mathcal{L}_{\text{Base}} \left(\theta\right) + \lambda \sum_{k} \omega_{k} \left(\theta_{k}^\star - \theta_k\right)^2.
\end{align}
Unlike L2 Regularization and EWC, the single tensor of optimal weights \(\theta_{k}^\star\) is updated with each task, as are the weight importances \(\omega_{k}\).

\subsection{Synaptic Intelligence}

Synaptic Intelligence \citep{Zenke2017} is a sequential learning technique again focused around a quadratic regularization term with weight importances. The importance measure is now based on the path integral of the task's gradient along the path through parameter space taken during training. Formally:

\begin{align}
    \omega_{k, \mu} &= \int_{t_{\mu-1}}^{t_{\mu}} \frac{\partial \mathcal{L}_{\text{Base}}(\theta(t)) }{\partial \theta_k(t)} \frac{\mathrm{d} \theta_k(t)}{\mathrm{d} t} \mathrm{d}t,
\end{align}
where again, the task is indexed by \(\mu\), and the model parameters \(\theta\) dependence on time represents the changes made by training the model. Here, \(t_\mu\) indicates the time step (be that in batches or epochs) when we finish training on task \(\mu\), so the interval \([t_{\mu-1}, {t_{\mu}}]\) represents the training on task \(\mu\). In practice, \citeauthor{Zenke2017} approximate the continuous path integral with steps taken each epoch or batch:
\begin{align}
    \omega_{k, \mu} &= \sum_{t = t_{\mu-1}}^{t_{\mu}} \frac{\partial \mathcal{L}_{\text{Base}}(\theta(t)) }{\partial \theta_k(t)} \left(\theta_k(t) - \theta_k(t-1)\right).
\end{align}
At the end of each task, the total displacement of each parameter over the course of learning that task is computed
\begin{align}
    \Delta_{k, \mu} &= \theta_k(t_\mu) - \theta_k(t_{\mu-1}).
\end{align}
which is combined with the weight importances of all previous tasks into a single term,
\begin{align}
    \Omega_{k,\mu} &= \sum_{\nu<\mu} \frac{\omega_{k, \nu}}{\left(\Delta_{k, \nu}\right)^2 + \epsilon},
\end{align}
where \(\epsilon\) is a small constant to aid in numerical stability. The combined importance measure is then used in a single quadratic surrogate term for task \(\mu\):
\begin{align}
    \mathcal{L}\left(\theta\right) &= \mathcal{L}_{\text{Base}} \left(\theta\right) + \lambda \sum_{k} \Omega_{k,\mu} \left(\theta_{k,\mu}^\star - \theta_k\right)^2
\end{align}

Synaptic Intelligence is similar to Memory Aware Synapses in that the optimal weights \(\theta_{k,\mu}^\star\) are updated every task, unlike Elastic Weight Consolidation which has a new set of optimal weights for each task. This can help with memory overhead, particularly for very large models, however it also means we lose some information on previous tasks. Unlike Memory Aware Synapses, which accumulates all importance measures from previous tasks into a single matrix, Synaptic Intelligence holds information about previous tasks separately, and combines these measures into an aggregate measure \(\Omega_{\mu}\) weighted by the total displacement of each weight in a task. We can compare these three similar methods by observing that Elastic Weight Consolidation uses the most information about previous tasks (as all weight importances are introduced in separate quadratic terms), Synaptic Intelligence uses less (combining each weight importance into one quadratic term, but weighted by the total displacement), and Memory Aware Synapses less information again (simply summing the importances of each task). L2 Regularization is the baseline method, with zero information on the weight importances or tasks. However, information quantity alone is not a good evaluation of these methods, as the derivation of the weight importances is very different between these methods and may result in better performance with less information.

\section{Sequential Learning Datasets}
\label{Appendix: Datasets}

There are a variety of sequential learning datasets that are standard across the literature. A decent amount of sequential learning has focused on classification tasks, as these networks were and are prevalent in the current age of machine learning. Some such datasets have been based around well known, widely available datasets such as MNIST digit classification \citep{LeCun1998} and CIFAR \citep{Krizhevsky2009}. These are image based classification datasets which proved popular during the rise of image classifications models after AlexNet \citep{Krizhevsky2012}. To convert these to sequential learning datasets, the original data is split into a series of tasks. The degree to which different tasks are related to one another may vary depending on what the aspect of learning is being focused on; for example, forward and backwards transfer may want tasks that are highly related, while capacity investigations may want entirely unrelated tasks. Common MNIST variations include Rotated MNIST (in which images are rotated by some amount, disrupting the data very little and leading to highly related tasks) and Permuted MNIST \citep{Kirkpatrick2017} (in which the pixels of images belonging to one task are permuted some amount, leading to entirely unrelated tasks, effectively an entire new MNIST dataset for each task). Another common transformation is to take only some of the classes of the dataset in each task. This is commonly seen in Split CIFAR-10 and Split CIFAR-100 \citep{Zenke2017}, as the structure of the CIFAR dataset does not lend itself to permutation without the tasks being extremely large. Split MNIST \citep{Goodfellow2015} is also well known.

Recently, perhaps in tandem with growing hardware power, other datasets have become more popular in a sequential learning environment. CUB \citep{Welinder2010} and AWA \citep{Lampert2009} have both seen use in sequential learning \citep{Chaudhry2019} as Split CUB and Split AWA. With the rise of transformers, attention, and Large Language Models, sequential learning over natural language datasets has also become common \citep{Houlsby2019, Douillard2022, Li2022}. Again, these datasets are usually derived from well known natural language datasets, although with more of a focus on classification tasks. Examples of NLP sequential learning datasets are Split Reuters (news article classification) \citep{Lewis2004}, Split Wiki-30K (Wikipedia article classification) \citep{Zubiaga2012}, and Split Arxiv (Arxiv paper classification) \citep{Yang2018}. 

Although the DAM has been generalized to take continuous values \citep{HopfieldIsAllYouNeed2021}, we are looking to investigate sequential learning in the most basic of the network architectures. This will allow us to measure if the adjustment of the interaction vertex alone impacts sequential learning, without possible interference of other factors such as the alterations that allow for the continuous network. Therefore, we will use the binary-valued DAM as first proposed by Krotov and Hopfield \citep{KrotovHopfield2016} with modifications that provably do not alter the original network properties \citep{McAlister2024B}. This paper will focus on binary datasets (or datasets that can be thresholded without loss of information), such as Permuted MNIST.

\section{Sequential Learning Example}
\label{Appendix: Sequential Learning Example}

To give a clarifying example of sequential learning in the Hopfield network, consider a simple problem of two tasks, each of some random states:
\begin{align*}
    \bar{\xi}_1 = \{\xi_{1,1}, \xi_{1,2}, \dots, \xi_{1,k}\},\\
    \bar{\xi}_2 = \{\xi_{2,1}, \xi_{2,2}, \dots, \xi_{2,k}\}. 
\end{align*}
For a sufficiently large Hopfield network such that the capacity is not exceeded \citep{Hopfield1982,Hertz1991}, the network is capable of learning both tasks if presented together. However, we present the tasks sequentially. After presenting the first task the network will form strong attractors for only the first task states, meaning that probes that are close to a learned state will iterate towards the nearby attractor, and the network will recall the learned states correctly. We can say that \(\xi_{1,i} + \delta \to \xi_{1,i} \; \forall i\) for small \(\lVert\delta\rVert\). Then, keeping the model parameters initialized to those learned from the first task, we present the second task. The network is capable of learning the second task just as well as the first, so \(\xi_{2,i} + \delta \to \xi_{2,i} \; \forall i\) after task two, but we did not present any items from the first task during the latest training. Therefore, we cannot make any guarantees about the stability of the first task attractors. It is possible that the second task induced minimal disruption and the first task attractors are still present and strong, but more likely is that the first task attractors are diminished and replaced by the second task. For example, after training on the second task we may find that a probe \(\xi_{1,1} \to \xi_{2,5}\), that is even the learned states of the first task themselves are unstable. The exact indices of the states here are not important, and only used as an example. What is important is that the attractor that once existed for \(\xi_{1,1}\) is now gone. We say the network has forgotten the first task, or that it has suffered from catastrophic forgetting. Learning a third task would cause catastrophic forgetting of the second, and further forgetting of the first. The degree to which the first task items are disrupted is a measure of that forgetting --- if the model forgets all attractors, that is worse than only forgetting half. The aim of sequential learning methods is to alleviate the impact of forgetting while still allowing enough plasticity to learn new tasks in full.

\section{General Dense Associative Memory Hyperparameter Search}
\label{Appendix: General Hyperparameter Search}

We conducted a hyperparameter search over the general network hyperparameters. Our aim is to find the combination of hyperparameters that maximizes the minimum accuracy on these tasks. This ensures our network can not only learn the first task well (overcoming initialization) but also remain plastic enough to learn subsequent tasks. We present a select few of these gridsearches that performed best, searching over the initial learning rate and the temperature. We repeated these experiments to tune the learning rate decay and momentum values, although the results are excluded for brevity. Note that the optimal regions in the searches below do not move considerably as we increase the interaction vertex, something attributed to modifications to the Dense Associative Memory \citep{McAlister2024B}. We also include the average accuracy over the same gridsearch for completeness. We determined the optimal general network hyperparameters to be a number of memories \(|\bar{\zeta}|=512\), a number of training epochs \(\text{MaxEpochs} = 500\), an initial learning rate \(\text{lrInit} = 1\times10^{-1}\), learning rate decay \(\text{lrDecay} = 0.999\), momentum \(p = 0.6\), an initial and final temperature value \(\frac{1}{\beta} = T_{i} = T_{f} = 0.875\), and an error exponent \(m = 1\).

\begin{figure}[H]
    \centering
    \begin{subfigure}[t]{0.48\textwidth}
        \includegraphics[width=\textwidth]{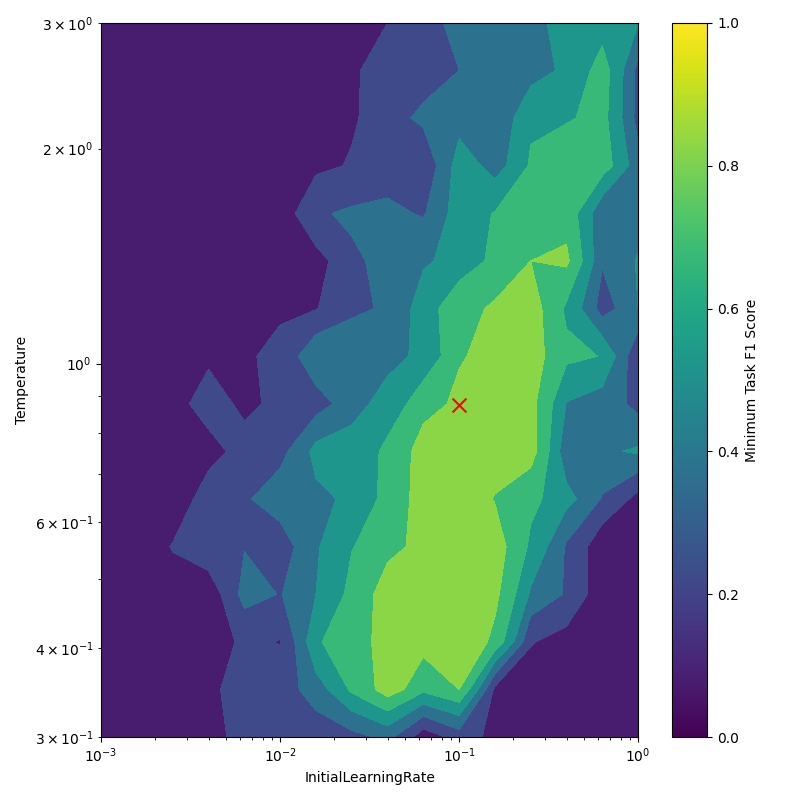}
        \caption{\(n=2\)}
    \end{subfigure}
    \begin{subfigure}[t]{0.48\textwidth}
        \includegraphics[width=\textwidth]{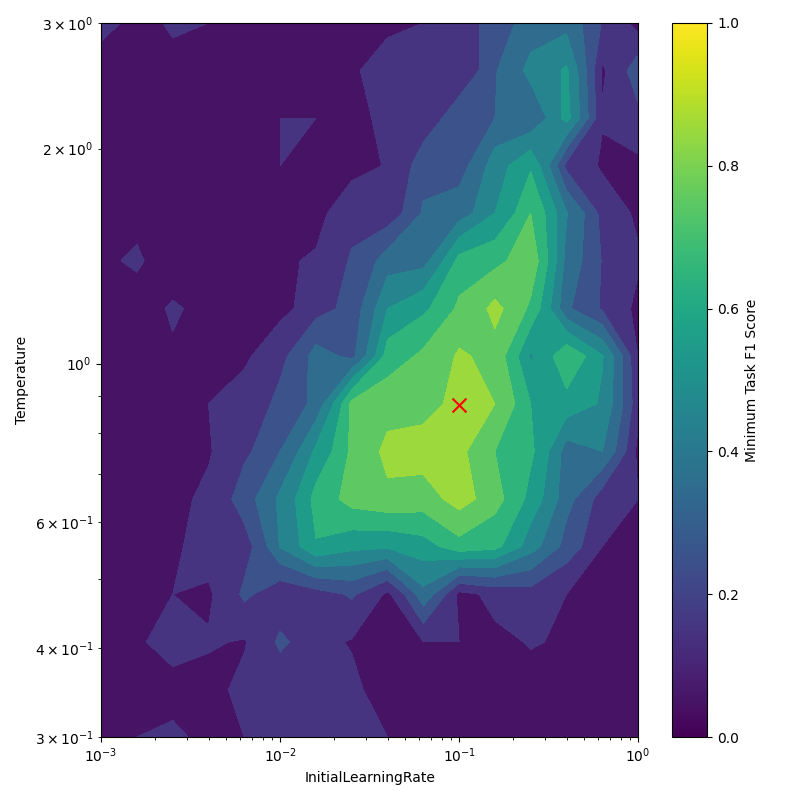}
        \caption{\(n=5\)}
    \end{subfigure}
    \hfill
    \begin{subfigure}[t]{0.48\textwidth}
        \includegraphics[width=\textwidth]{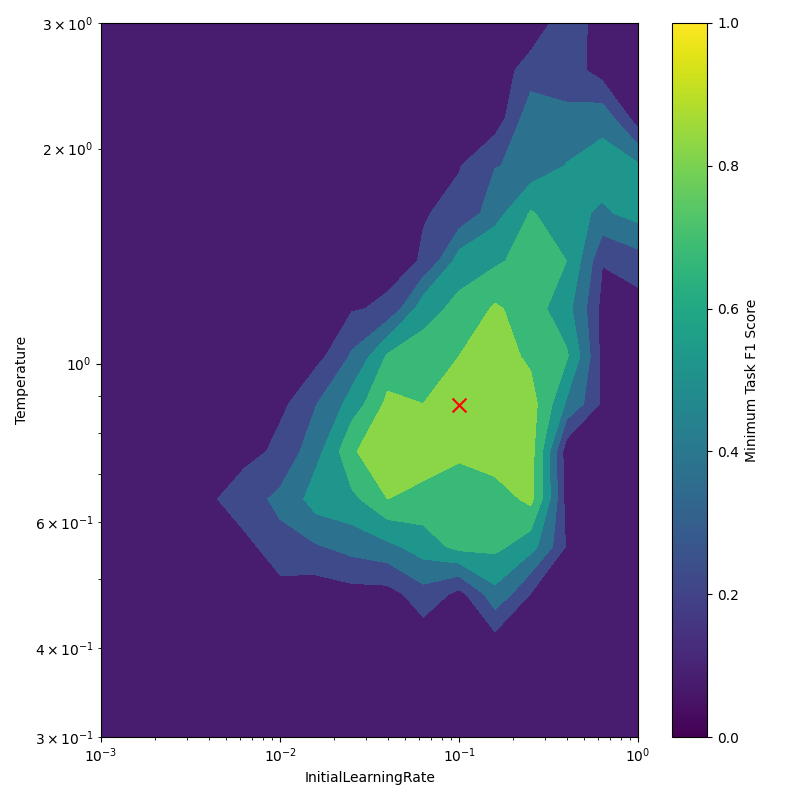}
        \caption{\(n=10\)}
    \end{subfigure}
    \begin{subfigure}[t]{0.48\textwidth}
        \includegraphics[width=\textwidth]{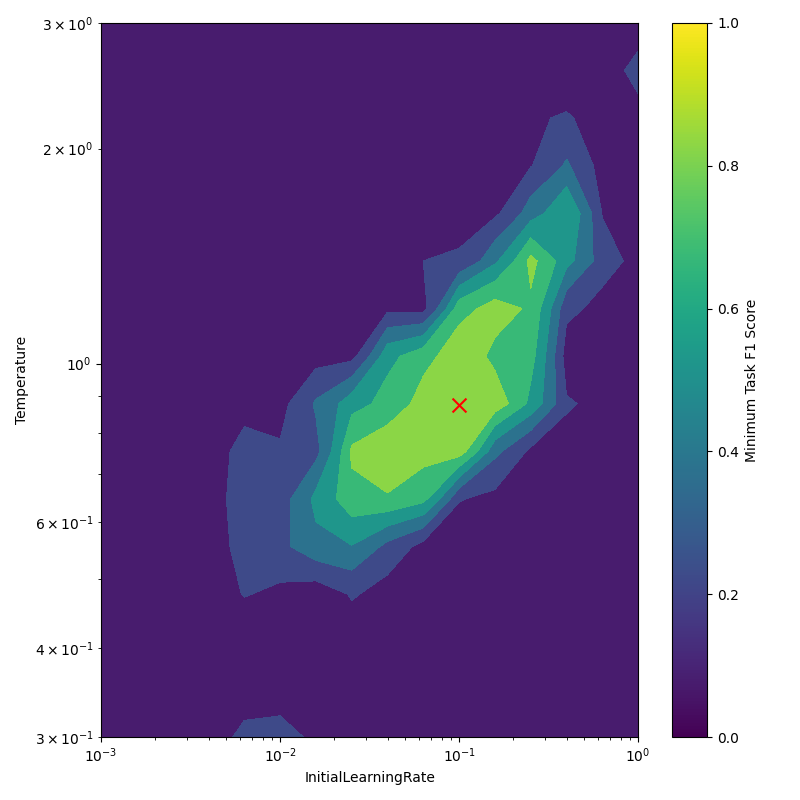}
        \caption{\(n=20\)}
    \end{subfigure}
    \caption{General network hyperparameter search on five Permuted MNIST tasks, measuring the minimum F1 score on the test data. A larger minimum F1 score corresponds to a better performing network. The red cross indicates our choice of hyperparameters \(\text{lrInit} = 1\times10^{-1}\) and \(\frac{1}{\beta} = T_{i} = T_{f} = 0.875\).}
    \label{Fig: General Hyperparameter Search Worst Task F1 Score}
\end{figure}

\begin{figure}[H]
    \centering
    \begin{subfigure}[t]{0.48\textwidth}
        \includegraphics[width=\textwidth]{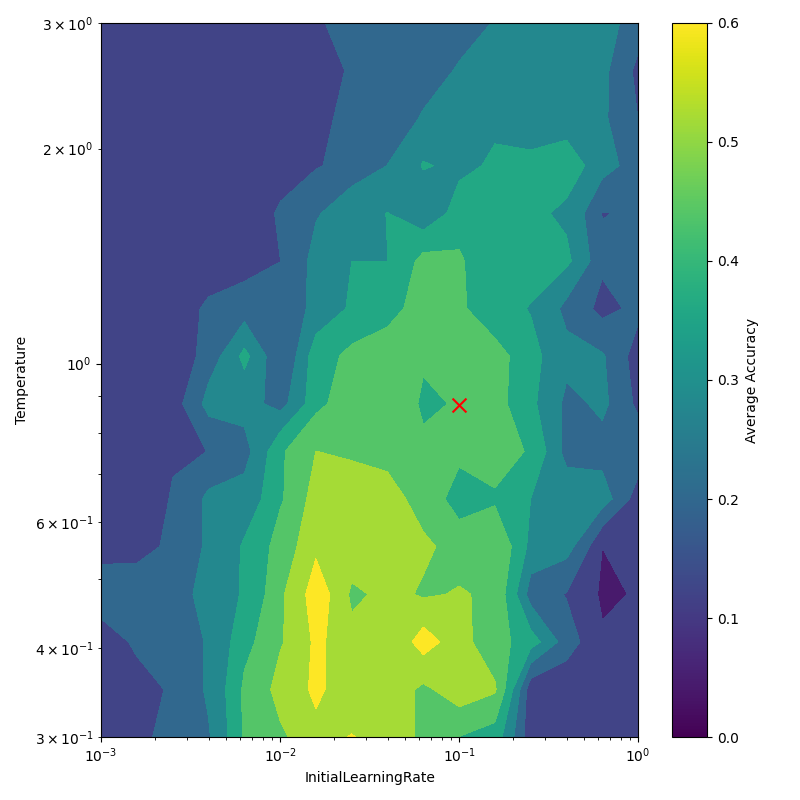}
        \caption{\(n=2\)}
    \end{subfigure}
    \begin{subfigure}[t]{0.48\textwidth}
        \includegraphics[width=\textwidth]{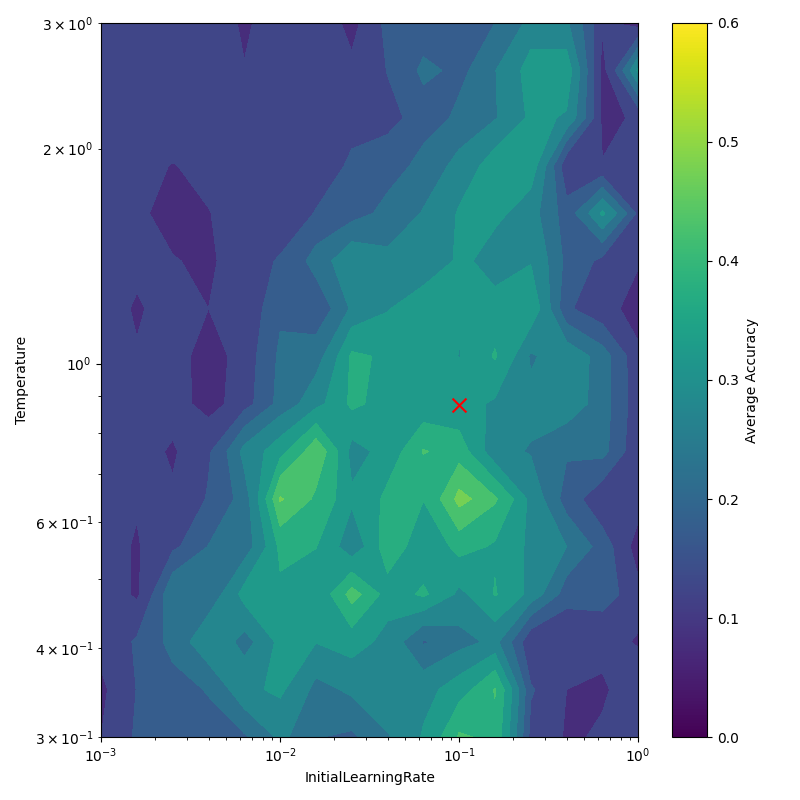}
        \caption{\(n=5\)}
    \end{subfigure}
    \hfill
    \begin{subfigure}[t]{0.48\textwidth}
        \includegraphics[width=\textwidth]{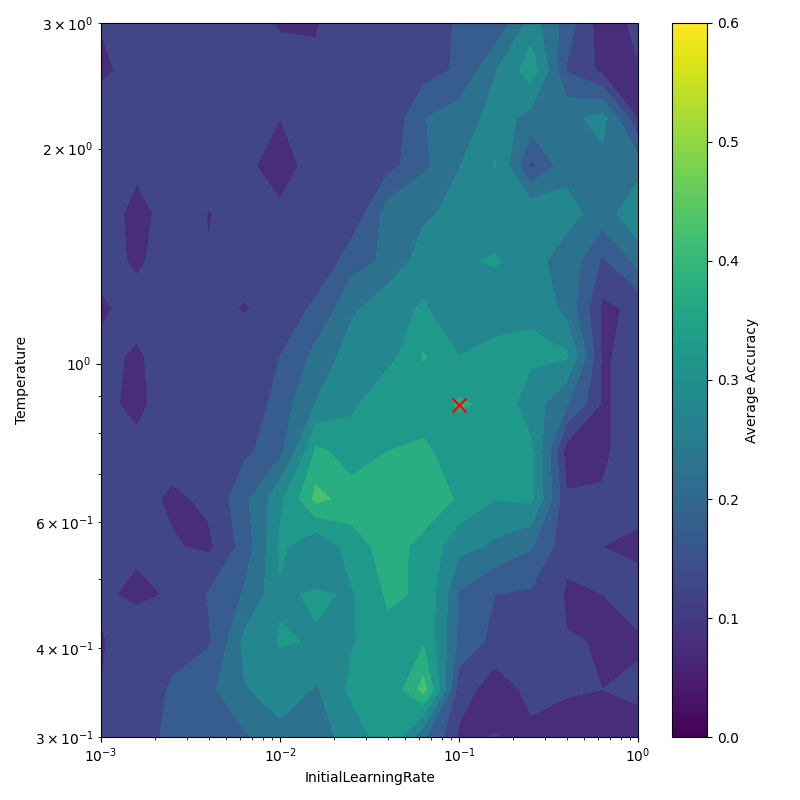}
        \caption{\(n=10\)}
    \end{subfigure}
    \begin{subfigure}[t]{0.48\textwidth}
        \includegraphics[width=\textwidth]{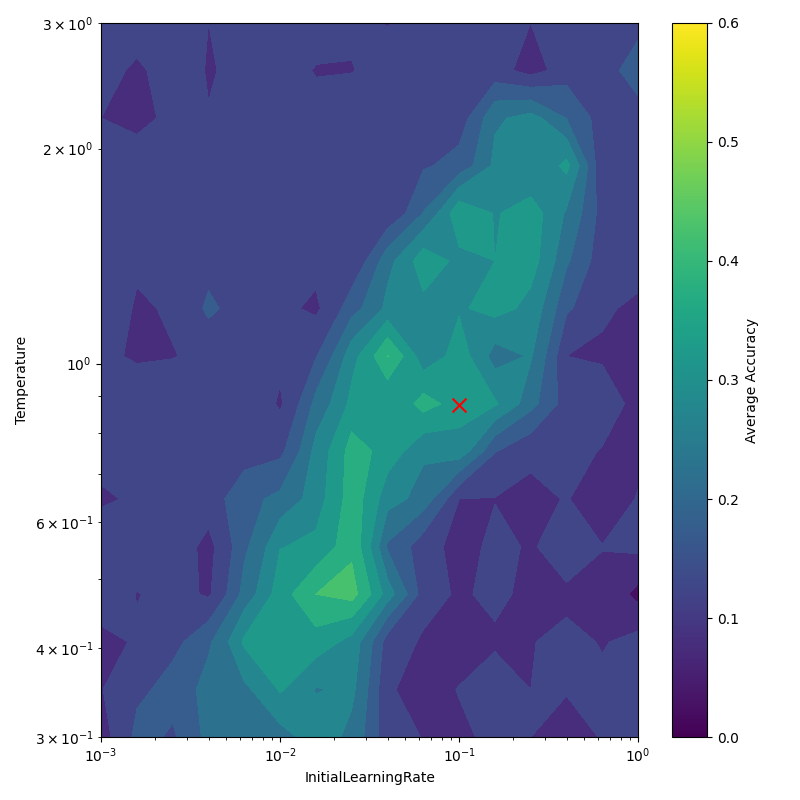}
        \caption{\(n=20\)}
    \end{subfigure}
    \caption{General network hyperparameter search on five Permuted MNIST tasks, measuring the average accuracy on the test data. A larger average accuracy corresponds to a better performing network. The red cross indicates our choice of hyperparameters \(\text{lrInit} = 1\times10^{-1}\) and \(\frac{1}{\beta} = T_{i} = T_{f} = 0.875\).}
    \label{Fig: General Hyperparameter Search Average Accuracy}
\end{figure}

\section{Task Performances by Sequential Learning Method}
\label{Appendix: Appendix Task Performances}

\begin{figure}[H]
    \centering
    \begin{subfigure}[t]{0.43\textwidth}
        \centering
        \includegraphics[width=\textwidth]{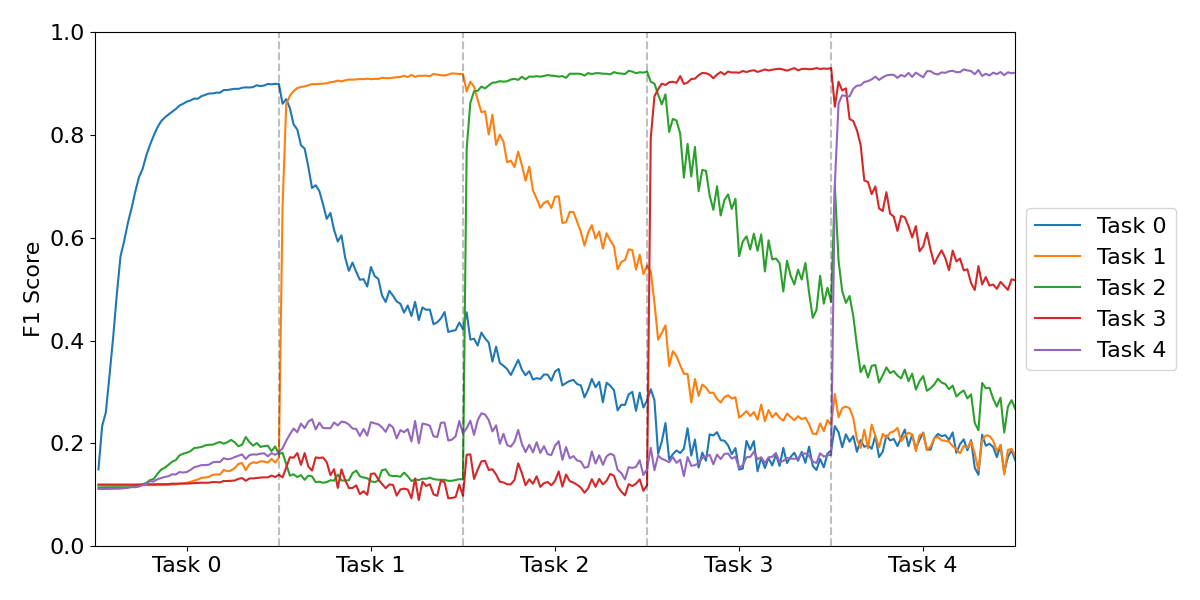}
        \caption{\(n=2\)}
    \end{subfigure}
    \begin{subfigure}[t]{0.43\textwidth}
        \centering
        \includegraphics[width=\textwidth]{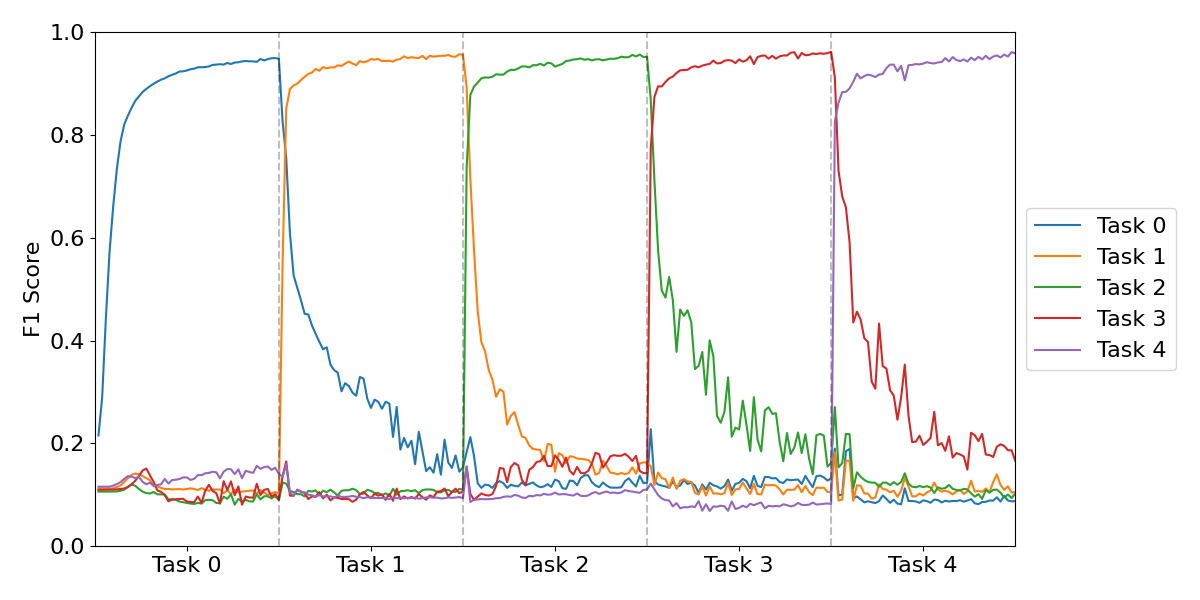}
        \caption{\(n=5\)}
    \end{subfigure}
    \hfill  
    \begin{subfigure}[t]{0.43\textwidth}
        \centering
        \includegraphics[width=\textwidth]{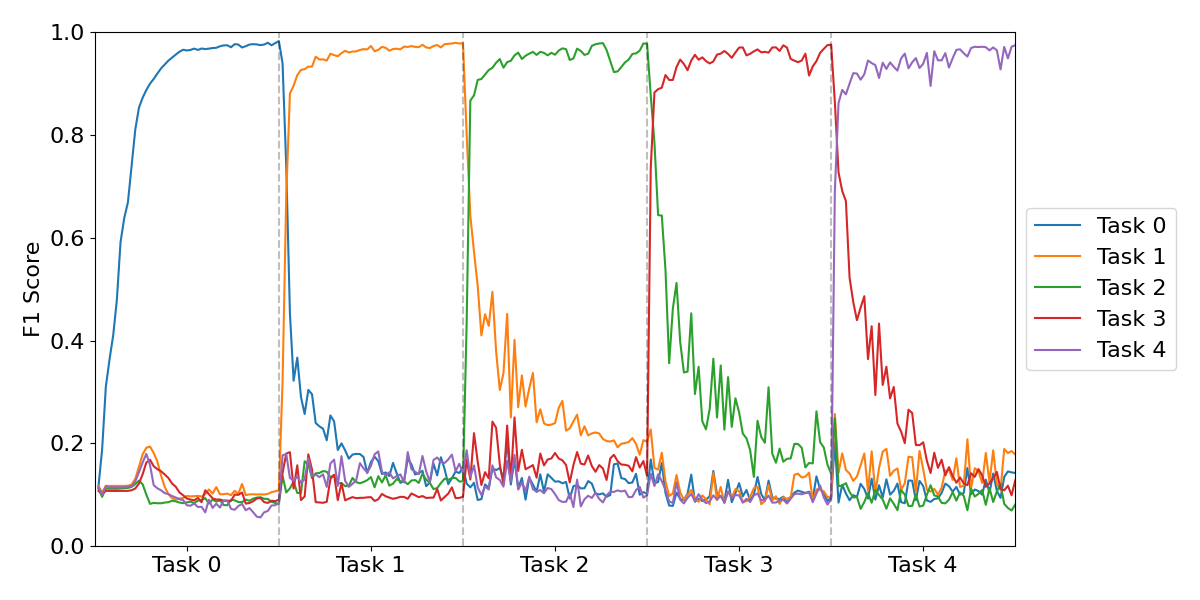}
        \caption{\(n=10\)}
    \end{subfigure}
    \begin{subfigure}[t]{0.43\textwidth}
        \centering
        \includegraphics[width=\textwidth]{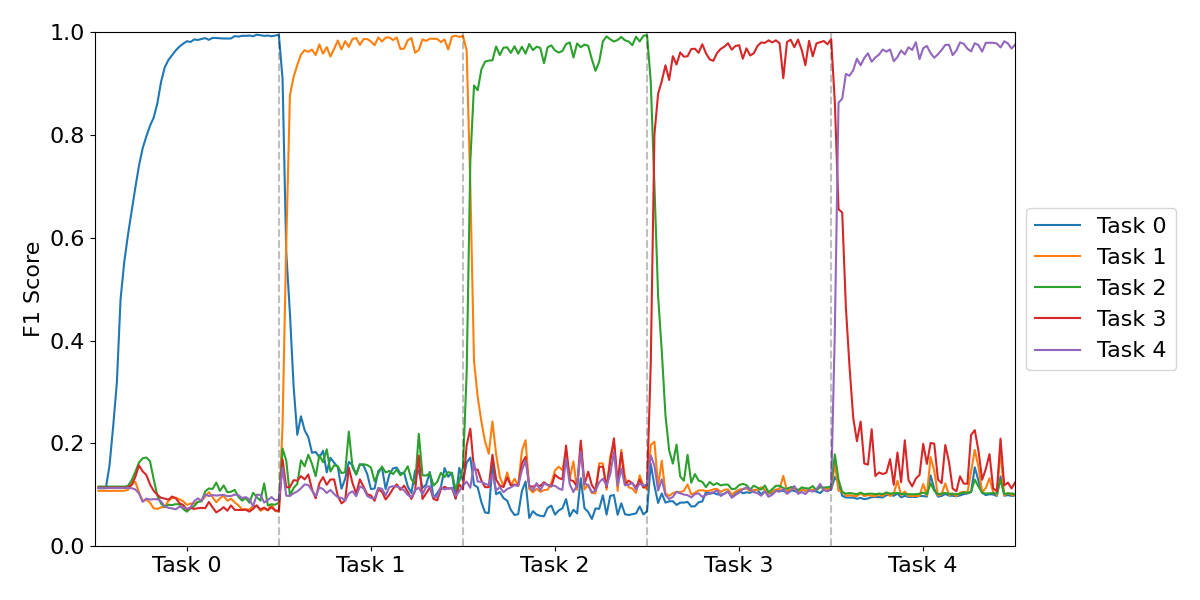}
        \caption{\(n=20\)}
    \end{subfigure}
    
    \caption{Task Performances for Vanilla Sequential Learning}
\end{figure}

\begin{figure}[H]
    \centering
    \begin{subfigure}[t]{0.43\textwidth}
        \centering
        \includegraphics[width=\textwidth]{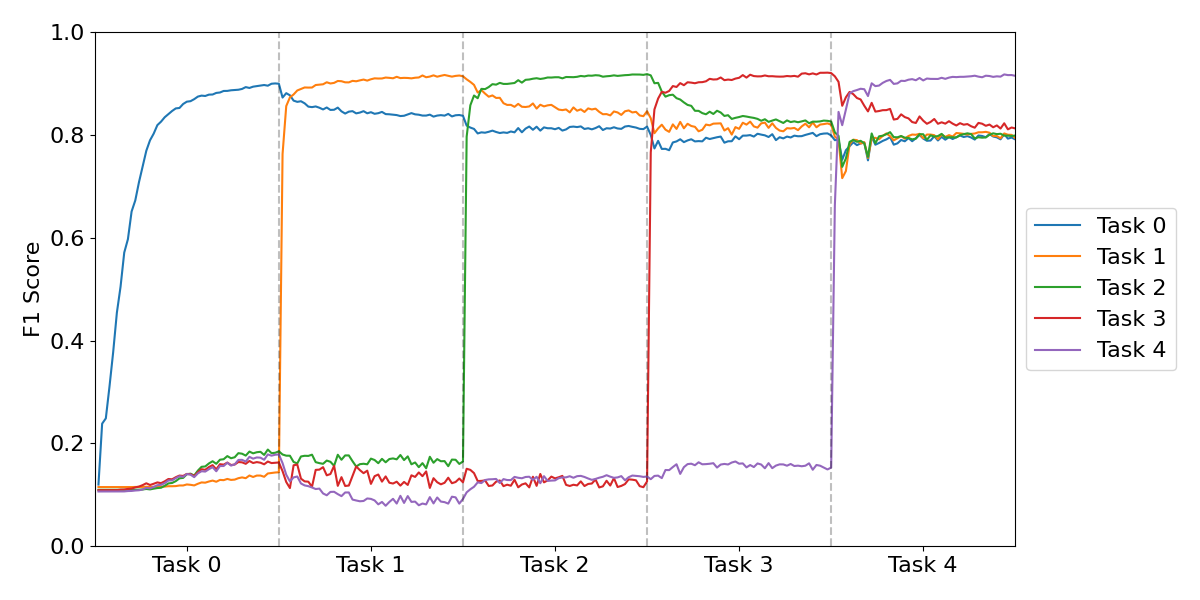}
        \caption{\(n=2\)}
    \end{subfigure}
    \begin{subfigure}[t]{0.43\textwidth}
        \centering
        \includegraphics[width=\textwidth]{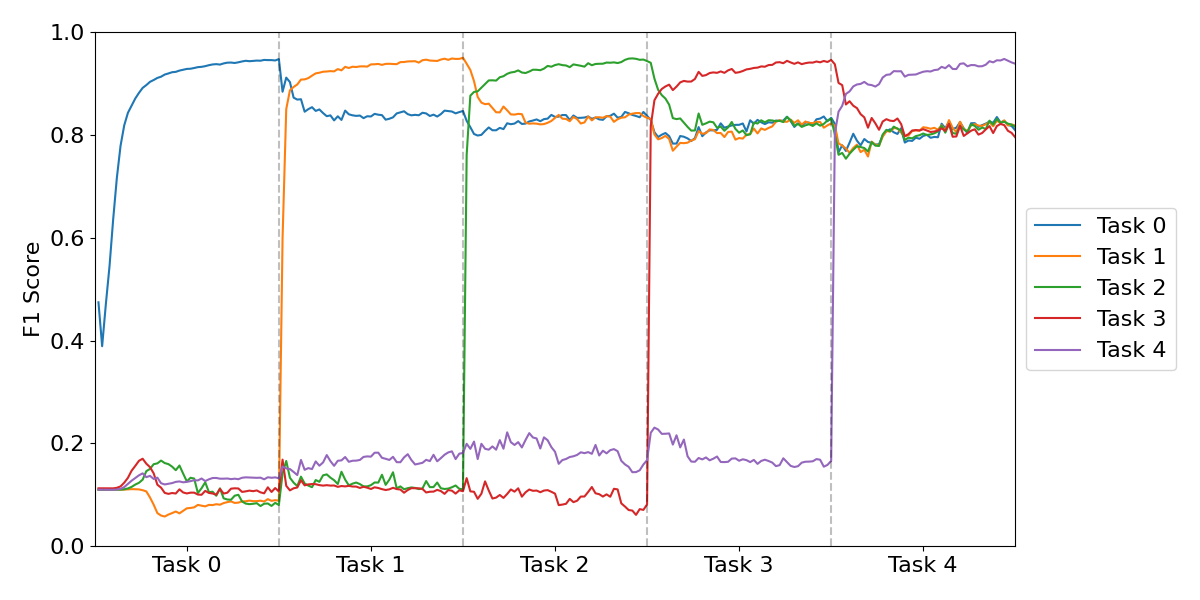}
        \caption{\(n=5\)}
    \end{subfigure}
    \hfill  
    \begin{subfigure}[t]{0.43\textwidth}
        \centering
        \includegraphics[width=\textwidth]{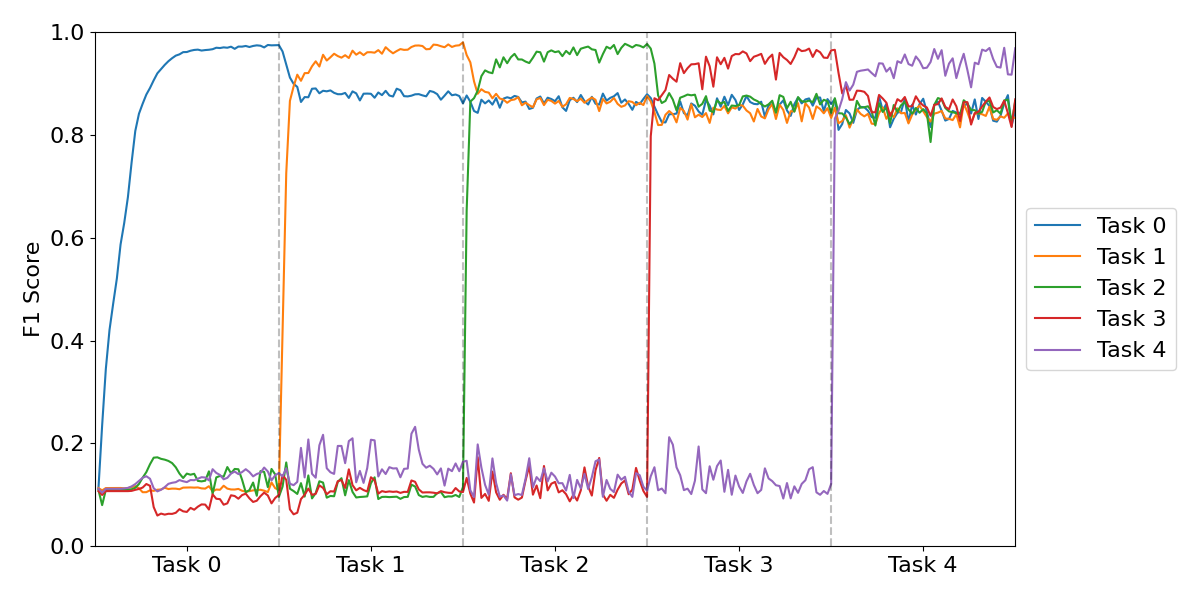}
        \caption{\(n=10\)}
    \end{subfigure}
    \begin{subfigure}[t]{0.43\textwidth}
        \centering
        \includegraphics[width=\textwidth]{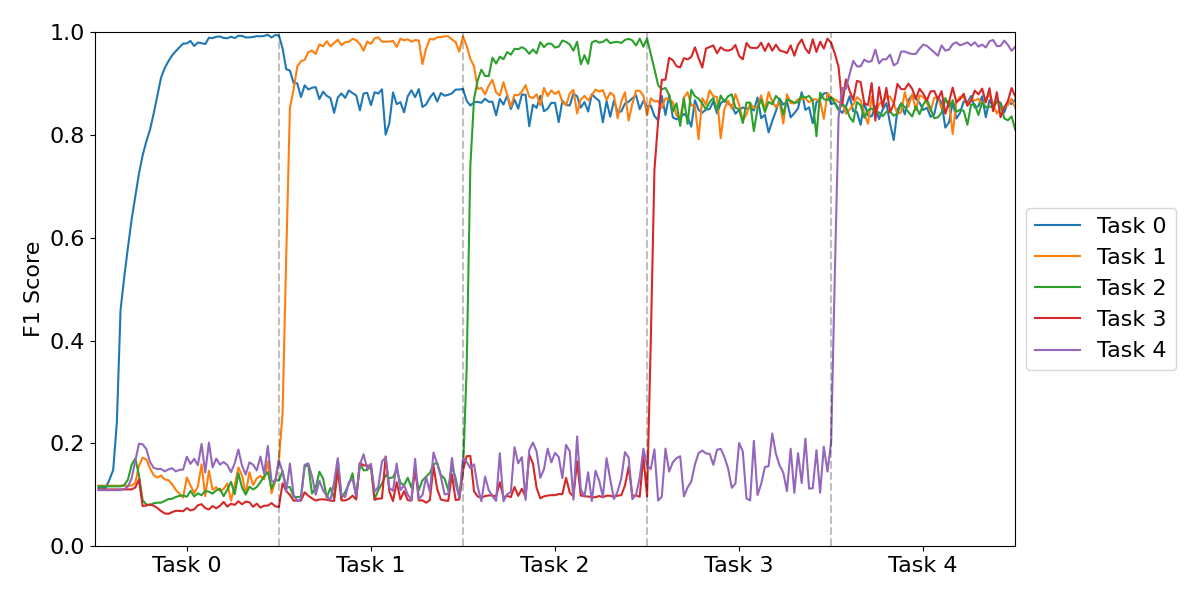}
        \caption{\(n=20\)}
    \end{subfigure}
    
    \caption{Task Performances for Rehearsal using a Rehearsal Proportion of \(0.05\).}
\end{figure}

\begin{figure}[H]
    \centering
    \begin{subfigure}[t]{0.43\textwidth}
        \centering
        \includegraphics[width=\textwidth]{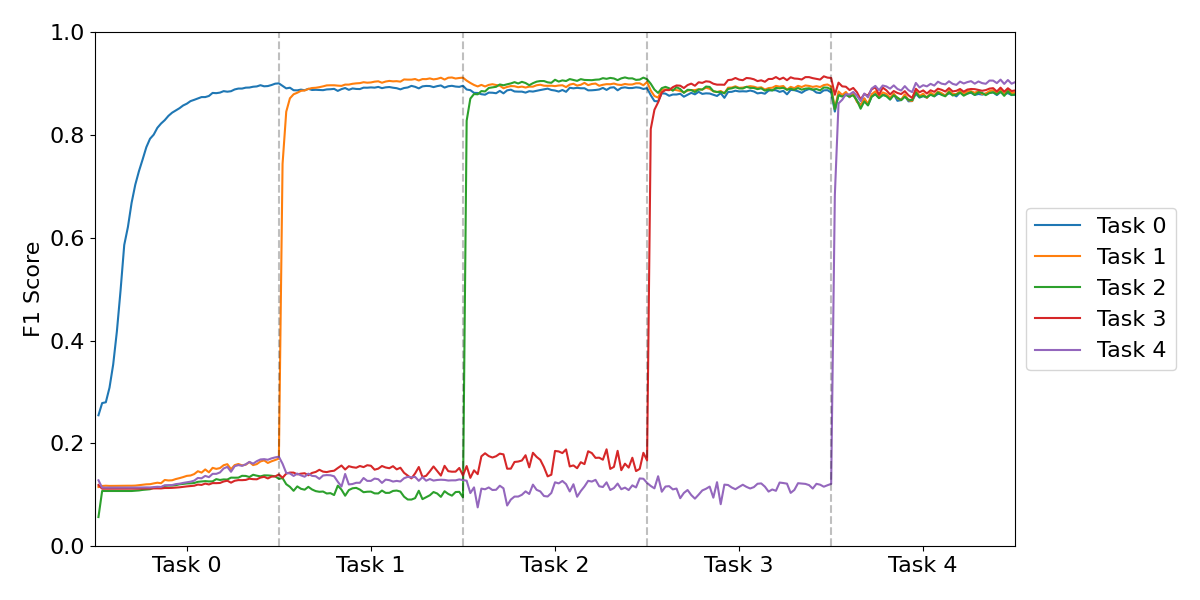}
        \caption{\(n=2\)}
    \end{subfigure}
    \begin{subfigure}[t]{0.43\textwidth}
        \centering
        \includegraphics[width=\textwidth]{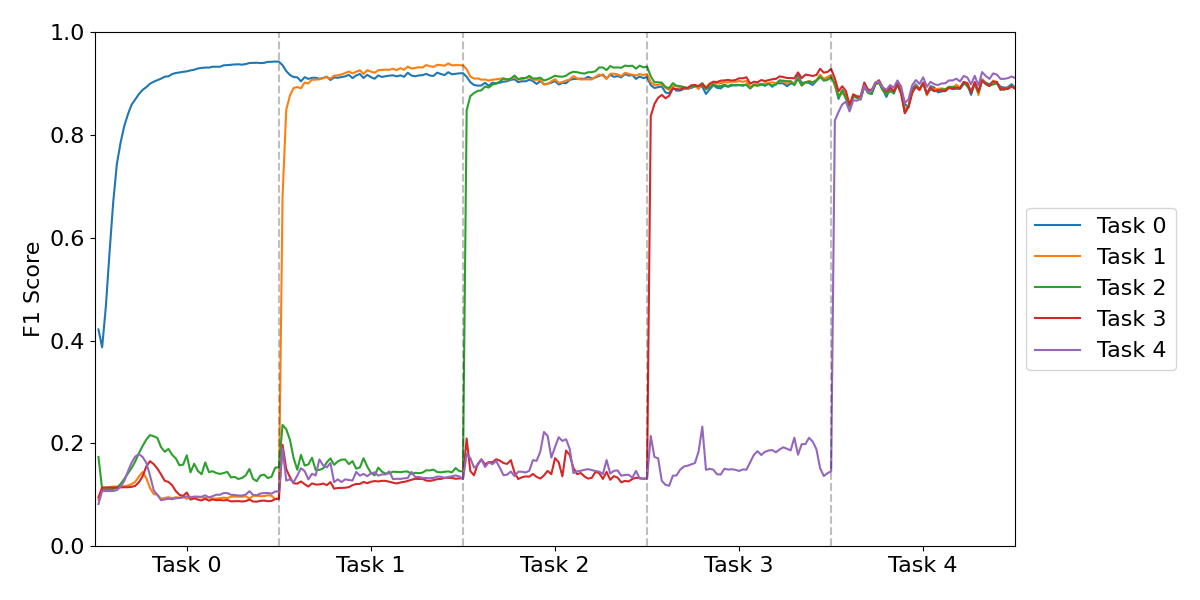}
        \caption{\(n=5\)}
    \end{subfigure}
    \hfill  
    \begin{subfigure}[t]{0.43\textwidth}
        \centering
        \includegraphics[width=\textwidth]{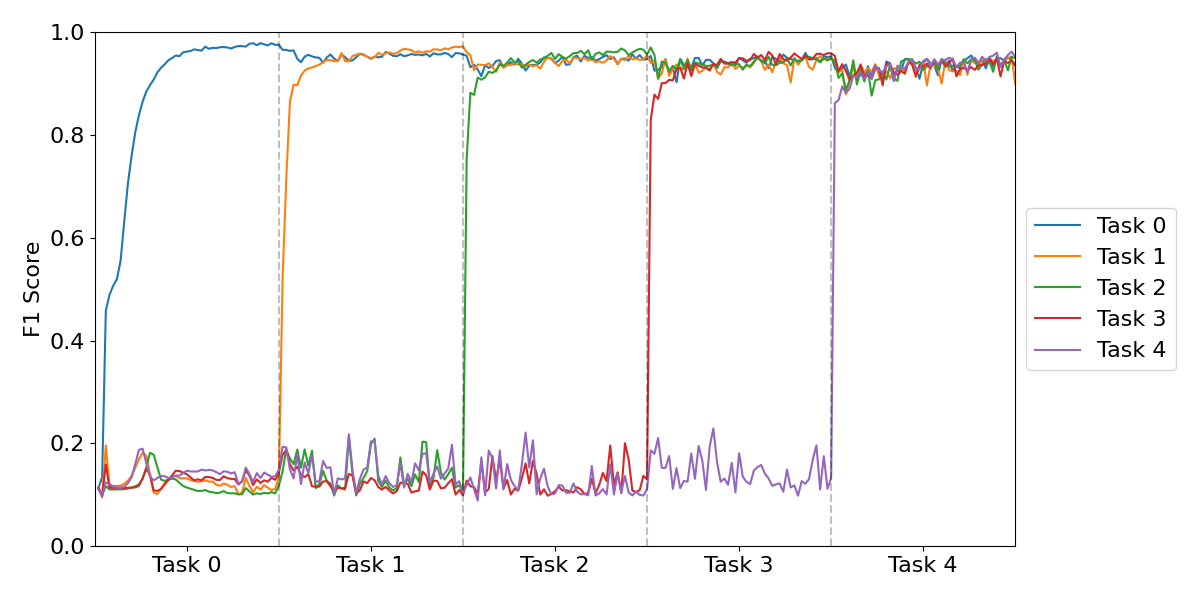}
        \caption{\(n=10\)}
    \end{subfigure}
    \begin{subfigure}[t]{0.43\textwidth}
        \centering
        \includegraphics[width=\textwidth]{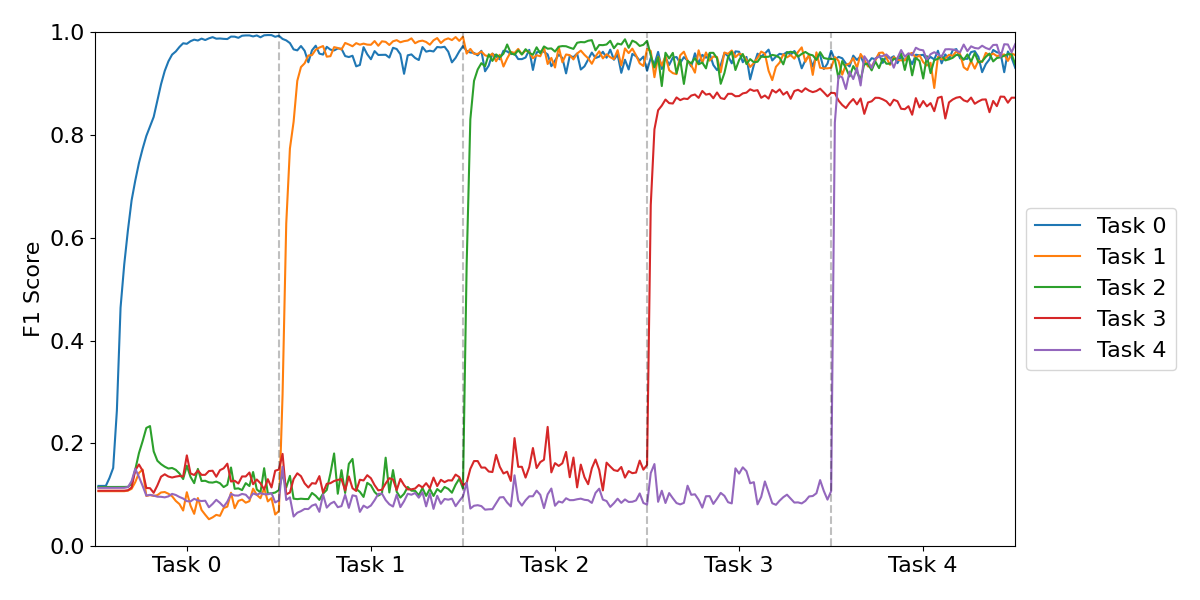}
        \caption{\(n=20\)}
    \end{subfigure}
    
    \caption{Task Performances for Rehearsal using a Rehearsal Proportion of \(0.5\).}
\end{figure}

\begin{figure}[H]
    \centering
    \begin{subfigure}[t]{0.43\textwidth}
        \centering
        \includegraphics[width=\textwidth]{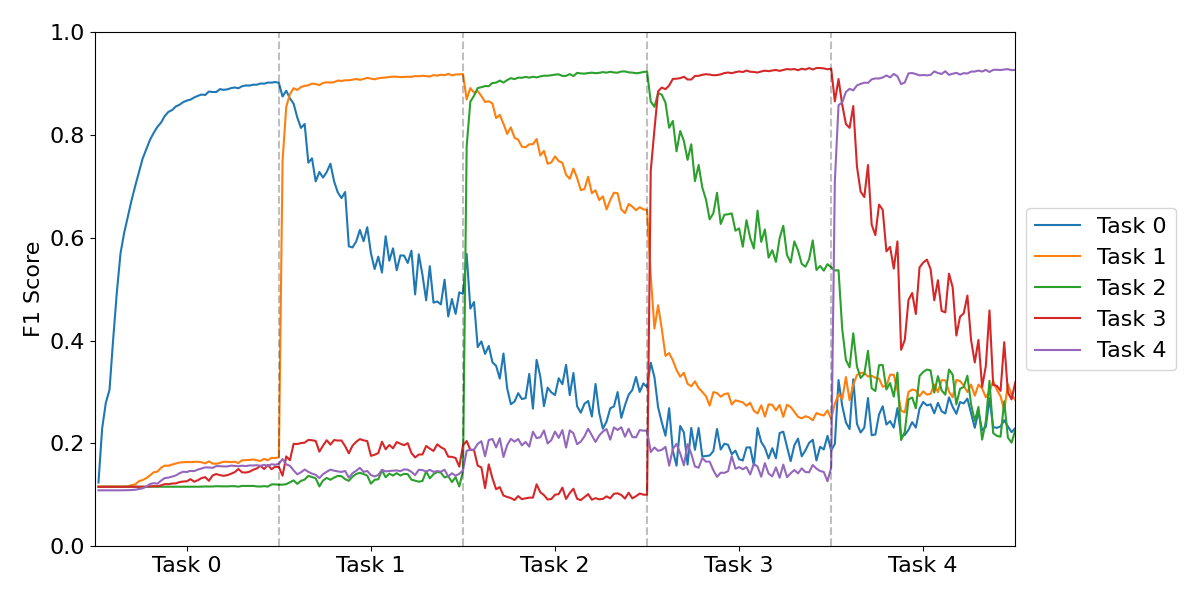}
        \caption{\(n=2\)}
    \end{subfigure}
    \begin{subfigure}[t]{0.43\textwidth}
        \centering
        \includegraphics[width=\textwidth]{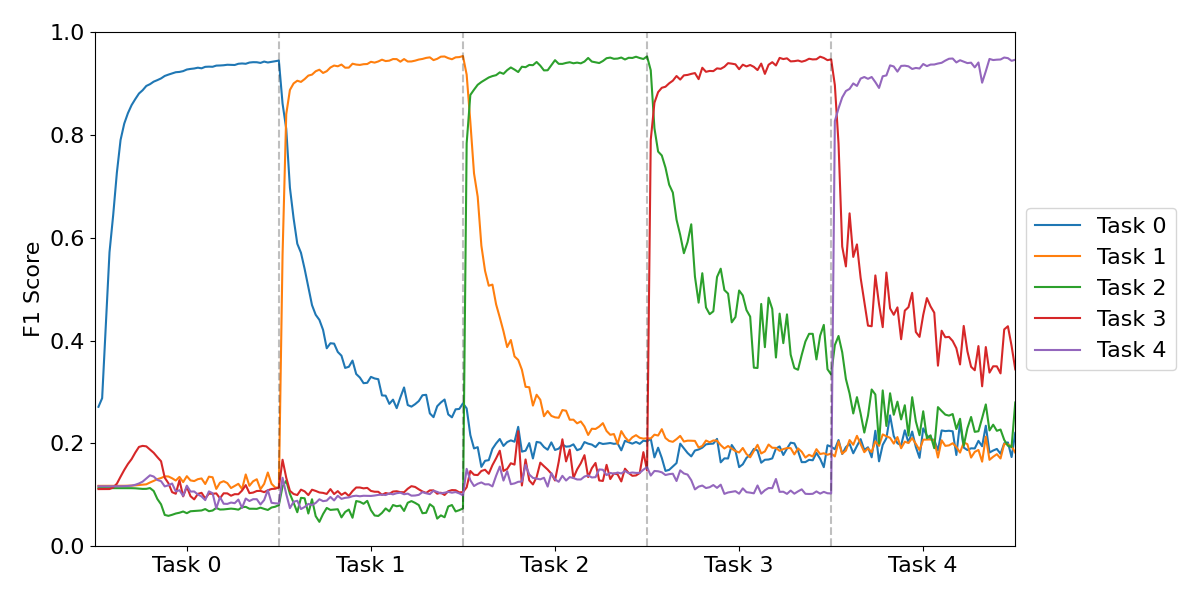}
        \caption{\(n=5\)}
    \end{subfigure}
    \hfill  
    \begin{subfigure}[t]{0.43\textwidth}
        \centering
        \includegraphics[width=\textwidth]{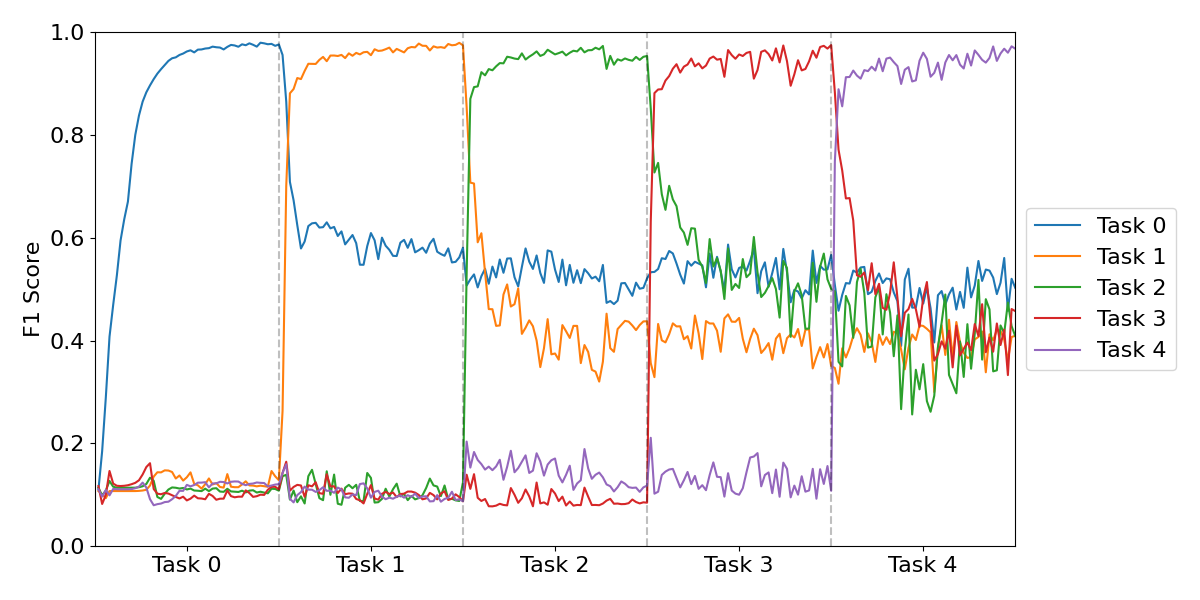}
        \caption{\(n=10\)}
    \end{subfigure}
    \begin{subfigure}[t]{0.43\textwidth}
        \centering
        \includegraphics[width=\textwidth]{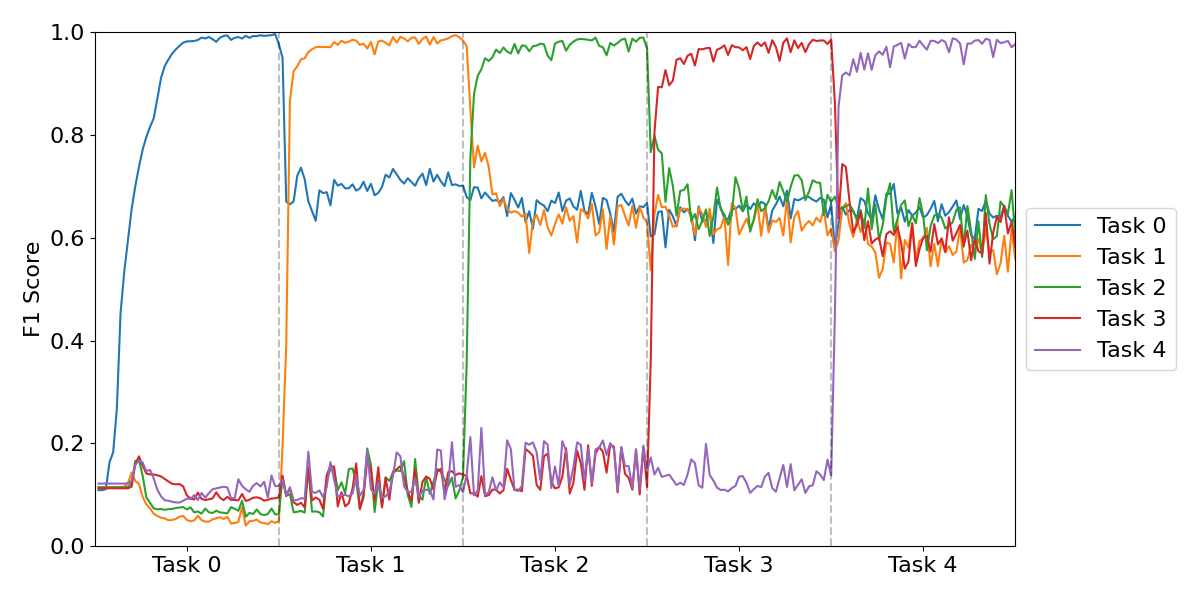}
        \caption{\(n=20\)}
    \end{subfigure}
    
    \caption{Task Performances for Pseudorehearsal using a Rehearsal Proportion of \(0.05\).}
\end{figure}

\begin{figure}[H]
    \centering
    \begin{subfigure}[t]{0.43\textwidth}
        \centering
        \includegraphics[width=\textwidth]{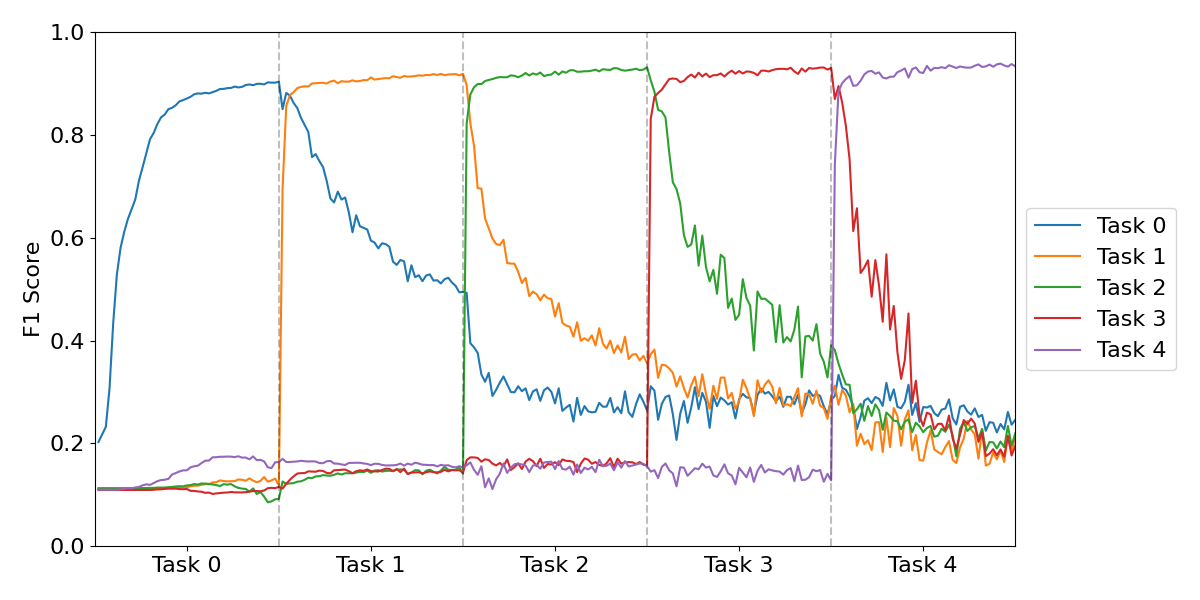}
        \caption{\(n=2\)}
    \end{subfigure}
    \begin{subfigure}[t]{0.43\textwidth}
        \centering
        \includegraphics[width=\textwidth]{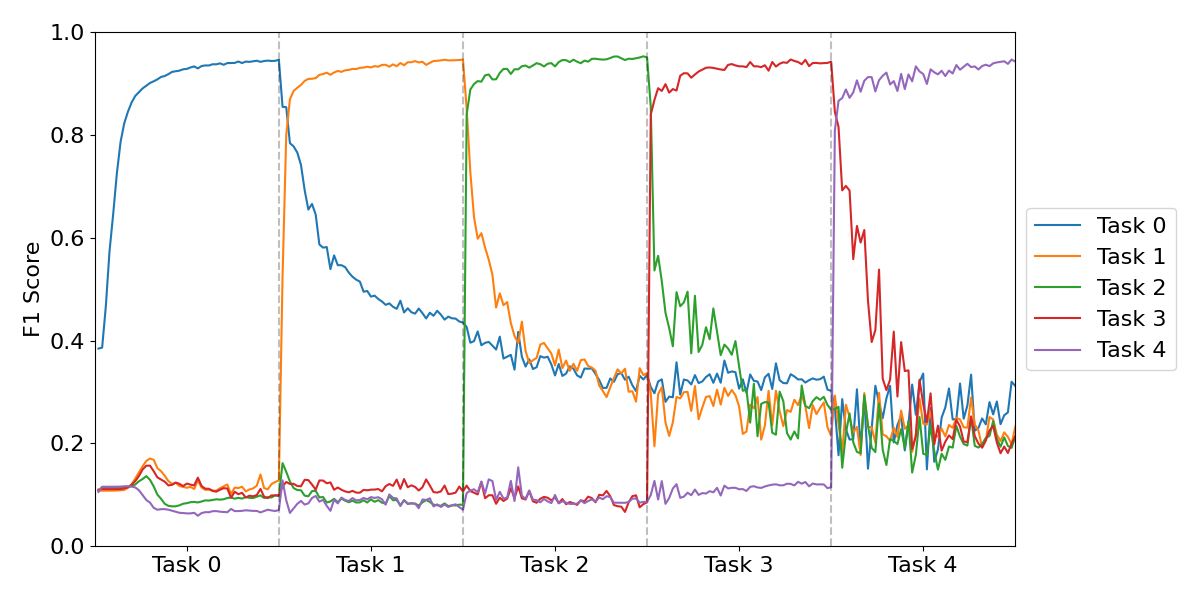}
        \caption{\(n=5\)}
    \end{subfigure}
    \hfill  
    \begin{subfigure}[t]{0.43\textwidth}
        \centering
        \includegraphics[width=\textwidth]{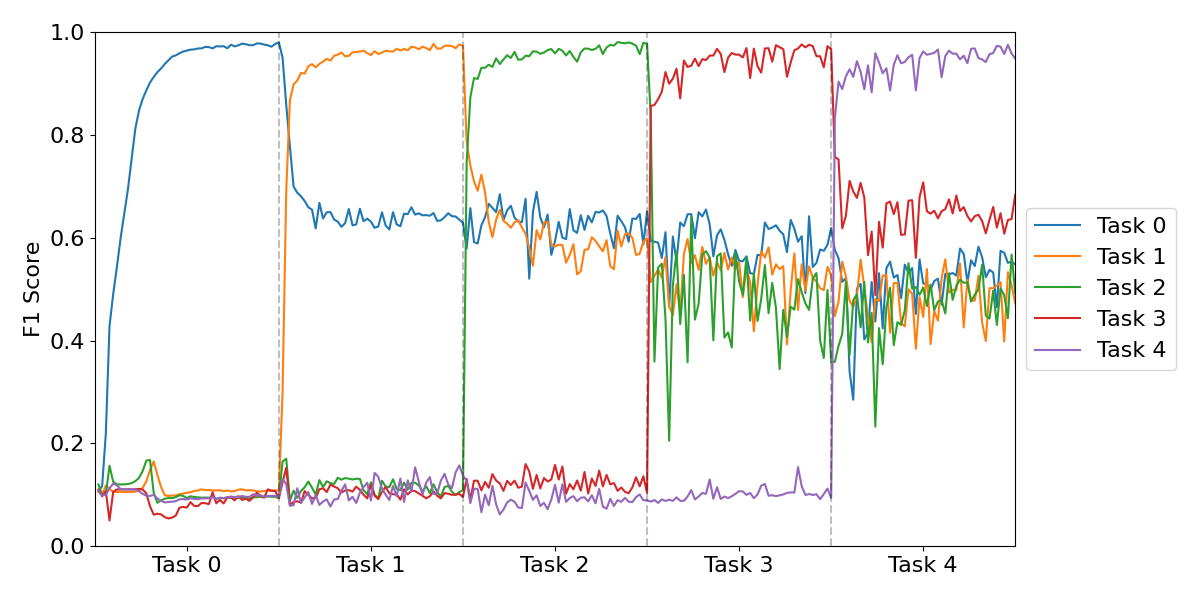}
        \caption{\(n=10\)}
    \end{subfigure}
    \begin{subfigure}[t]{0.43\textwidth}
        \centering
        \includegraphics[width=\textwidth]{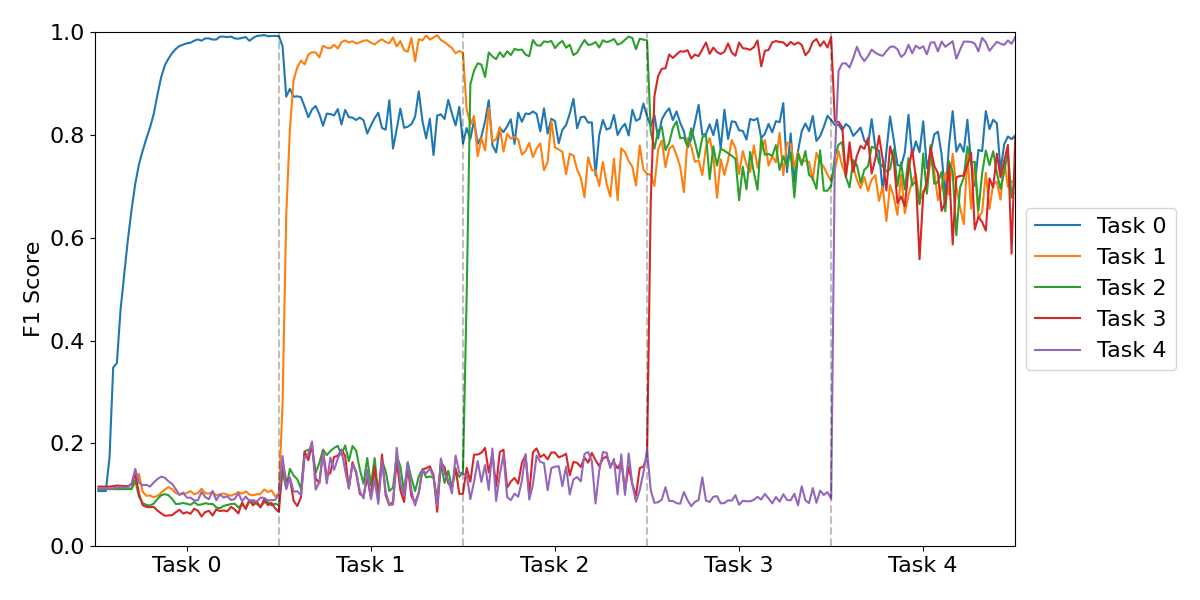}
        \caption{\(n=20\)}
    \end{subfigure}
    
    \caption{Task Performances for Pseudorehearsal using a Rehearsal Proportion of \(0.5\).}
\end{figure}

\begin{figure}[H]
    \centering
    \begin{subfigure}[t]{0.43\textwidth}
        \centering
        \includegraphics[width=\textwidth]{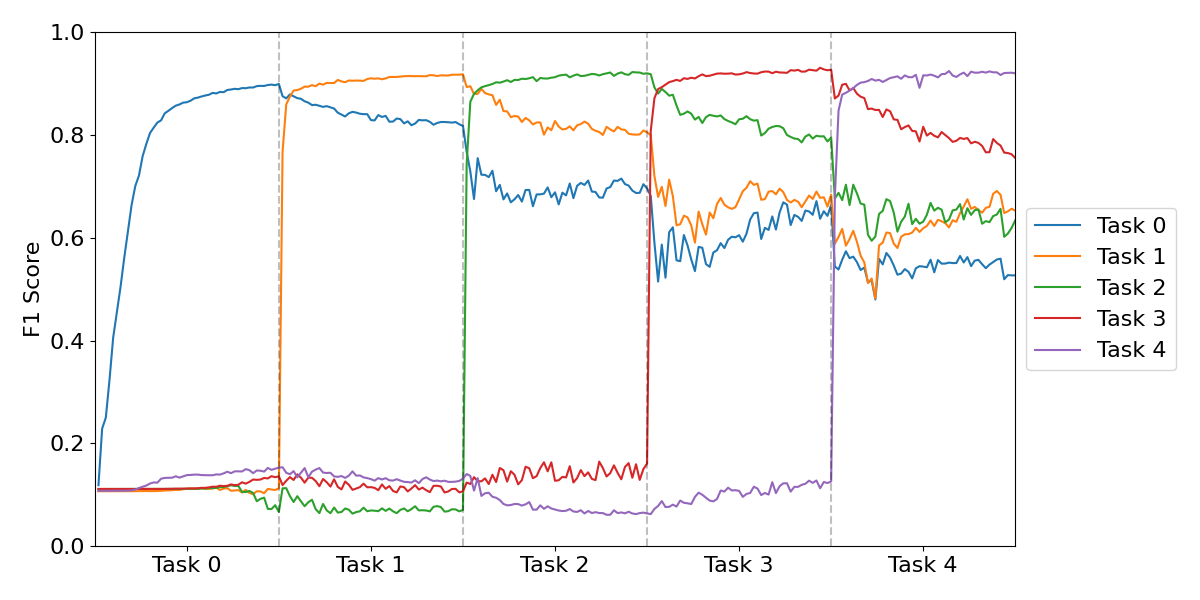}
        \caption{\(n=2\)}
    \end{subfigure}
    \begin{subfigure}[t]{0.43\textwidth}
        \centering
        \includegraphics[width=\textwidth]{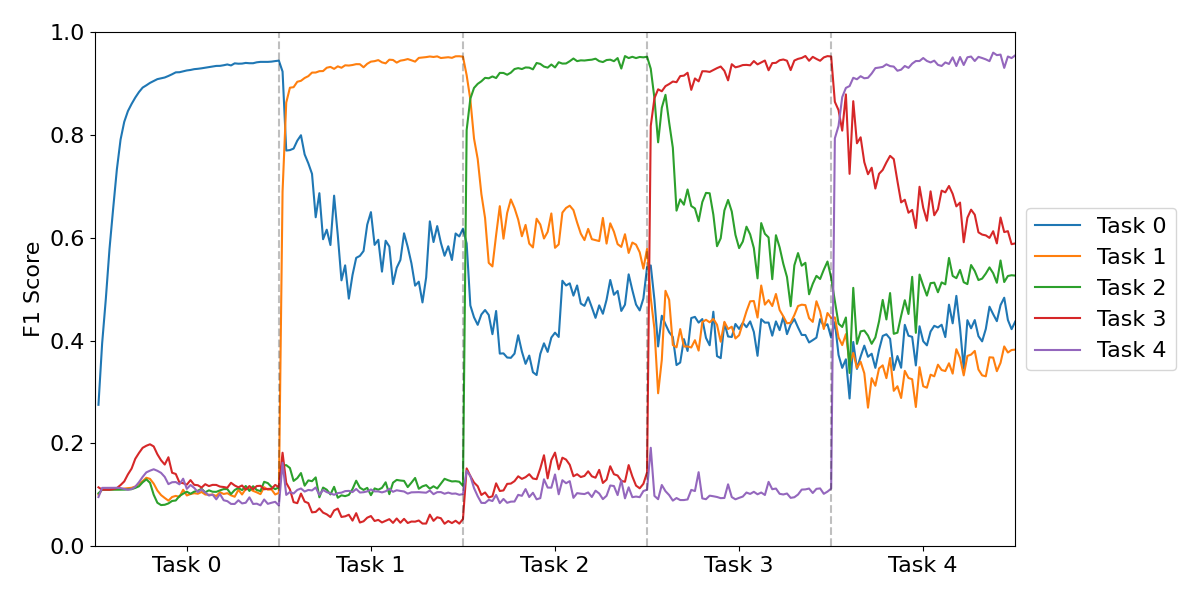}
        \caption{\(n=5\)}
    \end{subfigure}
    \hfill  
    \begin{subfigure}[t]{0.43\textwidth}
        \centering
        \includegraphics[width=\textwidth]{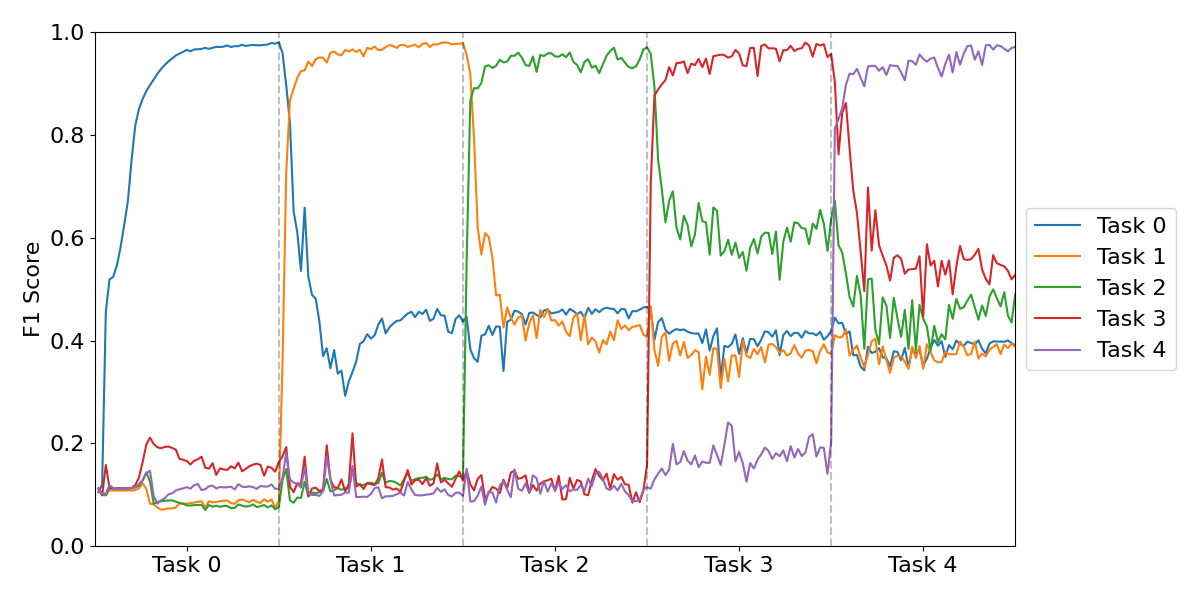}
        \caption{\(n=10\)}
    \end{subfigure}
    \begin{subfigure}[t]{0.43\textwidth}
        \centering
        \includegraphics[width=\textwidth]{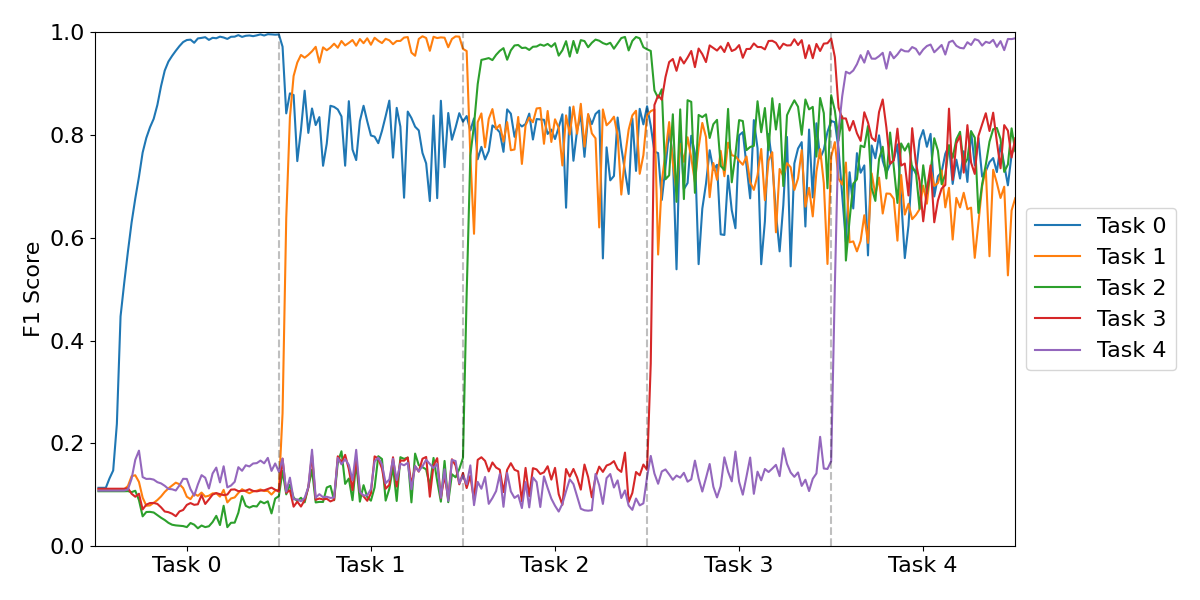}
        \caption{\(n=20\)}
    \end{subfigure}
    
    \caption{Task Performances for GEM using a Rehearsal Proportion of \(0.05\).}
\end{figure}

\begin{figure}[H]
    \centering
    \begin{subfigure}[t]{0.43\textwidth}
        \centering
        \includegraphics[width=\textwidth]{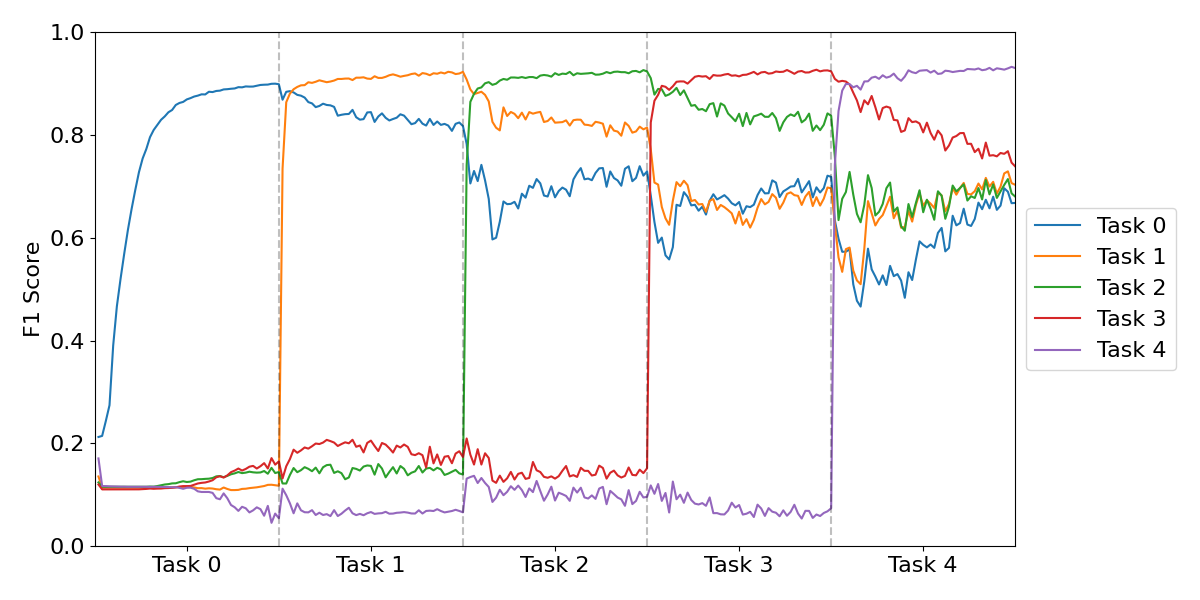}
        \caption{\(n=2\)}
    \end{subfigure}
    \begin{subfigure}[t]{0.43\textwidth}
        \centering
        \includegraphics[width=\textwidth]{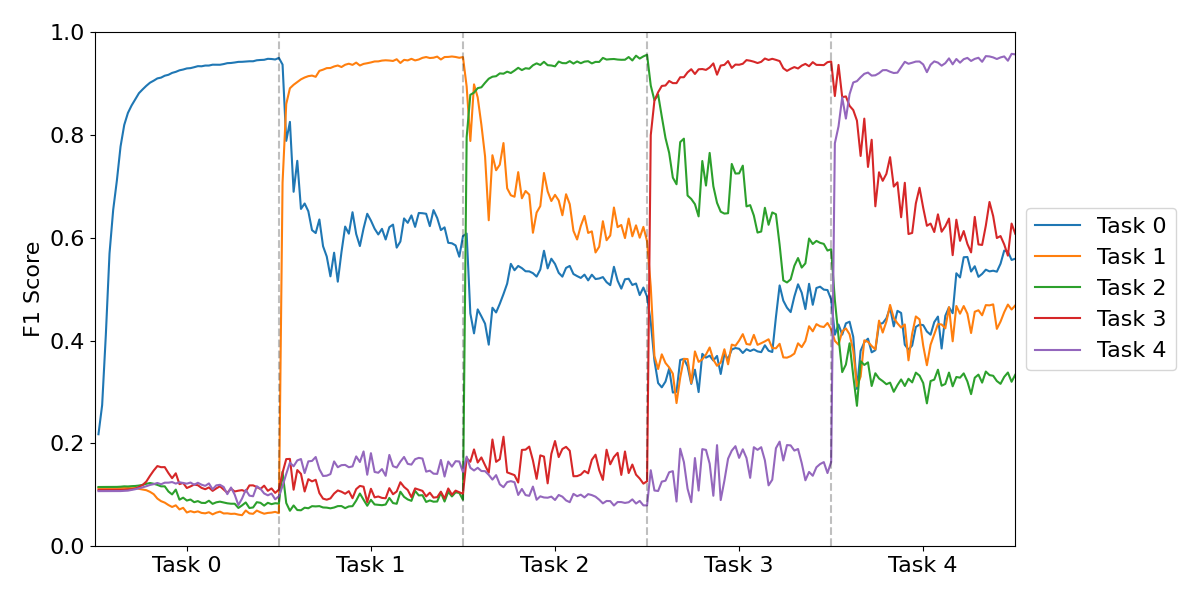}
        \caption{\(n=5\)}
    \end{subfigure}
    \hfill  
    \begin{subfigure}[t]{0.43\textwidth}
        \centering
        \includegraphics[width=\textwidth]{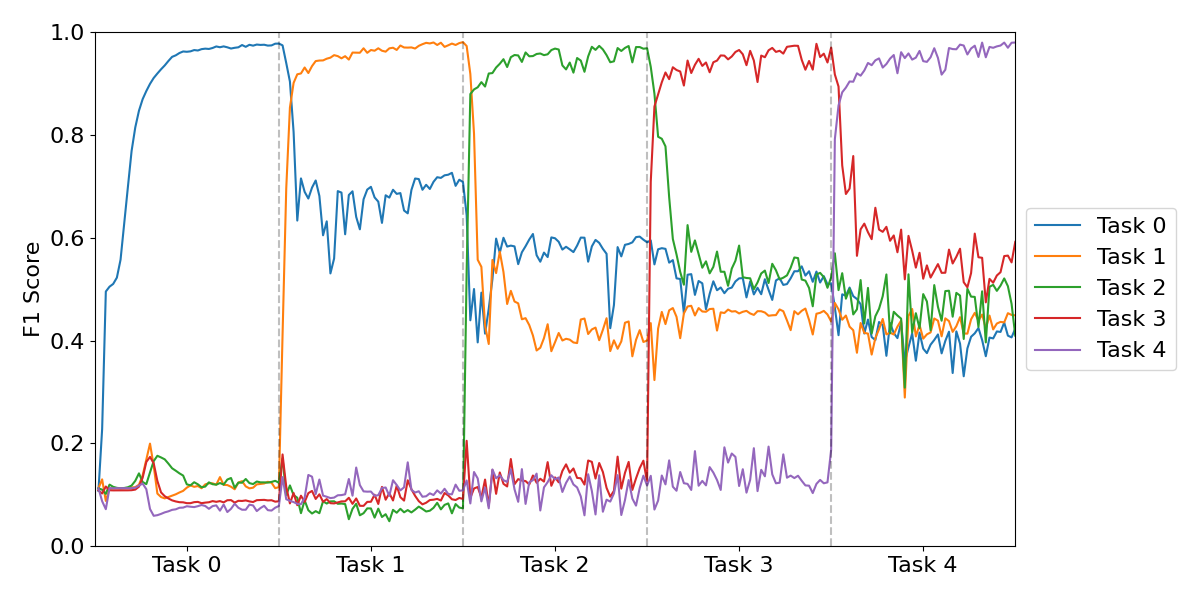}
        \caption{\(n=10\)}
    \end{subfigure}
    \begin{subfigure}[t]{0.43\textwidth}
        \centering
        \includegraphics[width=\textwidth]{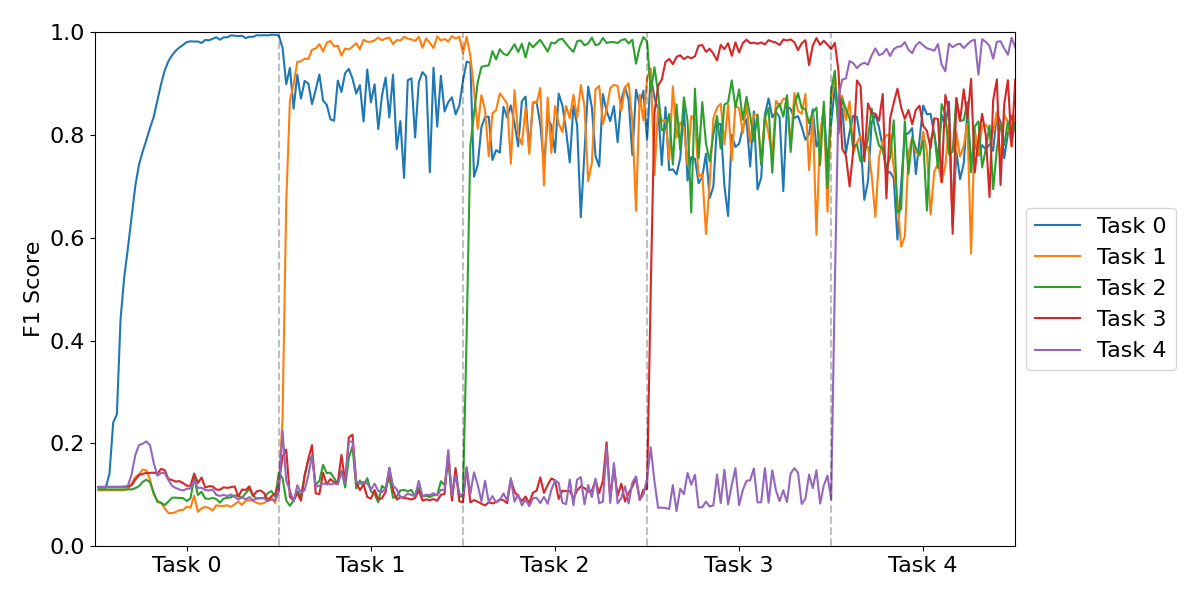}
        \caption{\(n=20\)}
    \end{subfigure}
    
    \caption{Task Performances for GEM using a Rehearsal Proportion of \(0.5\).}
\end{figure}

\begin{figure}[H]
    \centering
    \begin{subfigure}[t]{0.43\textwidth}
        \centering
        \includegraphics[width=\textwidth]{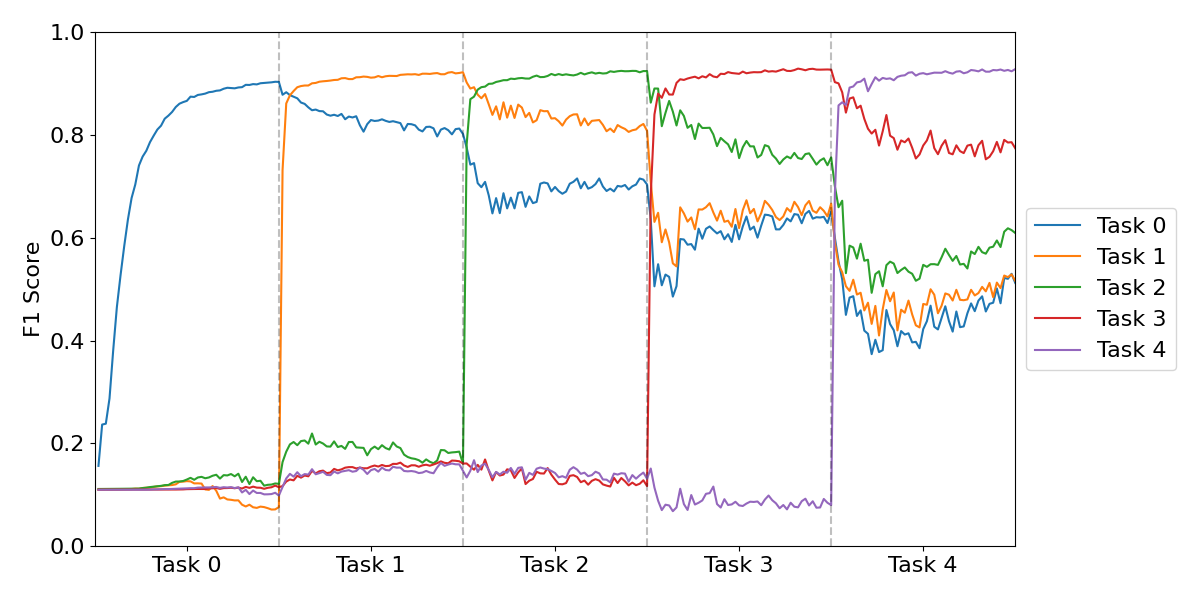}
        \caption{\(n=2\)}
    \end{subfigure}
    \begin{subfigure}[t]{0.43\textwidth}
        \centering
        \includegraphics[width=\textwidth]{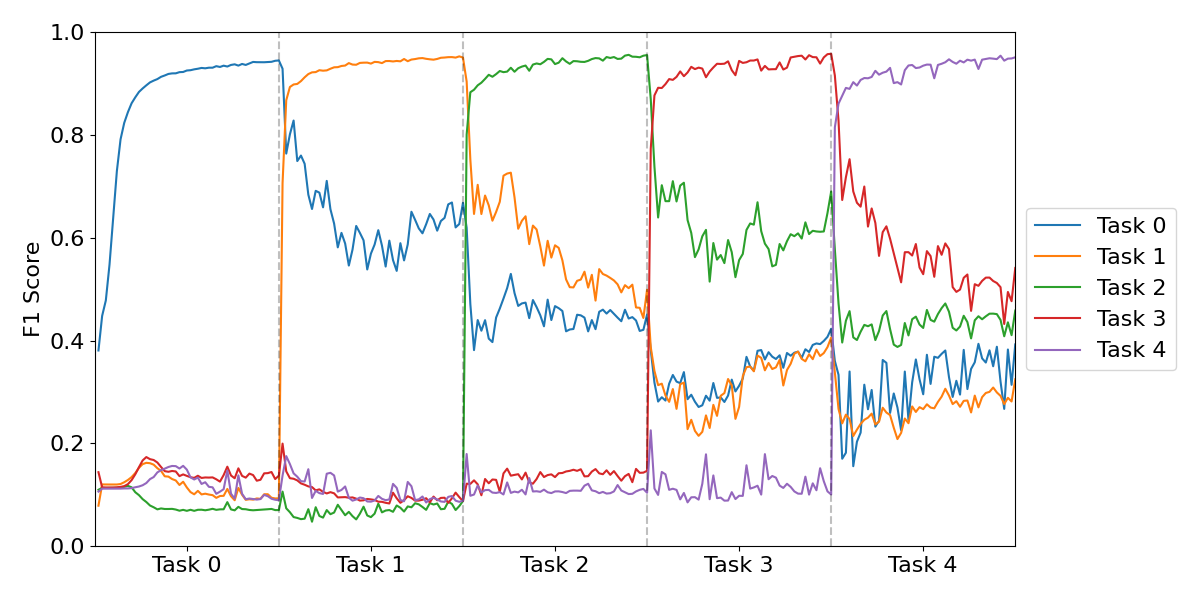}
        \caption{\(n=5\)}
    \end{subfigure}
    \hfill  
    \begin{subfigure}[t]{0.43\textwidth}
        \centering
        \includegraphics[width=\textwidth]{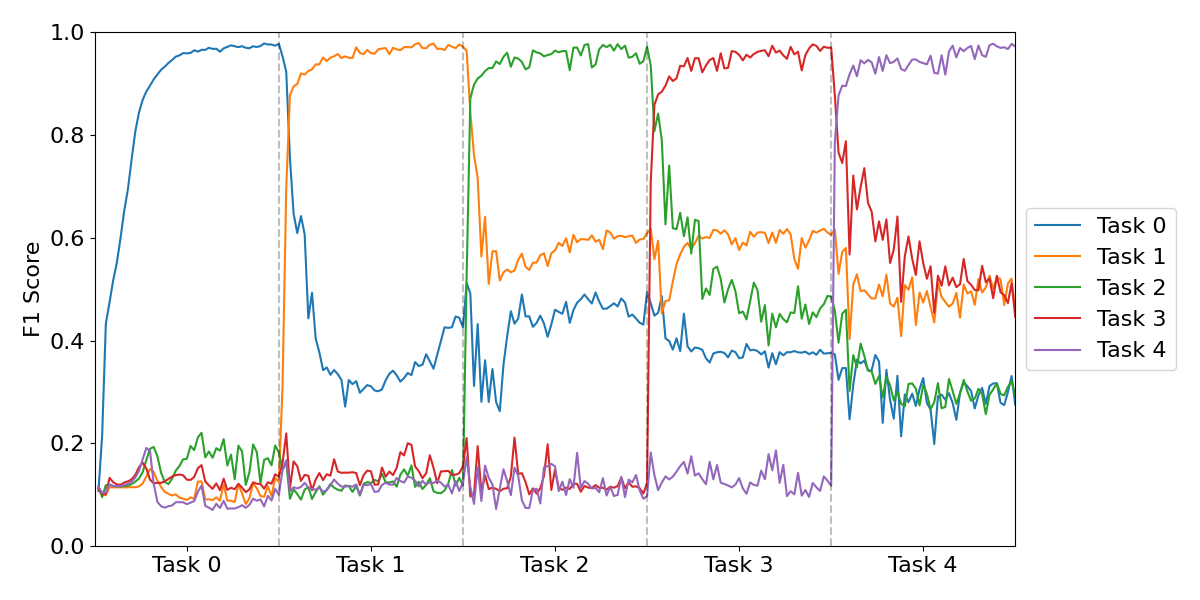}
        \caption{\(n=10\)}
    \end{subfigure}
    \begin{subfigure}[t]{0.43\textwidth}
        \centering
        \includegraphics[width=\textwidth]{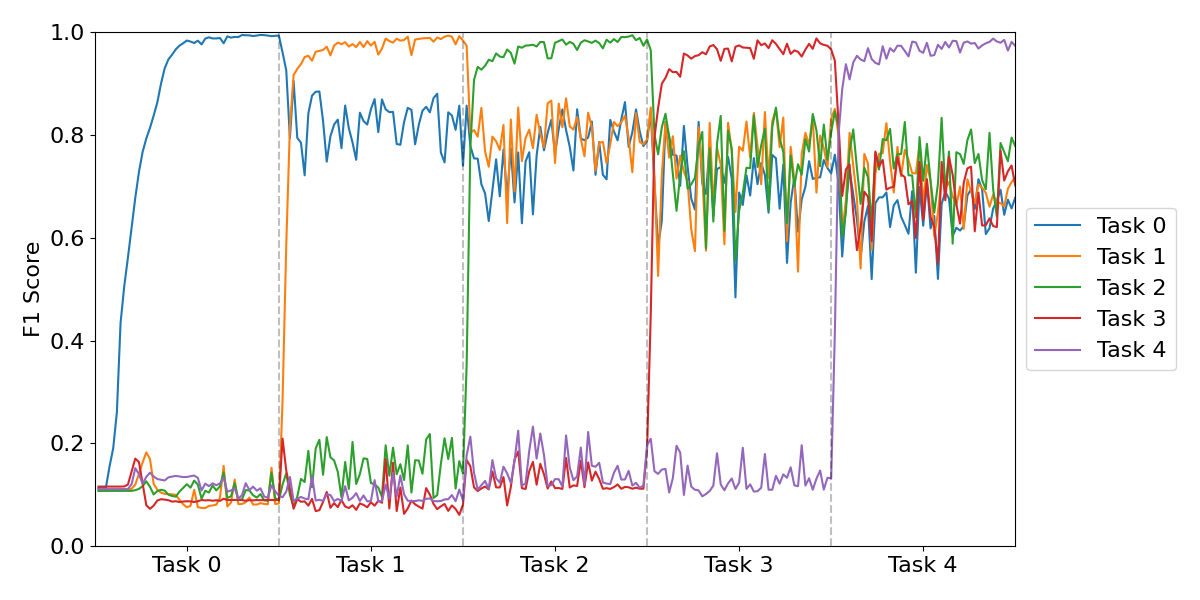}
        \caption{\(n=20\)}
    \end{subfigure}
    
    \caption{Task Performances for A-GEM using a Rehearsal Proportion of \(0.05\).}
\end{figure}

\begin{figure}[H]
    \centering
    \begin{subfigure}[t]{0.43\textwidth}
        \centering
        \includegraphics[width=\textwidth]{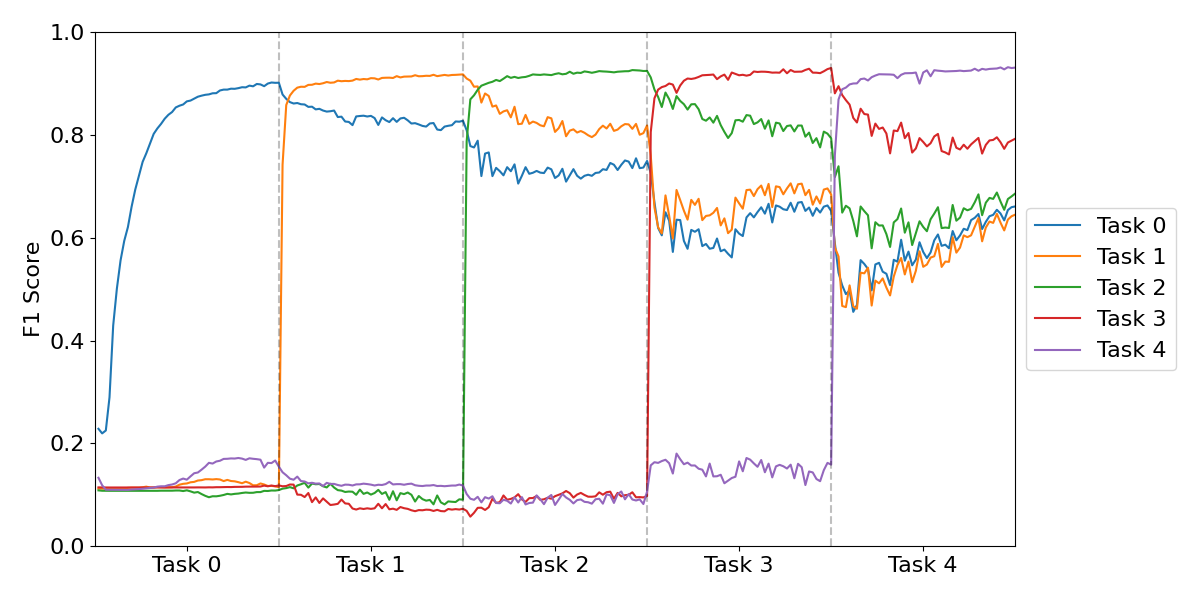}
        \caption{\(n=2\)}
    \end{subfigure}
    \begin{subfigure}[t]{0.43\textwidth}
        \centering
        \includegraphics[width=\textwidth]{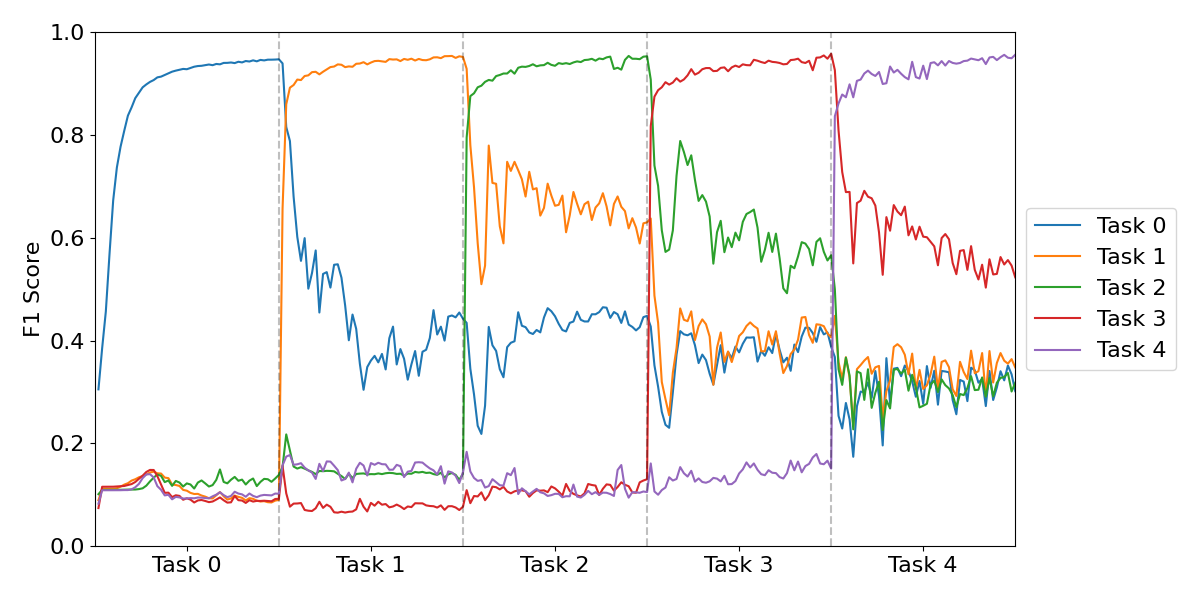}
        \caption{\(n=5\)}
    \end{subfigure}
    \hfill  
    \begin{subfigure}[t]{0.43\textwidth}
        \centering
        \includegraphics[width=\textwidth]{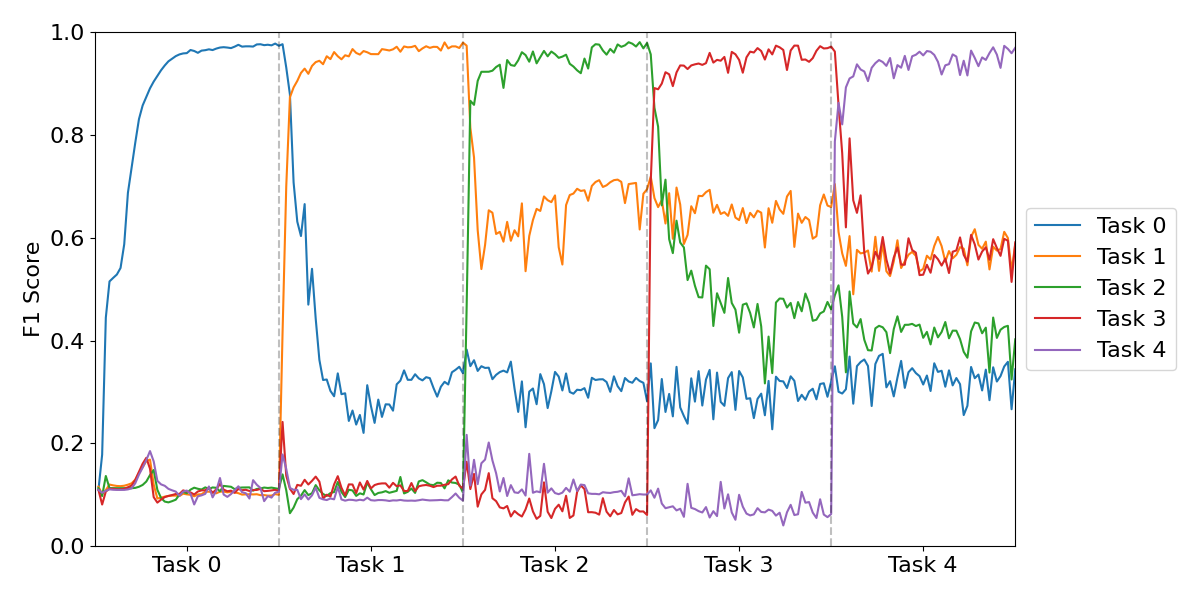}
        \caption{\(n=10\)}
    \end{subfigure}
    \begin{subfigure}[t]{0.43\textwidth}
        \centering
        \includegraphics[width=\textwidth]{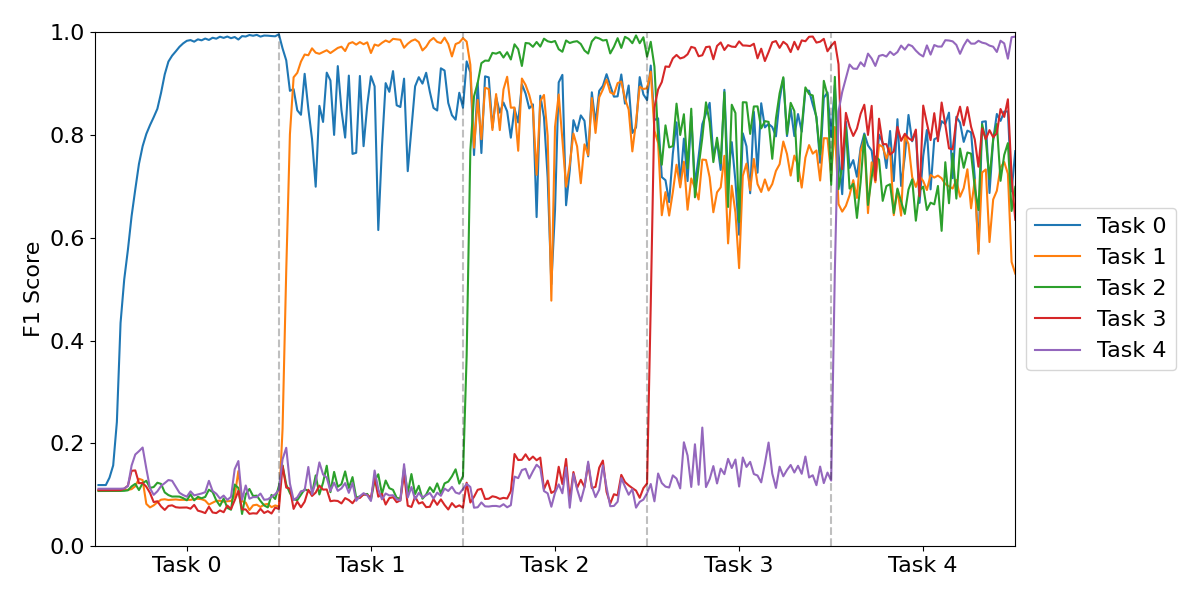}
        \caption{\(n=20\)}
    \end{subfigure}
    
    \caption{Task Performances for A-GEM using a Rehearsal Proportion of \(0.5\).}
\end{figure}

\begin{figure}[H]
    \centering
    \begin{subfigure}[t]{0.43\textwidth}
        \centering
        \includegraphics[width=\textwidth]{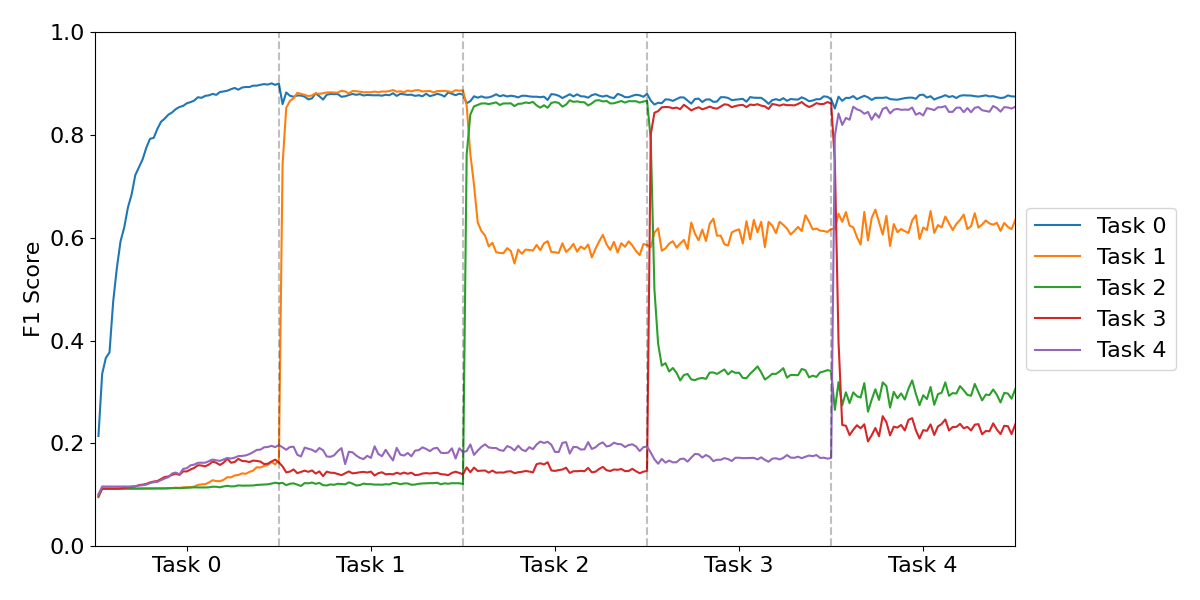}
        \caption{\(n=2\)}
    \end{subfigure}
    \begin{subfigure}[t]{0.43\textwidth}
        \centering
        \includegraphics[width=\textwidth]{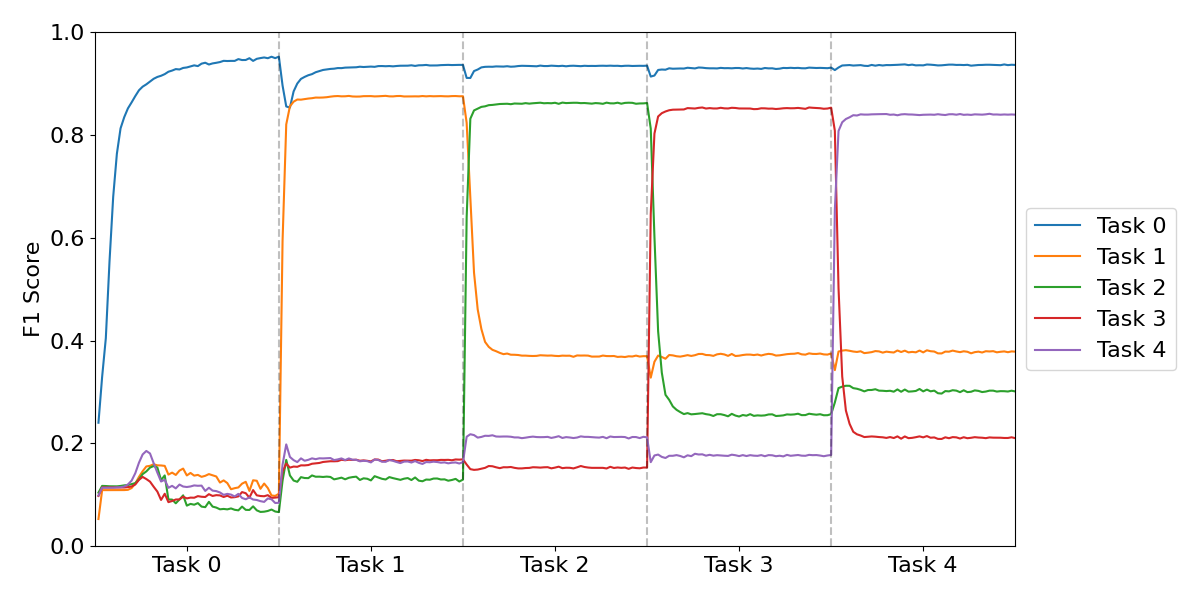}
        \caption{\(n=5\)}
    \end{subfigure}
    \hfill  
    \begin{subfigure}[t]{0.43\textwidth}
        \centering
        \includegraphics[width=\textwidth]{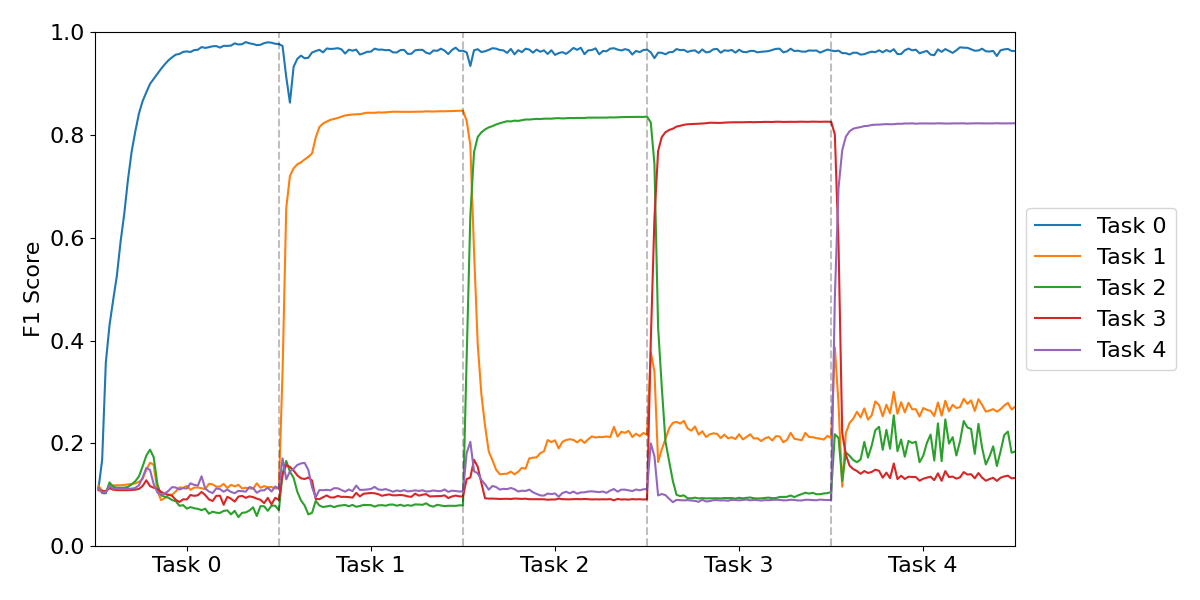}
        \caption{\(n=10\)}
    \end{subfigure}
    \begin{subfigure}[t]{0.43\textwidth}
        \centering
        \includegraphics[width=\textwidth]{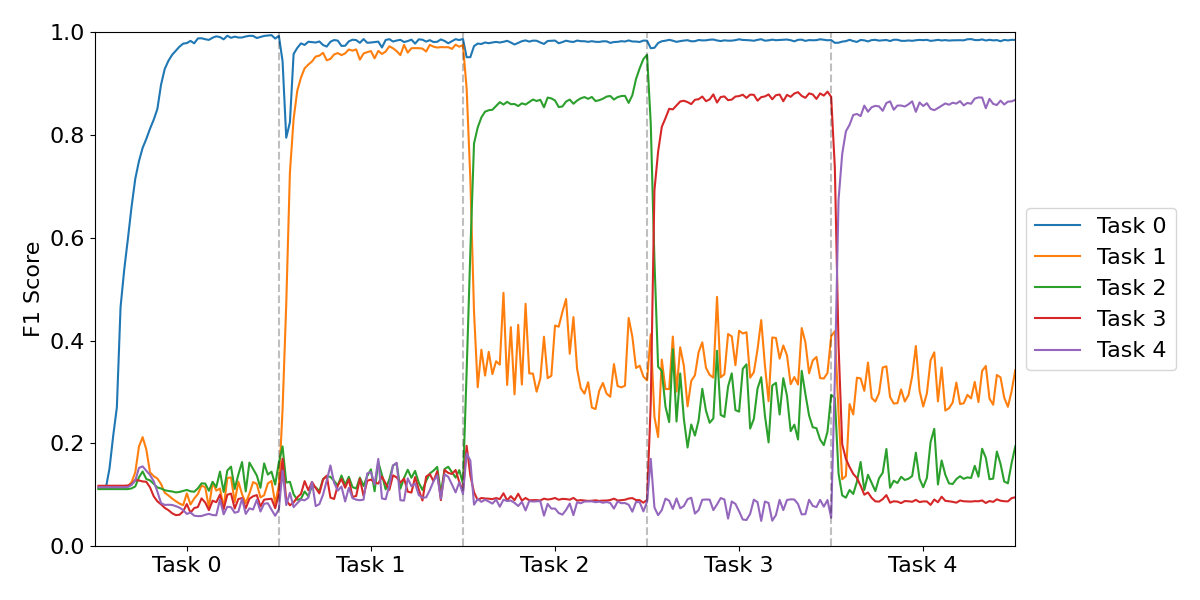}
        \caption{\(n=20\)}
    \end{subfigure}
    
    \caption{Task Performances for L2 Regularization.}
\end{figure}

\begin{figure}[H]
    \centering
    \begin{subfigure}[t]{0.43\textwidth}
        \centering
        \includegraphics[width=\textwidth]{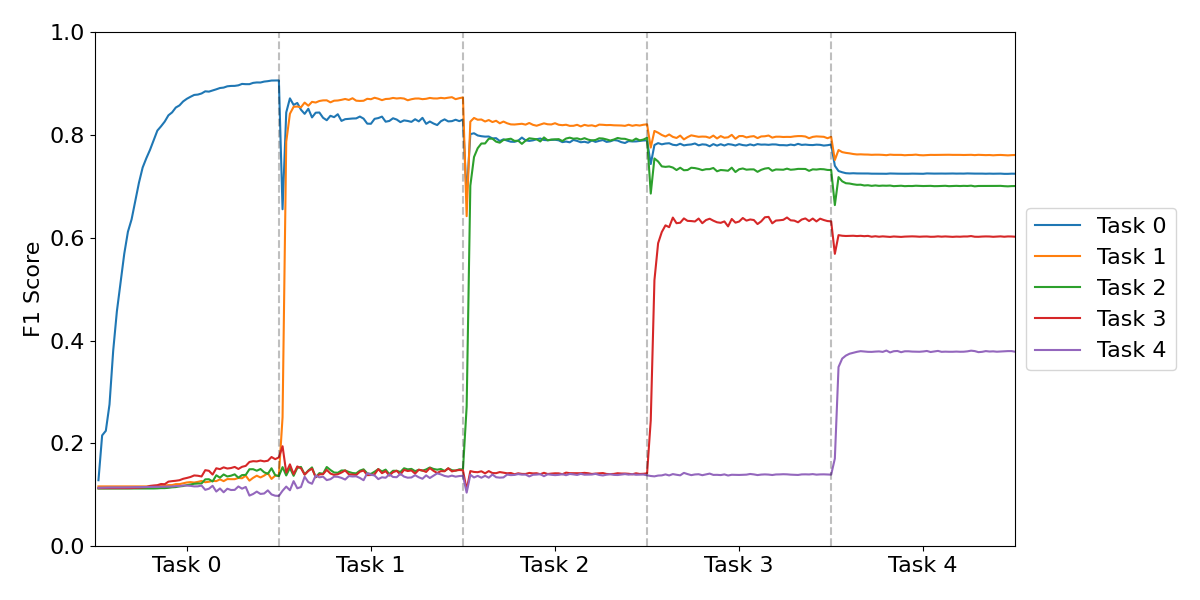}
        \caption{\(n=2\)}
    \end{subfigure}
    \begin{subfigure}[t]{0.43\textwidth}
        \centering
        \includegraphics[width=\textwidth]{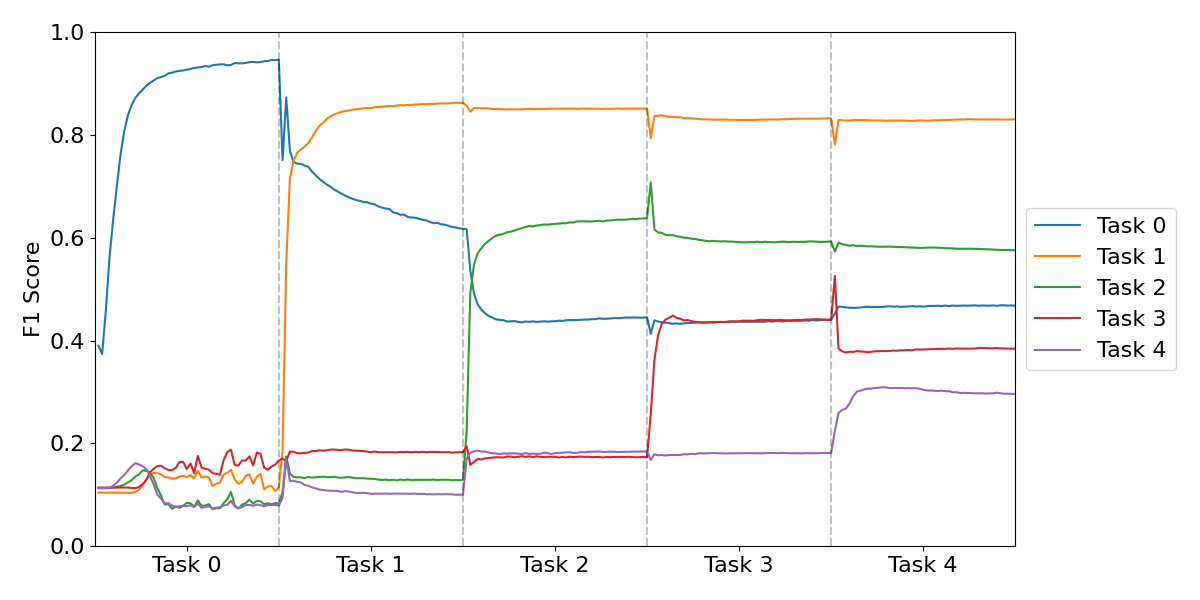}
        \caption{\(n=5\)}
    \end{subfigure}
    \hfill  
    \begin{subfigure}[t]{0.43\textwidth}
        \centering
        \includegraphics[width=\textwidth]{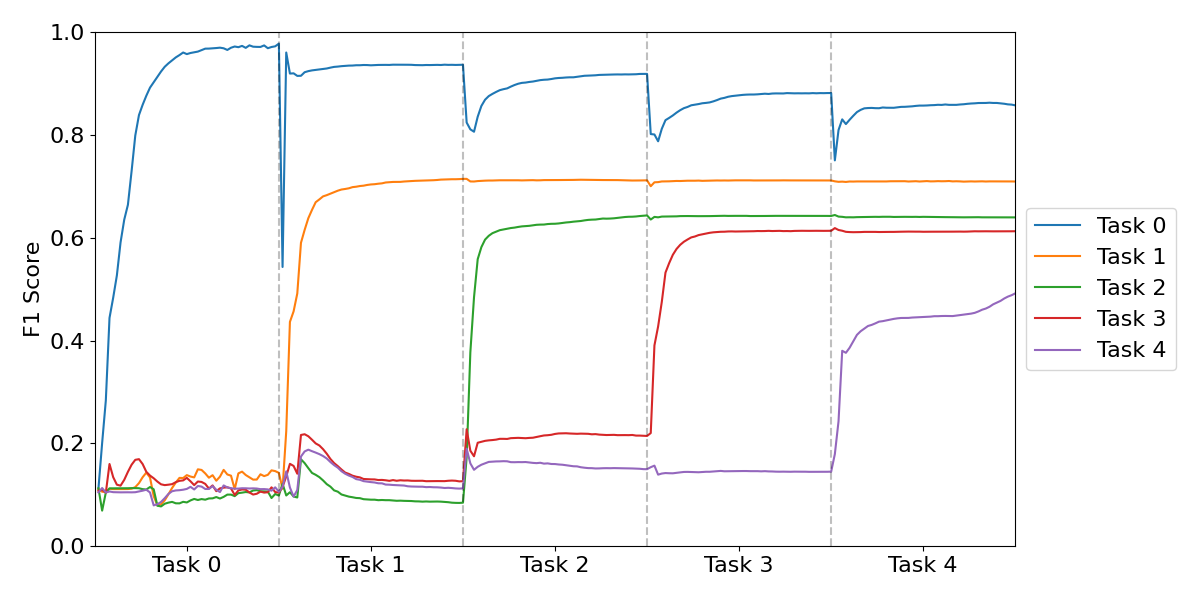}
        \caption{\(n=10\)}
    \end{subfigure}
    \begin{subfigure}[t]{0.43\textwidth}
        \centering
        \includegraphics[width=\textwidth]{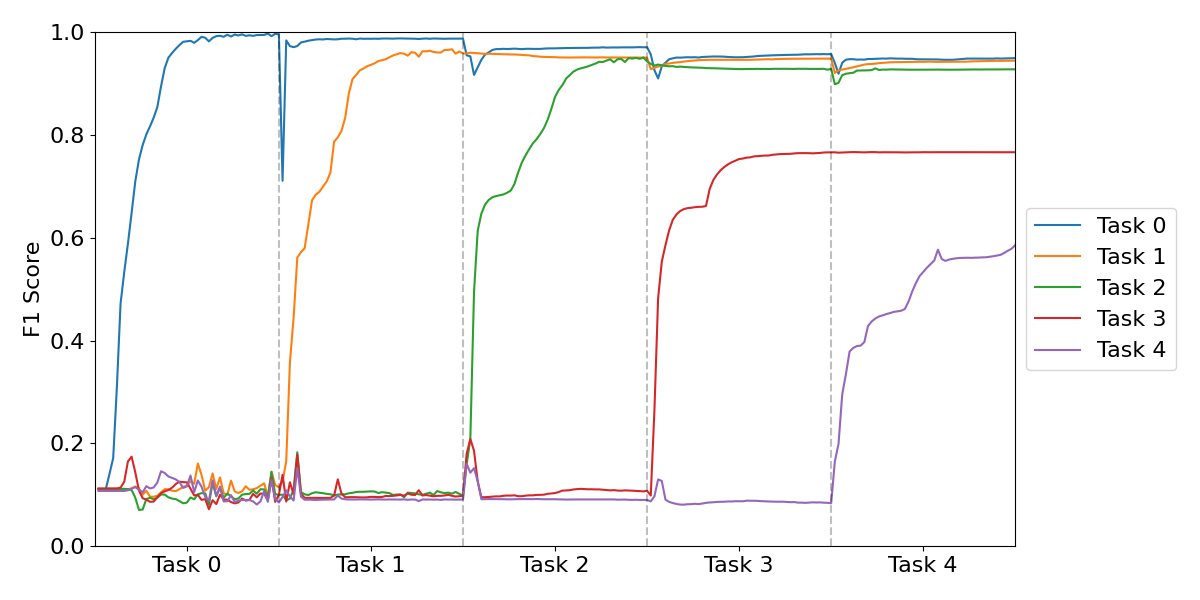}
        \caption{\(n=20\)}
    \end{subfigure}
    
    \caption{Task Performances for Elastic Weight Consolidation.}
\end{figure}

\begin{figure}[H]
    \centering
    \begin{subfigure}[t]{0.43\textwidth}
        \centering
        \includegraphics[width=\textwidth]{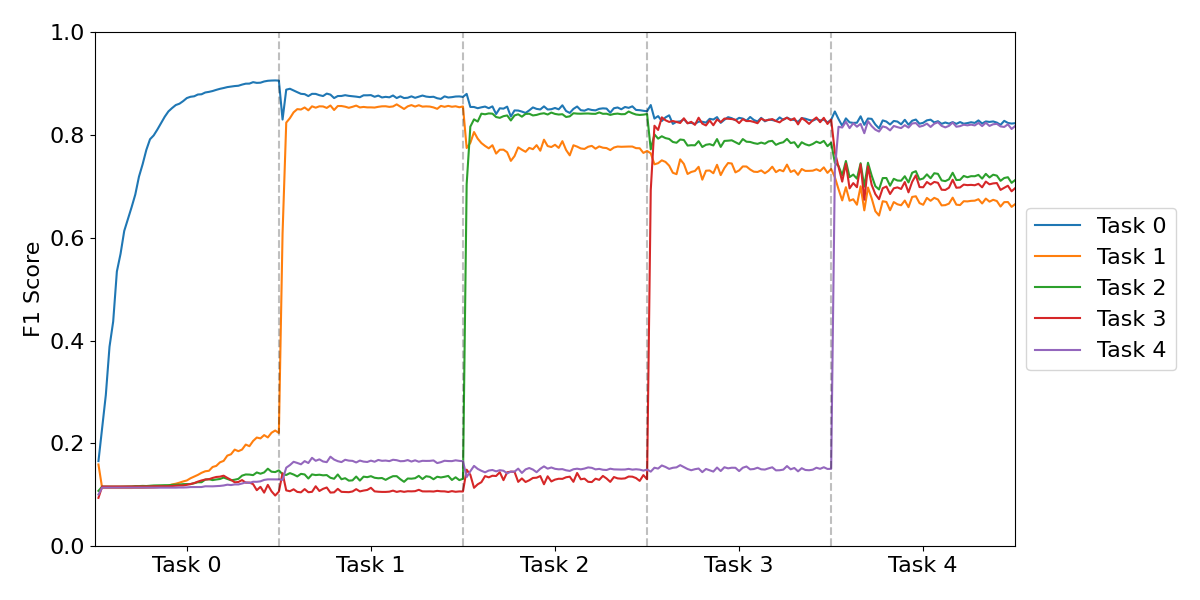}
        \caption{\(n=2\)}
    \end{subfigure}
    \begin{subfigure}[t]{0.43\textwidth}
        \centering
        \includegraphics[width=\textwidth]{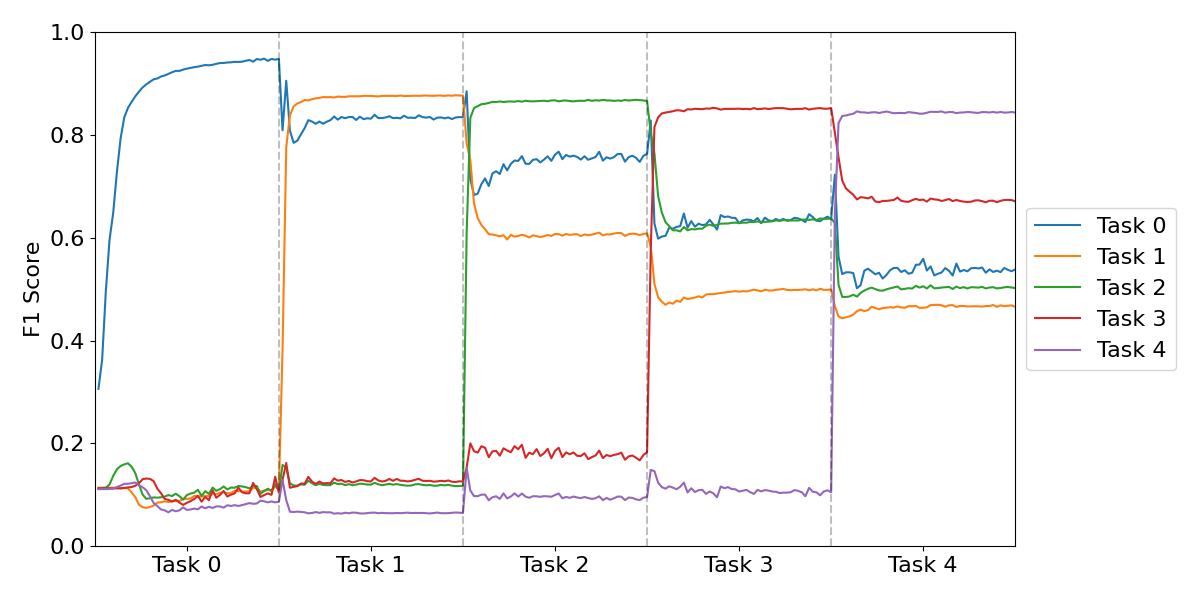}
        \caption{\(n=5\)}
    \end{subfigure}
    \hfill  
    \begin{subfigure}[t]{0.43\textwidth}
        \centering
        \includegraphics[width=\textwidth]{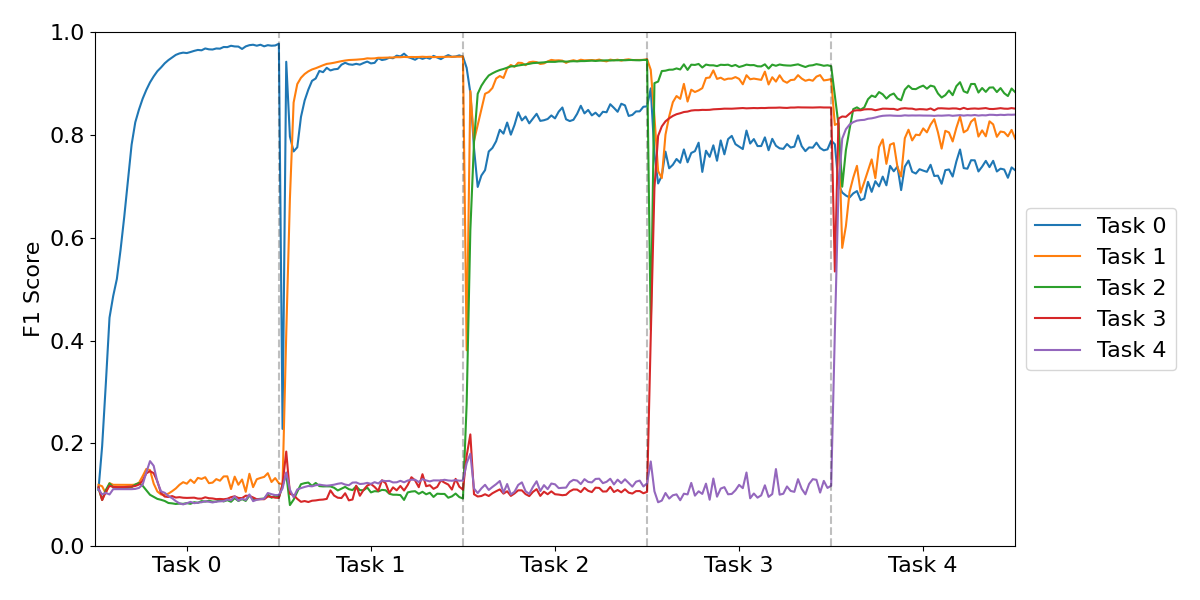}
        \caption{\(n=10\)}
    \end{subfigure}
    \begin{subfigure}[t]{0.43\textwidth}
        \centering
        \includegraphics[width=\textwidth]{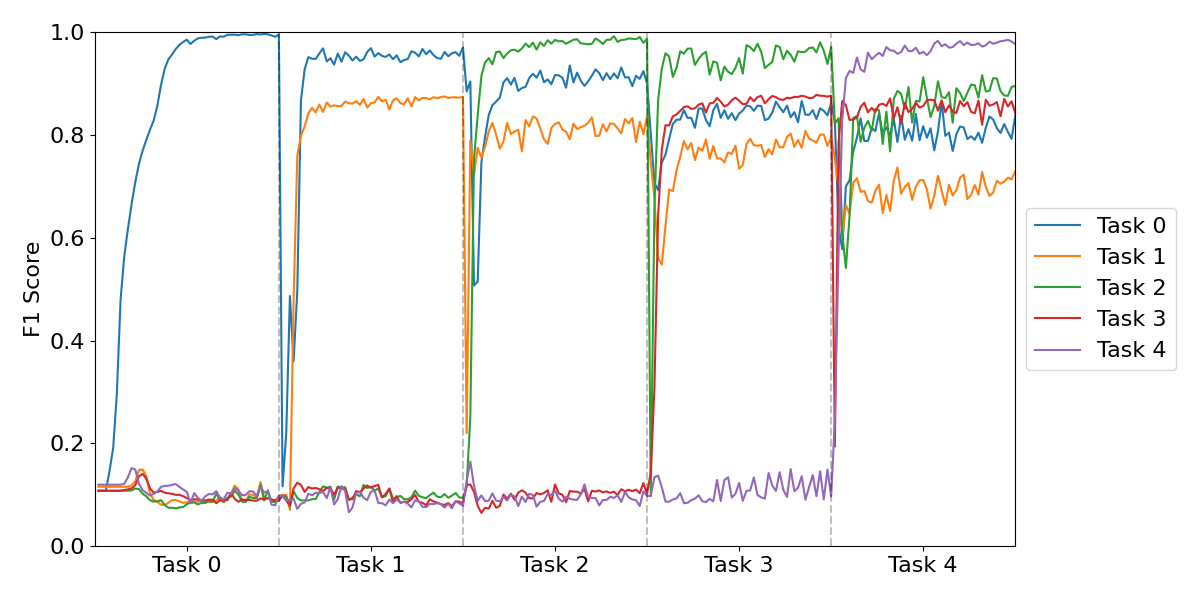}
        \caption{\(n=20\)}
    \end{subfigure}
    
    \caption{Task Performances for Memory Aware Synapses.}
\end{figure}

\begin{figure}[H]
    \centering
    \begin{subfigure}[t]{0.43\textwidth}
        \centering
        \includegraphics[width=\textwidth]{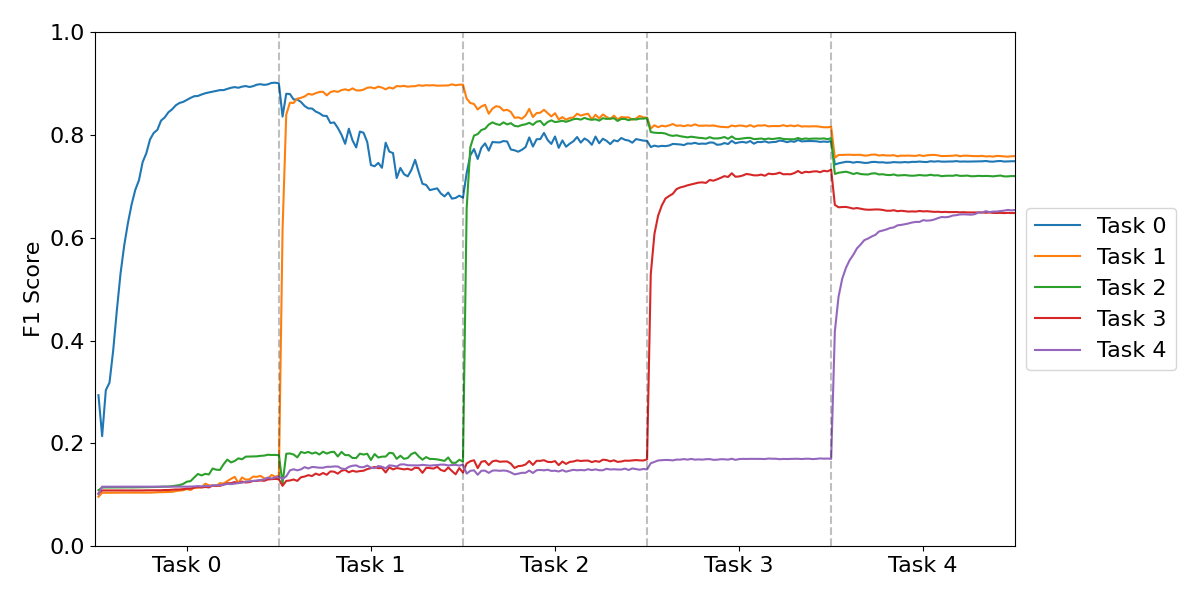}
        \caption{\(n=2\)}
    \end{subfigure}
    \begin{subfigure}[t]{0.43\textwidth}
        \centering
        \includegraphics[width=\textwidth]{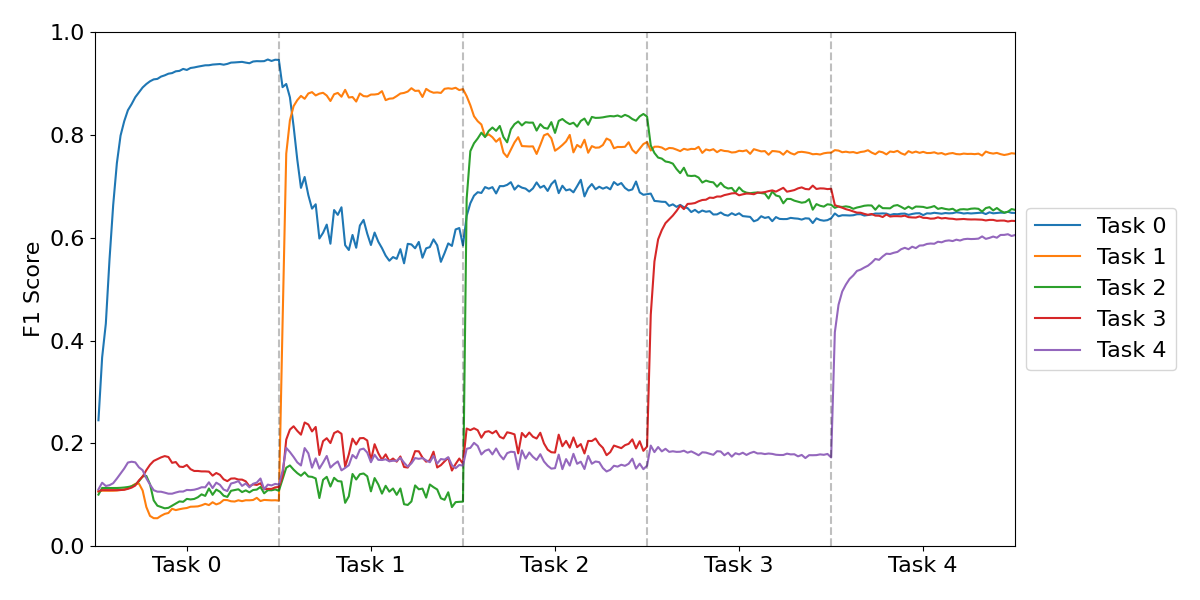}
        \caption{\(n=5\)}
    \end{subfigure}
    \hfill  
    \begin{subfigure}[t]{0.43\textwidth}
        \centering
        \includegraphics[width=\textwidth]{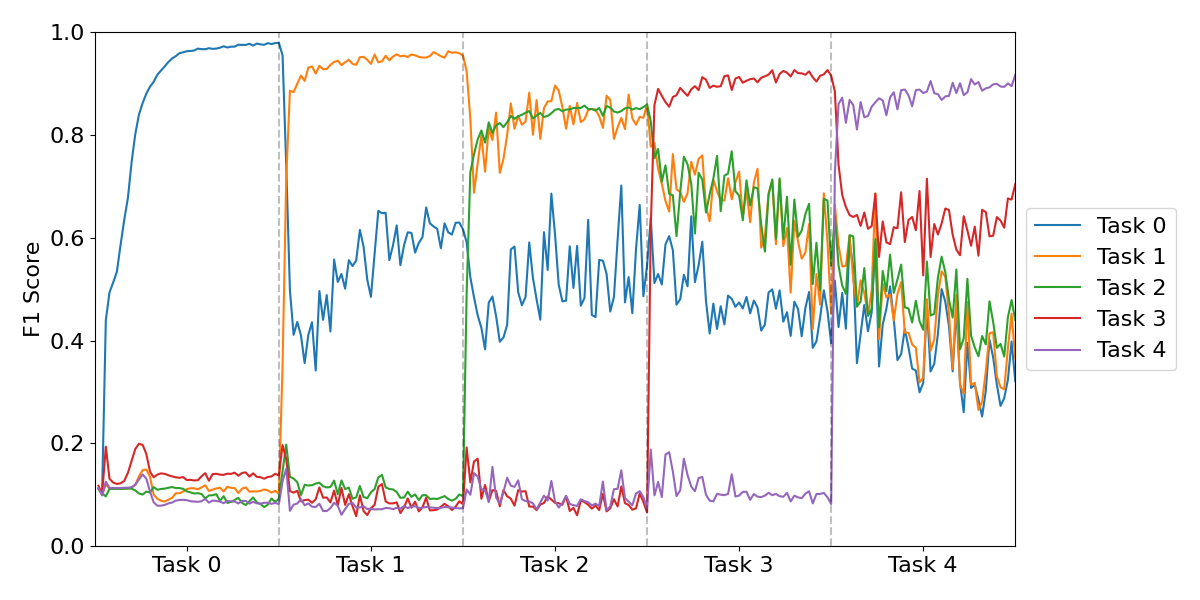}
        \caption{\(n=10\)}
    \end{subfigure}
    \begin{subfigure}[t]{0.43\textwidth}
        \centering
        \includegraphics[width=\textwidth]{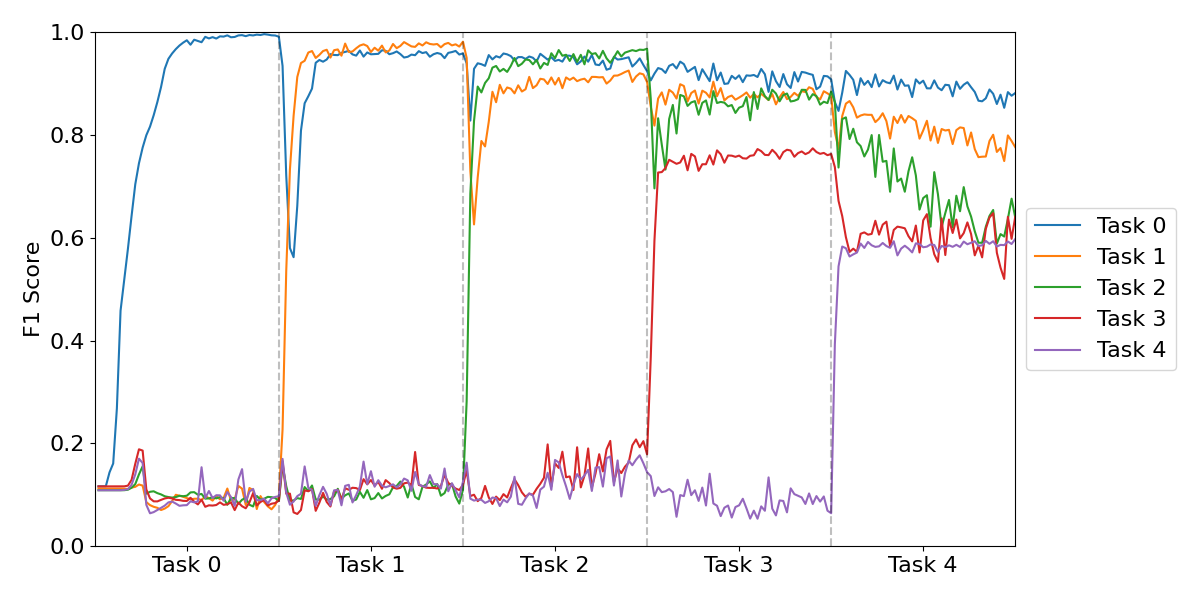}
        \caption{\(n=20\)}
    \end{subfigure}
    
    \caption{Task Performances for Synaptic Intelligence.}
\end{figure}

\end{document}